\pdfoutput=1
\documentclass[11pt,a4paper]{article}
\usepackage{times,latexsym}
\usepackage{url}
\usepackage[T1]{fontenc}
\newcommand{\codeurl}{https://github.com/GaoJianXin05/RuleCraft}
\newcommand{\dataurl}{https://huggingface.co/datasets/GGjx/RuleCraft}
\newcommand{\caseok}{\textcolor{green!45!black}{$\checkmark$}}
\newcommand{\casereal}{\textcolor{orange!85!black}{\textit{real}}}
\usepackage[acceptedWithA]{tacl2021v1}\usepackage[utf8]{inputenc}
\usepackage{microtype}
\usepackage{inconsolata}
\usepackage{graphicx}
\usepackage{booktabs}
\usepackage{multirow}
\usepackage{array}
\usepackage{amsmath}
\usepackage{amssymb}
\usepackage{enumitem}
\usepackage{placeins}
\usepackage{colortbl}

\newcommand{\nCardDS}{101}
\newcommand{\nCardQw}{68}
\newcommand{\nCardSol}{77}

\newcommand{\nRevMechN}{18}

\newcommand{\nRevNeutConsDDS}{+0.0}
\newcommand{\nRevNeutConsDQw}{$-$0.6}

\newcommand{\nRevNeutDCIDS}{[2.1, 11.6]}
\newcommand{\nRevNeutDCIQw}{[1.7, 10.3]}
\newcommand{\nRevNeutDDS}{+6.7}
\newcommand{\nRevNeutDQw}{+6.0}

\newcommand{\nRevNeutShareDS}{34}
\newcommand{\nRevNeutShareQw}{50}

\newcommand{\nRevStableImageMax}{25.4}
\newcommand{\nRevStableImageMin}{11.8}

\newcommand{\nRevStableRetestMax}{13.4}

\providecommand{\nHumanConf}{78.3}
\providecommand{\nHumanConfCI}{[71.7, 84.1]}
\providecommand{\nHumanCons}{77.8}

\providecommand{\nHumanFallback}{74}

\providecommand{\nHumanHConf}{84.3}
\providecommand{\nHumanHConfCI}{[77.9, 89.7]}
\providecommand{\nHumanHCons}{74.2}

\providecommand{\nHumanHFallback}{85}

\providecommand{\nHumanHoursText}{none (1), 10--100 hours (6), 100--500 hours (5), over 500 hours (3)}
\providecommand{\nHumanN}{15}

\newcommand{\aBalHardDirectQF}{59.0}
\newcommand{\aBalHardDirectSol}{68.2}

\newcommand{\aBalHardToldQF}{72.5}
\newcommand{\aBalHardToldSol}{98.1}

\newcommand{\aBlindModels}{six}
\newcommand{\aBlindN}{seven}
\newcommand{\aBlindReal}{100}

\newcommand{\aCoTMentionSol}{38}

\newcommand{\aGDCvsDirectHSCIDS}{[$-$49.0, $-$14.1]}
\newcommand{\aGDCvsDirectHSCIQF}{[$-$25.3, $-$3.5]}

\newcommand{\aGDCvsDirectHSDS}{$-$30.4}
\newcommand{\aGDCvsDirectHSQF}{$-$14.4}

\newcommand{\aGDCvsFixedGameHCCIDS}{[3.9, 27.1]}
\newcommand{\aGDCvsFixedGameHCCIQF}{[4.4, 16.6]}

\newcommand{\aGDCvsFixedGameHCDS}{+15.3}
\newcommand{\aGDCvsFixedGameHCQF}{+9.9}

\newcommand{\aHAGapMax}{18}
\newcommand{\aHAGapMin}{9}

\newcommand{\aHBalDiffMax}{24}
\newcommand{\aHBalDiffMin}{9}

\newcommand{\aHBiasImageMax}{$-$16}
\newcommand{\aHBiasImageMin}{$-$68}

\newcommand{\aHConsDropDS}{20.5}

\newcommand{\aHConsImageDP}{85.6}
\newcommand{\aHConsImageDS}{96.5}
\newcommand{\aHConsImageMax}{99}
\newcommand{\aHConsImageMin}{74}

\newcommand{\aHConsToldDP}{62.5}
\newcommand{\aHConsToldDS}{76.0}

\newcommand{\aHGapDP}{+41.5}
\newcommand{\aHGapDS}{+52.0}
\newcommand{\aHGapMax}{52.7}
\newcommand{\aHGapMin}{33.8}

\newcommand{\aHGapQM}{+33.8}

\newcommand{\aHGapSol}{+51.2}
\newcommand{\aHImageDL}{31.1}

\newcommand{\aHImageDS}{28.9}

\newcommand{\aHNAct}{10}

\newcommand{\aHNCons}{13}
\newcommand{\aHNModels}{six}

\newcommand{\aHNPas}{37}

\newcommand{\aHOverPooled}{97}
\newcommand{\aHOverTPooled}{99}

\newcommand{\aHPGapMax}{63}
\newcommand{\aHPGapMin}{40}

\newcommand{\aHPImageSol}{37.4}

\newcommand{\aHPTextMax}{100}
\newcommand{\aHPTextMin}{86}

\newcommand{\aHRealPooled}{96}

\newcommand{\aHSRulesBelow}{12}
\newcommand{\aHSRulesN}{13}

\newcommand{\aHToldSol}{98.2}

\newcommand{\aHUnusedMax}{67}
\newcommand{\aHUnusedMin}{34}

\newcommand{\aHighConsImage}{99.0}

\newcommand{\aHighDiff}{+8.3}
\newcommand{\aHighDiffCI}{[3.9, 13.7]}
\newcommand{\aHighGap}{+43.1}
\newcommand{\aHighGapCI}{[29.8, 56.6]}
\newcommand{\aHighImage}{55.3}

\newcommand{\aHighPImage}{45.6}

\newcommand{\aHighTold}{98.4}
\newcommand{\aInstrMax}{7}

\newcommand{\aMAGapMax}{12}
\newcommand{\aMAGapMin}{1}

\newcommand{\aMBiasImageMax}{$-$11}
\newcommand{\aMBiasImageMin}{$-$38}

\newcommand{\aMGapMax}{19.8}
\newcommand{\aMGapMin}{8.2}

\newcommand{\aMOverPooled}{94}
\newcommand{\aMOverTPooled}{97}

\newcommand{\aMPGapMax}{30}
\newcommand{\aMPGapMin}{13}

\newcommand{\aMRealPooled}{92}

\newcommand{\aMUnusedMax}{24}
\newcommand{\aMUnusedMin}{12}

\newcommand{\aMechGapSig}{all six}
\newcommand{\aMethHCCoTDS}{33.6}

\newcommand{\aMethHCDirectDS}{23.9}
\newcommand{\aMethHCDirectQF}{35.4}
\newcommand{\aMethHCDirectSol}{37.4}
\newcommand{\aMethHCFixedGameDS}{78.6}
\newcommand{\aMethHCFixedGameQF}{81.8}
\newcommand{\aMethHCFixedGameSol}{99.7}

\newcommand{\aMethHCToldDS}{85.0}
\newcommand{\aMethHCToldQF}{74.9}
\newcommand{\aMethHCToldSol}{100.0}

\newcommand{\aMethHSDirectDS}{96.5}
\newcommand{\aMethHSDirectQF}{82.7}

\newcommand{\aMethHSFixedGameQF}{61.9}

\newcommand{\aMethHSToldDS}{76.0}

\newcommand{\aMethHSToldSol}{96.2}

\newcommand{\aNConfigs}{eight}

\newcommand{\aNameOnlyShareMax}{90}
\newcommand{\aNameOnlyShareMin}{53}
\newcommand{\aNameVerShareMax}{99}
\newcommand{\aNameVerShareMin}{88}
\newcommand{\aNeutCostMax}{5.8}

\newcommand{\aNeutShareMax}{87}
\newcommand{\aNeutShareMin}{43}

\newcommand{\aNeutUpMax}{54}
\newcommand{\aNeutUpMin}{25}
\newcommand{\aNeutvsDirectHCCIDS}{[16.6, 36.0]}
\newcommand{\aNeutvsDirectHCCIQF}{[18.1, 31.2]}
\newcommand{\aNeutvsDirectHCCISol}{[41.2, 67.5]}
\newcommand{\aNeutvsDirectHCDS}{+26.0}
\newcommand{\aNeutvsDirectHCQF}{+24.5}
\newcommand{\aNeutvsDirectHCSol}{+54.3}
\newcommand{\aNeutvsDirectHSCIDS}{[$-$10.3, 1.6]}
\newcommand{\aNeutvsDirectHSCIQF}{[$-$10.9, 5.4]}
\newcommand{\aNeutvsDirectHSCISol}{[$-$14.4, 0.0]}
\newcommand{\aNeutvsDirectHSDS}{$-$2.6}
\newcommand{\aNeutvsDirectHSQF}{$-$2.6}
\newcommand{\aNeutvsDirectHSSol}{$-$5.8}

\newcommand{\aOpenAgree}{93}
\newcommand{\aOpenGameImageDS}{23.3}

\newcommand{\aOpenGameImageQF}{41.2}
\newcommand{\aOpenGameImageSol}{28.4}

\newcommand{\aOpenGapMax}{69}
\newcommand{\aOpenGapMin}{34}

\newcommand{\aOpenN}{296}
\newcommand{\aOpenNamedGameMax}{100}
\newcommand{\aOpenNamedGameMin}{58}
\newcommand{\aOpenNamedMax}{32}
\newcommand{\aOpenNamedMin}{14}

\newcommand{\aOpenRealImageMax}{71}
\newcommand{\aOpenRealImageMin}{51}

\newcommand{\aOpenUnnamedRealMax}{82}
\newcommand{\aOpenUnnamedRealMin}{66}

\newcommand{\aSdvConsMin}{93}

\newcommand{\aSdvGapMax}{38}
\newcommand{\aSdvGapMin}{28}

\newcommand{\aSdvN}{142}

\newcommand{\aSdvRealPooled}{95}

\newcommand{\aSdvRules}{36}

\newcommand{\aTerConfImageQM}{43.5}

\newcommand{\aTerConfToldQM}{88.9}

\newcommand{\aTerGapMax}{45}
\newcommand{\aTerGapMin}{29}

\newcommand{\aTerN}{160}

\newcommand{\aTerRealPooled}{99}

\newcommand{\aTerRecogQM}{99}
\newcommand{\aTerRules}{40}

\newcommand{\aThinkConsImage}{95.2}
\newcommand{\aThinkConsTold}{86.5}
\newcommand{\aThinkDiff}{+22.7}
\newcommand{\aThinkDiffCI}{[13.0, 32.8]}
\newcommand{\aThinkGap}{+44.4}
\newcommand{\aThinkGapCI}{[32.4, 56.1]}
\newcommand{\aThinkImage}{51.6}
\newcommand{\aThinkOffConsImage}{96.5}
\newcommand{\aThinkOffConsTold}{76.0}
\newcommand{\aThinkOffImage}{28.9}

\newcommand{\aThinkOffTold}{80.9}
\newcommand{\aThinkPGap}{+51.0}
\newcommand{\aThinkPGapCI}{[38.5, 63.5]}
\newcommand{\aThinkPImage}{48.0}
\newcommand{\aThinkPTold}{99.0}
\newcommand{\aThinkRealShare}{97}
\newcommand{\aThinkReasked}{21}
\newcommand{\aThinkTold}{96.0}
\newcommand{\aTokImageDS}{1}
\newcommand{\aTokImageSol}{4}

\newcommand{\aTokImageThink}{397}
\newcommand{\aTokToldSol}{4}

\newcommand{\aTokToldThink}{238}

\newcommand{\aToldCoTvsToldHCCIQF}{[9.1, 21.4]}

\newcommand{\aToldCoTvsToldHCQF}{+14.9}

\newcommand{\aToldvsCoTHCCIDS}{[39.2, 63.6]}
\newcommand{\aToldvsCoTHCCIQF}{[11.9, 32.2]}
\newcommand{\aToldvsCoTHCCISol}{[49.3, 76.0]}
\newcommand{\aToldvsCoTHCDS}{+51.5}
\newcommand{\aToldvsCoTHCQF}{+21.8}
\newcommand{\aToldvsCoTHCSol}{+62.5}

\newcommand{\aToldvsGKPHCCIDS}{[19.1, 39.1]}
\newcommand{\aToldvsGKPHCCIQF}{[5.1, 20.7]}
\newcommand{\aToldvsGKPHCCISol}{[39.2, 66.2]}
\newcommand{\aToldvsGKPHCDS}{+29.2}
\newcommand{\aToldvsGKPHCQF}{+13.0}
\newcommand{\aToldvsGKPHCSol}{+52.7}
\newcommand{\aToldvsNeutHCCIDS}{[23.2, 47.7]}
\newcommand{\aToldvsNeutHCCIQF}{[8.3, 22.1]}
\newcommand{\aToldvsNeutHCCISol}{[2.6, 16.0]}
\newcommand{\aToldvsNeutHCDS}{+35.1}
\newcommand{\aToldvsNeutHCQF}{+15.0}
\newcommand{\aToldvsNeutHCSol}{+8.3}
\newcommand{\aTrConcede}{87}

\newcommand{\aTrDescribe}{43}

\newcommand{\aTrMentionImage}{373}

\newcommand{\aTrN}{376}

\newcommand{\aTrPasMentionAccImage}{48}
\newcommand{\aTrPasMentionAccTold}{99}

\newcommand{\aTrPasN}{296}

\newcommand{\aTrRight}{142}
\newcommand{\aTrRightByGame}{141}
\newcommand{\aTrWrong}{154}
\newcommand{\aTrWrongRule}{24}

\newcommand{\aOpenRecMin}{93}
\newcommand{\aOpenRecMax}{100}
\newcommand{\aUnnamedMin}{3}
\newcommand{\aUnnamedMax}{24}
\newcommand{\aUnnamedUpMin}{13}
\newcommand{\aUnnamedUpMax}{49}

\newcommand{\aTextBalMin}{75}
\newcommand{\aTextBalMax}{100}
\newcommand{\aTextLeanNS}{11}
\newcommand{\aLeanSigN}{11}
\newcommand{\aHolmGap}{12}

\newcommand{\aHolmBalHard}{four}

\newcommand{\aWDevItems}{80}
\newcommand{\aWDevRules}{10}
\newcommand{\aWFlipNI}{99}
\newcommand{\aWFlipNQ}{150}

\newcommand{\aWHCGapCII}{[3.7, 19.9]}
\newcommand{\aWHCGapCIQ}{[17.0, 29.8]}
\newcommand{\aWHCGapI}{11.4}
\newcommand{\aWHCGapQ}{23.4}
\newcommand{\aWHCImageI}{13.6}
\newcommand{\aWHCImageQ}{31.9}
\newcommand{\aWHCTextI}{13.8}
\newcommand{\aWHCTextQ}{47.9}
\newcommand{\aWHCToldI}{25.0}
\newcommand{\aWHCToldQ}{55.3}

\newcommand{\aWHPGapI}{12.8}
\newcommand{\aWHPGapQ}{26.7}
\newcommand{\aWHPImageI}{15.2}
\newcommand{\aWHPImageQ}{34.5}
\newcommand{\aWHPTextI}{14.5}
\newcommand{\aWHPTextQ}{56.1}
\newcommand{\aWHPToldI}{28.0}
\newcommand{\aWHPToldQ}{61.1}
\newcommand{\aWHRealI}{98}
\newcommand{\aWHRealQ}{98}

\newcommand{\aWHSGapI}{$-$5.8}
\newcommand{\aWHSGapQ}{$-$23.1}
\newcommand{\aWHSImageI}{78.8}
\newcommand{\aWHSImageQ}{74.0}
\newcommand{\aWHSTextI}{87.5}
\newcommand{\aWHSTextQ}{60.6}
\newcommand{\aWHSToldI}{73.1}
\newcommand{\aWHSToldQ}{51.0}
\newcommand{\aWLayersI}{36}
\newcommand{\aWLayersQ}{32}
\newcommand{\aWMCGapCII}{[$-$0.6, 11.4]}
\newcommand{\aWMCGapCIQ}{[6.7, 15.1]}
\newcommand{\aWMCGapI}{5.6}
\newcommand{\aWMCGapQ}{10.8}
\newcommand{\aWMCImageI}{29.1}
\newcommand{\aWMCImageQ}{39.7}
\newcommand{\aWMCTextI}{30.6}
\newcommand{\aWMCTextQ}{51.1}
\newcommand{\aWMCToldI}{34.7}
\newcommand{\aWMCToldQ}{50.6}

\newcommand{\aWMPGapI}{9.8}
\newcommand{\aWMPGapQ}{15.2}
\newcommand{\aWMPImageI}{27.3}
\newcommand{\aWMPImageQ}{43.0}
\newcommand{\aWMPTextI}{35.5}
\newcommand{\aWMPTextQ}{62.9}
\newcommand{\aWMPToldI}{37.1}
\newcommand{\aWMPToldQ}{58.2}
\newcommand{\aWMRealI}{92}
\newcommand{\aWMRealQ}{95}

\newcommand{\aWMSGapI}{4.9}
\newcommand{\aWMSGapQ}{$-$7.1}
\newcommand{\aWMSImageI}{76.8}
\newcommand{\aWMSImageQ}{80.4}
\newcommand{\aWMSTextI}{83.9}
\newcommand{\aWMSTextQ}{76.3}
\newcommand{\aWMSToldI}{81.7}
\newcommand{\aWMSToldQ}{73.2}

\newcommand{\aWPatchLeftI}{5}
\newcommand{\aWPatchLeftQ}{1}

\newcommand{\aWRealMax}{98}
\newcommand{\aWRealMin}{92}

\newcommand{\aWSplitI}{25}
\newcommand{\aWSplitQ}{21}

\newcommand{\aWStHBalSteerQ}{50.1}

\newcommand{\aWStHBalToldQ}{53.1}

\newcommand{\aWStHCRandomI}{25.0}

\newcommand{\aWStHCSteerI}{24.7}
\newcommand{\aWStHCSteerQ}{54.0}

\newcommand{\aWStHSRandomI}{75.0}

\newcommand{\aWStHSSteerI}{95.2}

\newcommand{\aWSteerLayerI}{19}
\newcommand{\aWSteerLayerQ}{12}
\newcommand{\aWSweepHiI}{35}
\newcommand{\aWSweepHiQ}{30}
\newcommand{\aWSweepLoI}{9}
\newcommand{\aWSweepLoQ}{8}
\newcommand{\tSweepNoneQ}{62.5}
\newcommand{\tSweepBestQ}{69.5}
\newcommand{\tSweepAnswerMoveQ}{3.1}
\newcommand{\tSweepNoneI}{28.1}
\newcommand{\tSweepBestI}{45.3}

\newcommand{\tRecOnlyShiftMax}{1.4}
\newcommand{\tStayPas}{100}
\newcommand{\tStayConf}{92}
\newcommand{\tStayCons}{3}
\newcommand{\tStayBal}{47}
\newcommand{\tCoTNoMentionN}{498}
\newcommand{\tCoTNoMentionRight}{1}
\newcommand{\tPromptedMin}{13.0}
\newcommand{\tPromptedMax}{62.5}
\newcommand{\tWordNativeRules}{23}
\newcommand{\tWordNativeMin}{29}
\newcommand{\tWordNativeMax}{55}
\newcommand{\tWordRealRules}{14}
\newcommand{\tWordRealMin}{45}
\newcommand{\tWordRealMax}{81}

\newcommand{\tWordActMin}{9}
\newcommand{\tWordActMax}{18}
\newcommand{\tChgDynN}{28}
\newcommand{\tChgDynMin}{1}
\newcommand{\tChgDynMax}{13}
\newcommand{\tChgDynUp}{two}
\newcommand{\tChgStatN}{39}
\newcommand{\tChgStatMin}{13}
\newcommand{\tChgStatMax}{25}

\newcommand{\tChgStillN}{14}

\newcommand{\tChgHardDynN}{5}
\newcommand{\tChgHardN}{47}
\newcommand{\tCamMImageMin}{66}
\newcommand{\tCamMImageMax}{70}

\newcommand{\tCamHImageMin}{39}
\newcommand{\tCamHImageMax}{44}

\newcommand{\tThirdTextErrShare}{20}

\newcommand{\tThirdImageErrShare}{4}

\newcommand{\tUnparsedN}{26}
\newcommand{\tUnparsedTotal}{37{,}600}
\newcommand{\tAgreeDSMin}{87}
\newcommand{\tAgreeDSMax}{91}
\newcommand{\tAgreeQFMin}{62}
\newcommand{\tAgreeQFMax}{70}
\newcommand{\tAgreeSolMin}{88}
\newcommand{\tAgreeSolMax}{100}
\newcommand{\tHumMMin}{56}
\newcommand{\tHumMMax}{95}
\newcommand{\tHumMOften}{86}
\newcommand{\tHumMLittle}{74}
\newcommand{\tHumMNever}{67}
\newcommand{\tHumHMin}{55}
\newcommand{\tHumHMax}{97}
\newcommand{\tHumHOften}{92}
\newcommand{\tHumHLittle}{78}
\newcommand{\tHumHNever}{87}
\newcommand{\tHumNOften}{6}
\newcommand{\tHumNLittle}{8}

\newcommand{\tMetaGMMin}{94}
\newcommand{\tMetaGMMax}{99}
\newcommand{\tMetaGapMMin}{6}
\newcommand{\tMetaGapMMax}{19}
\newcommand{\tMetaGHMin}{61}
\newcommand{\tMetaGHMax}{93}
\newcommand{\tMetaGapHMin}{18}
\newcommand{\tMetaGapHMax}{46}
\newcommand{\tMetaGapSig}{every}
\newcommand{\tMetaParsed}{99.9}

\newcommand{\tDepHSGainDP}{$-$14.4}

\newcommand{\tDepMSGainDL}{$-$9.8}

\newcommand{\tDepMMin}{$-$5}
\newcommand{\tDepMMax}{+7}
\newcommand{\tDepHCUp}{five}

\newcommand{\tDepHUpMin}{11}
\newcommand{\tDepHUpMax}{31}

\newcommand{\tDepHLossDS}{7}
\newcommand{\bench}{RuleCraft}
\makeatletter
\newcommand{\tabbox}[2]{\begin{minipage}[t]{#1}\vspace{0pt}\def\@captype{table}#2\end{minipage}}
\newcommand{\figbox}[2]{\begin{minipage}[t]{#1}\vspace{0pt}\def\@captype{figure}#2\end{minipage}}
\makeatother

\title{Multimodal LLMs Know the Game but Assume the Real World}

\hypersetup{pdftitle={Multimodal LLMs Know the Game but Assume the Real World},
  pdfauthor={Jianxin Gao, Bohan Chen, Runze Li, Qi Zhang}}
\author{Jianxin Gao$^{1}$\qquad Bohan Chen$^{2}$\qquad Runze Li$^{3}$\qquad Qi Zhang$^{4}$ \\[1pt]
{\normalfont\small $^{1}$College of Information and Electrical Engineering, China Agricultural University, Beijing, China} \\
{\normalfont\small $^{2}$China Agricultural University--University of Surrey Joint Institute, Beijing, China} \\
{\normalfont\small $^{3}$College of Software, Jilin University, Changchun, China} \\
{\normalfont\small $^{4}$College of Foreign Language, Wuhan Institute of Technology, Wuhan, China} \\[1pt]
{\normalfont\small\texttt{\{jxgao, chenbohan\}@cau.edu.cn, rzli5524@mails.jlu.edu.cn, zqaquarius@gmail.com}} \\[5pt]
{\normalfont\raisebox{-0.22em}{\includegraphics[height=1.15em]{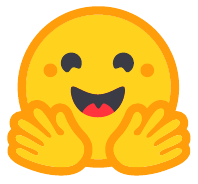}}\,\href{\dataurl}{Dataset and model answers}\qquad
\raisebox{-0.2em}{\includegraphics[height=1.05em]{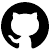}}\,\href{\codeurl}{Code for all experiments}}}

\date{}

\begin{document}
\maketitle

\begin{abstract}
Multimodal large language models can state the rules of a video game, but do they apply them when they only see the
game? We introduce \bench{}, 1{,}240 image questions over 155 Minecraft rules with outcomes verified in the game
engine, each asked in text, on the image with the game named, and on the bare image. Six widely used models know the
rules and recognize the game, yet answer the bare image by real-world physics: naming the game raises their accuracy
on rules where the game departs from reality by \aMGapMin--\aMGapMax{} points on a main set and by
\aHGapMin--\aHGapMax{} on a hard stress-test set. Their errors pick the real-world outcome, the gap is largest where the
game leaves a scene unchanged, and reasoning traces state the game's rule, then answer by reality. On the main set,
naming the game mostly swaps the real-world default for the game's. The pattern holds with reasoning, recurs in two
more games, and persists on questions the models themselves take to be about the game; one neutral sentence asking
for the rules of the world shown recovers much of the gap. Evaluations that name the game measure knowledge of its
rules, not world selection.
\end{abstract}

\section{Introduction}

Multimodal large language models (MLLMs) increasingly reason about worlds other than our own. They detect glitches in
game footage \citep{taesiri2024glitchbench}, play games as agents \citep{gao2026mirrorcraft,zhou2026discovery,zheng2026vmage},
and judge whether generated video is physically plausible \citep{li2025worldmodelbench,gu2026phyworldbench,zhang2026physioneval}.
Many of these worlds run on physics of their own. In Minecraft, a snow block next to lava never melts, and a block whose
support is removed stays in the air. A model that brings real-world physics to such an image gives the wrong answer,
and a judge that does so flags legitimate footage as broken \citep{kim2026gameability}. Games also make this failure
measurable: their rules are fully specified and their engines compute every outcome, so an answer can be checked
against the world a scene belongs to rather than against an annotator's intuition.

\begin{figure}[t]
\centering
\includegraphics[width=\columnwidth]{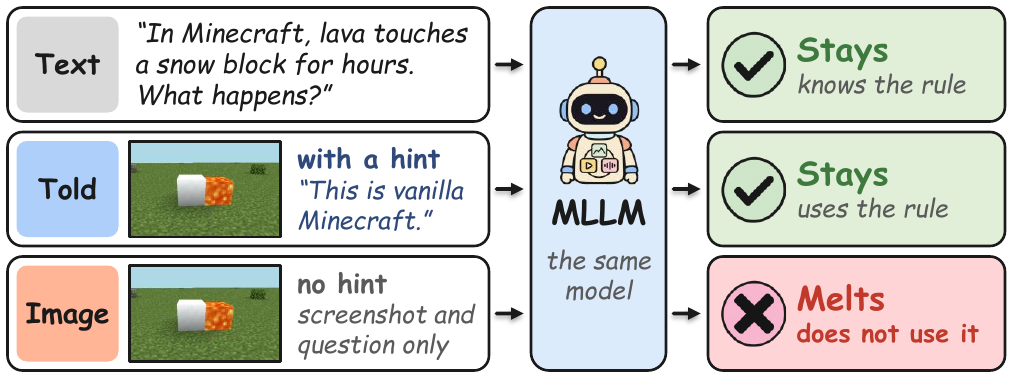}
\caption{One rule in three forms. Qwen3.8-Max picks the game outcome on all eight variants of the rule in text and
when told the world, and the real-world outcome on five of the eight from the bare image.}
\label{fig:teaser}
\end{figure}

Do MLLMs use what they know about a game when they see it? In Figure~\ref{fig:teaser}, Qwen3.8-Max states correctly in
text what happens to a snow block next to lava in Minecraft, and again when told that an image of the scene comes
from Minecraft. Shown the same image without the hint, it mostly predicts that the snow melts. The model holds the
rule and can recognize the game but does not connect the two on its own. Asked which video game an image comes from,
the six models we test name Minecraft on \aOpenRecMin--\aOpenRecMax\% of our images, and the gap holds on the images
each model itself names; what they seldom do unprompted is treat the scene as the game's.

This differs from language priors, where a model overrides what an image shows with what is typical
\citep{vo2025biased,lee2025vlind,sun2026prive}, and from knowledge conflicts, where a context competes with memory
\citep{longpre2021entity,xie2024chameleon}. Here nothing in the input is false or competing: the image renders the game
faithfully, the prompt states no rule, and the model knows the rules of both the game and reality. Each question asks
what happens in the scene shown. A Minecraft scene is a state of the game, so its outcome is the one the game computes,
which is also the answer a game agent or a judge of game footage needs; an answer by reality takes the scene for a
picture of the real world. Choosing between the two is what we call \emph{world selection}: a model makes it from
pixels when it answers the bare image as it answers once the world is named. The models themselves
take our questions to be about the game: asked which world a question concerns, they mostly choose the
game, and on those questions naming it still lifts their answers (Section~\ref{sec:cue}). As in
memory, where a recognized cue need not retrieve what is stored with it \citep{tulving1973encoding}, the image is
recognized, yet the game's rules reach the answer only once its name is given.

We build \bench{} to measure it. Each of its 155 rules pairs a Minecraft scene and an action with the outcome that the
unmodified game produces, verified on a game server. A \emph{main set} of 95 rules estimates how common the gap is; a
\emph{hard set} of 60 rules is a stress test, built around processes that the game does not simulate. Every question
is asked as a text description (\emph{Text}), on the image with the game named (\emph{Told}), and on the bare image
(\emph{Image}). The three conditions separate the steps of an answer: Text tests knowledge of the rules, Told adds
perception of the scene, and Image adds the choice of world, so the difference Told $-$ Image measures what naming
the world changes in the answers. Our contributions are:
\begin{itemize}[leftmargin=*,itemsep=1pt,topsep=2pt]
\item \bench{}, a benchmark of world selection whose game outcomes are verified in the engine. It pairs \emph{conflict}
rules, where the game departs from reality, with \emph{consistent} rules, where the two agree, and carries the design to
Terraria and Stardew Valley (Section~\ref{sec:bench}). We release the benchmark, its engine scripts and every model
answer.
\item Evidence from six widely used models that knowing a world's rules and recognizing it do not select it: naming
the game lifts accuracy on conflict rules for every model on both sets, errors on the bare image pick the real-world
outcome (a real-world default), the models fail on the same rules, and the gap is largest where the game leaves a
scene unchanged, above all when only time or weather acts (a passive gap; Section~\ref{sec:phen}).
\item A lightweight repair and its analysis: one sentence that names no game recovers much of the gap and stating the
game's name recovers most of it, while requested or built-in reasoning does not, as traces state the game's rule and
then answer by reality (concession; Section~\ref{sec:select}).
\end{itemize}

\section{Related Work}

\begin{figure*}[t]
\centering
\includegraphics[width=\textwidth]{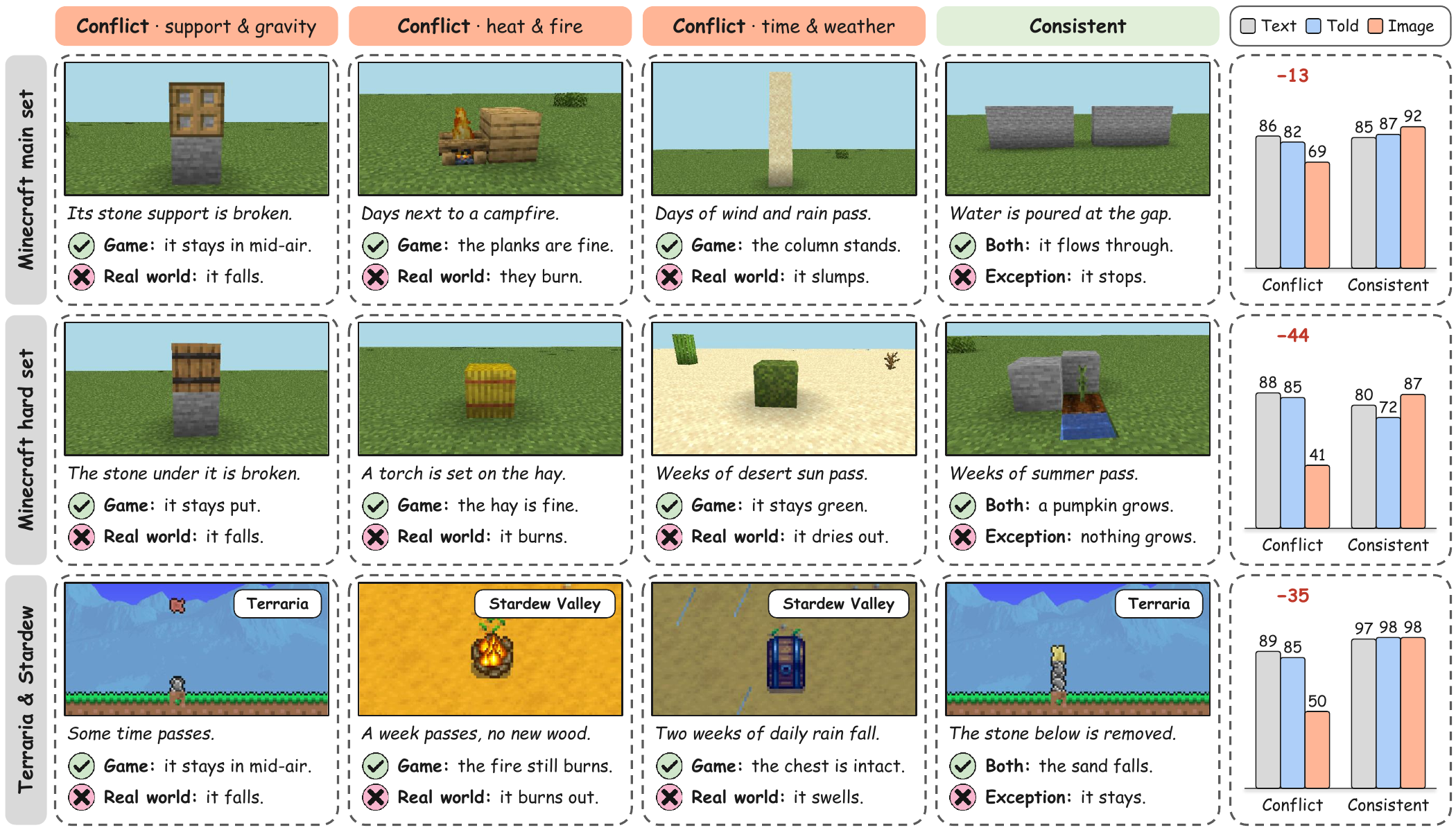}
\caption{Example rules of the two Minecraft sets and of two more games (rows), by kind of rule (columns). In a
conflict rule the game outcome differs from the real-world one; in a consistent rule the two agree, and the wrong option
applies a game exception where none holds. Right: mean accuracy (\%) on all conflict and all consistent questions of the
row in text, when told and on the bare image, and in red the drop from told to bare image on conflict questions (six
models on Minecraft, Section~\ref{sec:phen}; three on the other games, Section~\ref{sec:games}).}
\label{fig:examples}
\end{figure*}

\paragraph{Priors and defaults.} Vision-language models often answer from language priors when an image contradicts
them \citep{lee2025vlind,golovanevsky2025pixels,li2026seeing,zhu2024unraveling,liu2025insight,vo2025biased,sun2026prive},
answer less accurately about an entity shown than named \citep{cohen2025performance}, and can perceive evidence without
acting on it \citep{dai2026seeing}. Our told-image gap extends this to systems of rules, where consistent rules show
that naming a world swaps one default for another. Language models fall back on default conditions in counterfactual
worlds \citep{wu2024reasoning,zhang2026parallel,kirkpatrick2025generics}, keep a default rule when an alternative is
stated \citep{alam2026fixation,wang2026pvfh}, and weigh context against memory
\citep{longpre2021entity,xie2024chameleon,xu2024conflicts}.{} In multi-agent systems,
aggregators count restated evidence as new \citep{gao2026copies}, and an agent's notes help or hurt depending on the
team it joins \citep{gao2026interchangeable}.{} In our setting the game is recognized, no rule is stated and
no context competes: only the image indicates which world applies, much as only context settles the reading of an
ambiguous question \citep{min2020ambigqa}.

\paragraph{Games and physics.} Minecraft is a standard testbed for embodied agents
\citep{baker2022vpt,fan2022minedojo,wang2024voyager,wang2023jarvis1}, whose prompts usually state the game. Game
benchmarks for MLLMs track entities \citep{cohen2026metbench}, hide rule changes \citep{gao2026mirrorcraft}, separate
rule discovery or perception from rule use \citep{zhou2026discovery,jin2026seeing,zheng2026vmage}, ground instructions
in game scenes \citep{qi2026ggbench}, play through a common harness \citep{hu2026lmgame}, supply rules in text, in
conflict or in visual context \citep{peper2026ludobench,kazemi2023boardgameqa,chen2026mmcl}, or teach new physics
\citep{sreekumar2026inphyre,zhang2026ipr}. GlitchBench \citep{taesiri2024glitchbench},
VideoGameQA-Bench \citep{taesiri2025videogameqa}, PhysGame \citep{cao2024physgame}, PhysBench \citep{chow2025physbench}
and QuantiPhy \citep{li2026quantiphy} test glitch detection and physical reasoning, rendered scenes with controlled
layouts test spatial relations \citep{kargin2026sparrta}, and judges of video physics
\citep{li2025worldmodelbench,gu2026phyworldbench,zhang2026physioneval} assume real-world physics, which game footage
violates by design \citep{kim2026gameability}. MLLMs also annotate game footage with rewards
\citep{schmid2026vlmannot} and judge Minecraft world models \citep{savva2026solaris}. In all of these the world is
either given or fixed to reality.

\paragraph{Eliciting and locating knowledge.} Chain of thought \citep{wei2022cot,wu2023cotvl}, generated knowledge
\citep{liu2022gkp}, recitation \citep{sun2023recitation}, step-back prompting \citep{zheng2024stepback} and learned rule
sets \citep{zhou2024walle} elicit knowledge before answering, though stated reasoning need not drive the answer
\citep{turpin2023unfaithful}. These methods assume that the source of the knowledge is settled;
Section~\ref{sec:reason} tests whether they settle it. World selection composes two steps, from the image to its world
and from the world to its rule; language models often answer each step of such a composition but not the whole
\citep{press2023compositionality}, and resolve the second step late in the network or not at all
\citep{yang2024latent,biran2024hopping}. Inside models, the logit lens \citep{nostalgebraist2020logitlens} and its tuned
variant \citep{belrose2023tuned}, activation patching \citep{meng2022locating} and steering vectors
\citep{rimsky2024steering} locate where a prediction forms and which states carry it; Section~\ref{sec:open} applies
them to the choice of world.

\section{The \bench{} Benchmark}
\label{sec:bench}

\bench{} tests world selection in Minecraft for three reasons. The game's departures from real-world physics are many
and well documented, so whether a model knows them can be tested in text. Its engine is deterministic and scriptable,
so every label can be checked by running the scene rather than by annotation. Its look is also distinctive: asked which
game an image shows, models name Minecraft on nearly every image (Section~\ref{sec:cue}), so a failure to apply its
rules is not a failure to recognize it.

\subsection{Rules and Questions}

A rule pairs a scene and an action with the outcome of vanilla Minecraft (Java Edition 1.20.1). In a \emph{conflict}
rule the game outcome differs from the real-world one; in a \emph{consistent} rule the two agree. Each rule is a
parameterized scene with eight variants, which cycle through its target materials (and supports, where it has several)
and four camera positions. The question asks what happens after the action, or at the end of a stated period, and
offers three shuffled options: the game outcome, which the engine verifies, and two that we wrote, the real-world
outcome, the change a real scene of the same objects would undergo, and an outcome that holds in neither world. For a
consistent rule the real-world option is replaced by a \emph{game exception} that does not hold there, so preferring
the unusual answer never pays (Figure~\ref{fig:examples}).

\subsection{Construction and Verification}

The main set holds 95 rules (67 conflict, 28 consistent) over eight families, from gravity and fluids to plants and
heat (Table~\ref{tab:app_families}). The hard set is a stress test of 60 rules written to three principles that make
the bare image harder, so it shows how far the gap can go rather than how common it is. Bare images fail most
when nobody acts on the scene, so in \aHNPas{} of its conflict rules only time or weather acts (\emph{passive} rules)
and in \aHNAct{} something is done to the scene (\emph{action} rules). A real-world option worded as total destruction
is easy to reject, so its real-world options state a mild, realistic change, and its questions avoid game-specific
words. Its \aHNCons{} consistent rules are processes the game does simulate, such as crops ripening or fire spreading,
so that ``nothing changes'' is not a safe answer. The principles come from a separate pool of ten development rules, on
which a milder real-world option lowered GPT-6.1-Sol's bare-image accuracy from 55\% to 19\%. The hard rules were
written before any model answered them, and none was dropped or reworded after (Appendix~\ref{app:bench}); GPT-6.1-Sol's
gain on them (\aHGapSol{} points) lies within the range of the five models that played no part in the pool.

Every variant is built on a vanilla server by a script that performs the action or advances game time and weather, and
the resulting blocks are checked against the label with the server's own commands; all 1{,}240 scenes match. A stated
period, such as a winter or months of drought, runs as a burst of random block updates about 300 times the normal rate,
in the biome the scene is built in. Minecraft has no seasons, evaporation or weathering, so a conflict scene stays as it
was; for these passive rules the check confirms that no process of the game changes the scene over the period. Under
the same accelerated updates, the processes the game does simulate, such as crops ripening in the hard consistent
rules, complete as the game prescribes. Images ($768{\times}480$) are rendered with the game's textures and without its
interface from the same scene specification, which the text condition describes in one or two sentences (Appendix~\ref{app:bench}).
Table~\ref{tab:catalogue} in Appendix~\ref{app:replication} states every rule in one line.

\subsection{Evaluation Setup}

Each question is asked in three conditions. \emph{Image} shows the image and the question. \emph{Told} prefixes
``This image is a screenshot from vanilla Minecraft (Java Edition 1.20.1). Answer according to the rules of vanilla
Minecraft.'' \emph{Text} replaces the image with a description that begins ``In vanilla Minecraft'', so it tests the
rules without perception. In all three the model is asked for a single letter (Appendix~\ref{app:prompts}). The label
is the outcome the game's engine produces for the scene shown; reading the image as a sketch of a real scene is the
real-world default we measure. Image and Told share the question and the options word for word, so the difference between
them measures the effect of naming the world, and a preference for one kind of option, such as a change over no
change, cannot open a gap between them.

\paragraph{Terms.} The \emph{gain} is Told $-$ Image accuracy on conflict rules. The \emph{lean}, conflict minus
consistent accuracy, is negative when a model answers as if the scene were real; \emph{balanced accuracy}, their mean,
rises only if the model tells the two kinds of rule apart better. A \emph{real-world default} shows as a negative lean
and as bare-image errors that pick the real-world option. The \emph{passive gap} is the excess gain on passive over
action rules. \emph{Concession} is a reasoning trace that states the game's rule and then answers by reality. Chance is 33\%. An
item's score is the mean over its runs. Confidence intervals (95\%) come from a two-level bootstrap over rules and items
within rules, and differences are paired per item (Appendix~\ref{app:stats}).

\subsection{Models and Human Reference}

We evaluate six widely used models on both sets and in all three conditions: DeepSeek-V4.1-Flash, Qwen3.8-Max and
-Flash, Doubao-Seed-2.1-Pro and -Lite, and GPT-6.1-Sol (in tables DeepSeek, Qwen-Max, Qwen-Flash, Doubao-Pro,
Doubao-Lite and GPT-6.1-Sol). DeepSeek, Qwen and Doubao answer with reasoning switched off,
GPT-6.1-Sol at its lowest reasoning setting, all at the provider's default temperature; DeepSeek, Qwen3.8-Flash and
GPT-6.1-Sol answer the image conditions three times (Appendix~\ref{app:models}).

As a human reference, \nHumanN{} undergraduates answered the image questions. They took part
voluntarily, knew the purpose of the study, and agreed to the publication of their anonymized answers; the only
personal information collected is their experience with Minecraft, whose reported play time is \nHumanHoursText. A
short questionnaire on that experience preceded the questions, so participants knew which game the images came from,
but they were not asked to apply its rules. Each participant answered 100 of the 155 rules, one variant each, with the
image, action, question and options of the image condition and a Chinese translation of each line
(Appendix~\ref{app:human}). Since they knew the game, their accuracy compares with the told condition of the models.

\section{Models Know the Game but Assume the Real World}
\label{sec:phen}

\begin{table*}[t]
\centering
\footnotesize
\setlength{\tabcolsep}{0pt}
\begin{tabular*}{\textwidth}{@{\extracolsep{\fill}}lcccccccccccccc@{}}
\toprule
& \multicolumn{7}{c}{Main set} & \multicolumn{7}{c}{Hard set} \\
\cmidrule(lr){2-8}\cmidrule(l){9-15}
& \multicolumn{4}{c}{Conflict} & \multicolumn{2}{c}{Consistent} & Bal. & \multicolumn{4}{c}{Conflict} &
\multicolumn{2}{c}{Consistent} & Bal. \\
\cmidrule(lr){2-5}\cmidrule(lr){6-7}\cmidrule(lr){9-12}\cmidrule(lr){13-14}
Model & Text & Told & Image & Told $-$ Image & Told & Image & gain & Text & Told & Image & Told $-$ Image & Told & Image & gain \\
\midrule
DeepSeek & 83.6 & 75.7 & 55.8 & +19.8 {\scriptsize[12.6, 27.8]} & 87.5 & 93.5 & +6.9$^{\dagger}$ & 85.2 & 80.9 & 28.9 & +52.0 {\scriptsize[40.0, 63.4]} & 76.0 & 96.5 & +15.8$^{\dagger}$ \\
Qwen-Max & 89.2 & 85.3 & 75.0 & +10.3 {\scriptsize[4.7, 16.4]} & 83.5 & 92.0 & +0.9\phantom{$^{\dagger}$} & 88.8 & 91.5 & 57.7 & +33.8 {\scriptsize[22.3, 45.5]} & 58.7 & 74.0 & +9.2$^{\dagger}$ \\
Qwen-Flash & 74.4 & 67.1 & 58.9 & +8.2 {\scriptsize[2.2, 13.9]} & 79.0 & 84.7 & +1.3\phantom{$^{\dagger}$} & 78.5 & 71.4 & 36.5 & +34.8 {\scriptsize[26.1, 43.7]} & 70.2 & 82.7 & +11.2$^{\dagger}$ \\
Doubao-Pro & 84.5 & 85.6 & 72.4 & +13.2 {\scriptsize[6.5, 20.1]} & 85.7 & 91.5 & +3.7\phantom{$^{\dagger}$} & 89.1 & 86.2 & 44.7 & +41.5 {\scriptsize[30.1, 52.9]} & 62.5 & 85.6 & +9.2$^{\dagger}$ \\
Doubao-Lite & 82.3 & 76.7 & 61.2 & +15.5 {\scriptsize[9.0, 22.8]} & 84.4 & 93.3 & +3.3\phantom{$^{\dagger}$} & 88.8 & 83.8 & 31.1 & +52.7 {\scriptsize[41.0, 64.1]} & 68.3 & 81.7 & +19.6$^{\dagger}$ \\
GPT-6.1-Sol & 100.0 & 99.9 & 88.4 & +11.6 {\scriptsize[5.5, 18.7]} & 100.0 & 99.6 & +6.0$^{\dagger}$ & 98.7 & 98.2 & 47.0 & +51.2 {\scriptsize[38.1, 64.5]} & 96.2 & 99.0 & +24.2$^{\dagger}$ \\
\addlinespace[2pt]
Humans$^{*}$ & \textendash & \textendash & \nHumanConf & \textendash & \textendash & \nHumanCons & \textendash & \textendash & \textendash & \nHumanHConf & \textendash & \textendash & \nHumanHCons & \textendash \\
\bottomrule
\end{tabular*}
\caption{Accuracy (\%) by condition; chance is 33\%. Conflict: rules where the game departs from reality; Told $-$ Image is the gain from
naming the game (95\% CI). Consistent: rules where game and reality agree. Bal.\ gain: Told $-$ Image in balanced
accuracy, the mean of the two; $^{\dagger}$interval excludes zero. $^{*}$Participants saw the bare images knowing that they
came from Minecraft, without being asked to apply its rules.}
\label{tab:phen}
\end{table*}

\subsection{Knowledge Versus Use}

Table~\ref{tab:phen} compares the three conditions. For each of the \aHNModels{} models, Told
$-$ Image is positive with an interval above zero on both sets: \aMGapMin{} to \aMGapMax{} points on the main set,
\aHGapMin{} to \aHGapMax{} on the hard set, and \aHPGapMin{} to \aHPGapMax{} on its passive rules. All \aHolmGap{} gains
remain significant after Holm correction (Table~\ref{tab:app_holm}), and so do the main-set gains of \aMechGapSig{}
models when the bootstrap resamples the \nRevMechN{} game mechanisms behind the conflict rules, such as blocks that need
no support, instead of the rules (Table~\ref{tab:mech}).

Accuracy in text reflects knowledge of the rules rather than a habit of choosing the game's ``nothing changes''. In
text the lean is close to zero, with an interval that includes zero for \aTextLeanNS{} of the 12 pairs of model and
set, and balanced accuracy in text is \aTextBalMin--\aTextBalMax\% (Table~\ref{tab:app_default}). A habit would score
differently. Every hard passive conflict rule leaves the scene as it was, so always choosing the option in which
nothing changes scores \tStayPas\% on them and \tStayConf\% on all hard conflict rules, but \tStayCons\% on the hard
consistent rules: a balanced accuracy of \tStayBal\%, below every model in every condition on the hard set.

This knowledge goes unused on the bare image. GPT-6.1-Sol answers every hard passive rule correctly in text and when
told, yet only \aHPImageSol\% of the hard passive questions from the bare image. Of the conflict questions a model
answers correctly in every text and every told run, it misses \aMUnusedMin--\aMUnusedMax\% on the bare image in the
main set and \aHUnusedMin--\aHUnusedMax\% in the hard set. The larger model of each family keeps the gap: on the hard
set Qwen3.8-Max and Doubao-Seed-2.1-Pro gain \aHGapQM{} and \aHGapDP{} points.

\begin{table}[t]
\centering
\small
\setlength{\tabcolsep}{1pt}
\begin{tabular*}{\columnwidth}{@{\extracolsep{\fill}}lccccc@{}}
\toprule
& \multicolumn{2}{c}{Image} & \multicolumn{2}{c}{Told} & Gain in Game \\
\cmidrule(lr){2-3} \cmidrule(lr){4-5} \cmidrule(l){6-6}
Model & Game & Real & Game & Real & Told $-$ Image \\
\midrule
DeepSeek & 23.3 & 58.4 & 57.1 & 36.8 & +33.8 {\scriptsize[20.6, 46.3]} \\
GPT-6.1-Sol & 28.4 & 70.6 & 97.6 & 0.3 & +69.3 {\scriptsize[57.1, 81.1]} \\
Qwen-Flash & 41.2 & 51.0 & 86.1 & 10.5 & +44.9 {\scriptsize[32.8, 56.8]} \\
\bottomrule
\end{tabular*}
\caption{Open answers on the hard set's passive conflict rules: share (\%) of answers that describe the game outcome
(Game) and the real-world outcome (Real); the rest match neither. Last column: gain in the game outcome from naming the
game (95\% CI).}
\label{tab:app_open}
\end{table}

\begin{table*}[t]
\centering
\small
\begin{tabular*}{\textwidth}{@{\extracolsep{\fill}}>{\raggedright\arraybackslash}p{0.2\textwidth}>{\raggedright\arraybackslash}p{0.37\textwidth}>{\raggedright\arraybackslash}p{0.37\textwidth}@{}}
\toprule
Scene and time & Bare image & Told \\
\midrule
An anvil, many years of rain & ``The anvil becomes rusty and corroded after years of exposure to rain, leaving its
surface rough and reddish-brown.'' & ``The anvil remains unchanged. Rain and the passage of time do not rust or damage
anvils in vanilla Minecraft.'' \\
A campfire, a week without new wood & ``The campfire burns through its remaining wood and goes out. By the end of the
week, it is cold, with no flames or glowing embers.'' & ``The campfire remains lit after the week. In vanilla
Minecraft, campfires burn indefinitely without needing more wood.'' \\
\bottomrule
\end{tabular*}
\caption{Open answers of GPT-6.1-Sol to the first variant of two hard passive rules, without answer options.}
\label{tab:app_open_examples}
\end{table*}

The pattern also holds without answer options. Asked to describe the outcome of hard passive questions in their own
words, three models describe the real-world outcome in \aOpenRealImageMin--\aOpenRealImageMax\% of their answers from
the bare image, such as an anvil that rusts or a campfire that burns out, and the game outcome \aOpenGapMin{} to
\aOpenGapMax{} points more often when told (Tables~\ref{tab:app_open} and~\ref{tab:app_open_examples}).

\begin{table}[t]
\centering
\small
\setlength{\tabcolsep}{1pt}
\begin{tabular*}{\columnwidth}{@{\extracolsep{\fill}}llcc@{}}
\toprule
Scene and time & Real world & Image & Told \\
\midrule
Moss block, weeks of sun & dries out & 0 & 97 \\
Lily pads, a frosty winter & wither & 4 & 97 \\
Brick wall, centuries & crumbles & 8 & 100 \\
Tree, months of drought & wilts & 12 & 99 \\
Snow block, warm rain & melts & 12 & 91 \\
Flower pot, weeks of sun & wilts & 15 & 99 \\
Dirt path, heavy rain & turns muddy & 19 & 97 \\
\bottomrule
\end{tabular*}
\caption{Hard rules that the \aBlindModels{} models know and do not apply: their mean accuracy (\%) is at least
90 in text and when told and below 20 on the bare image. Real
world: the outcome the models choose instead; in Minecraft nothing changes; \aBlindReal\% of the models' image errors pick the real-world outcome.}
\label{tab:blind}
\end{table}

The models fail on the same rules (Figure~\ref{fig:perrule}). Table~\ref{tab:blind} lists the \aBlindN{} rules that
the six models answer correctly in text and when told and almost never from the image. Minecraft leaves each of these
scenes unchanged; from the image, the models predict the change that the scene would undergo in reality.

When the bare image misleads a model, the error seldom picks the outcome that holds in neither world.
\aMRealPooled\% of the errors on the main set and \aHRealPooled\% on the hard set pick the real-world outcome of the
block and action shown: the errors predict what reality would do with the scene. The third option does draw errors
when the world is named: in text it takes \tThirdTextErrShare\% of the errors on hard conflict rules, against
\tThirdImageErrShare\% on the bare image. The human participants, who knew
where the images came from, answered \nHumanConf\% {\small\nHumanConfCI} of the main set's and \nHumanHConf\%
{\small\nHumanHConfCI} of the hard set's conflict questions correctly, close to the models when told, and
\nHumanFallback\% and \nHumanHFallback\% of their errors also picked the real-world outcome.

\subsection{The Real-World Default}
\label{sec:default}

\begin{figure}[t]
\centering
\includegraphics[width=\columnwidth]{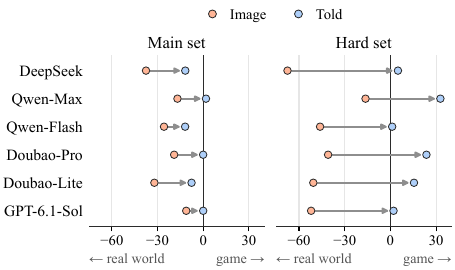}
\caption{Lean of the models, conflict minus consistent accuracy, on the bare image and when told; the arrow shows
the move when the game is named. Negative: the model answers as if the scene were real.}
\label{fig:default}
\end{figure}

A preference for the most plausible option would leave consistent rules untouched when the game is named. In \bench{}
they move with the conflict rules: the errors follow whichever world the model takes as the default. On the bare image
that default is reality. The lean of the models is \aMBiasImageMin{} to \aMBiasImageMax{} points on the main set and
\aHBiasImageMin{} to \aHBiasImageMax{} on the hard set (Figure~\ref{fig:default}; Table~\ref{tab:app_default}), below
zero with an interval that excludes zero in \aLeanSigN{} of the 12 pairs. Their high accuracy on consistent rules
(\aHConsImageMin--\aHConsImageMax\% on the hard set) comes from reality and the game agreeing there. On hard conflict
rules DeepSeek and Doubao-Seed-2.1-Lite even fall below the chance level of 33\% from the bare image (\aHImageDS{} and
\aHImageDL\%), which a default produces and a guess does not.

Naming the game swaps the default. The lean rises for every model on both sets and turns positive for every model on
the hard set, where accuracy on consistent rules falls: DeepSeek drops from \aHConsImageDS{} to \aHConsToldDS\%,
Doubao-Seed-2.1-Pro from \aHConsImageDP{} to \aHConsToldDP\%, and \aHSRulesBelow{} of the \aHSRulesN{} hard
consistent rules are answered worse when told than from the bare image. The new errors mirror the old ones. Told errors
on consistent rules pick the game exception (\aMOverPooled\% on the main set and \aHOverPooled\% on the hard set), just
as errors on conflict rules from the bare image pick reality (Table~\ref{tab:direction}).

Naming the game also adds discrimination, which a swap of defaults alone cannot do (Bal.\ gain in
Table~\ref{tab:phen}). On the hard set balanced accuracy rises for all \aHNModels{} models (\aHBalDiffMin{} to
\aHBalDiffMax{} points, \aHolmBalHard{} of them significant after Holm correction). On the main set it rises
significantly for DeepSeek and GPT-6.1-Sol, and for the other four models the gain from being told is mostly the swap
of defaults. Naming the world replaces reality as the default and, where the rules are hardest, lets models apply rules
they know.

\subsection{Passive and Action Rules}
\label{sec:where}

\begin{figure}[t]
\centering
\includegraphics[width=\columnwidth]{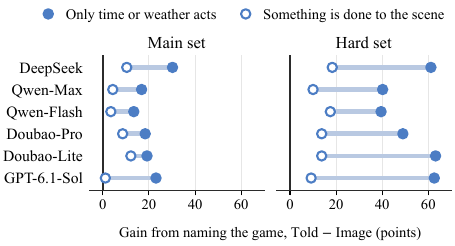}
\caption{Gain from naming the game on conflict rules where only time or weather acts (filled) and where something is
done to the scene (open), per model.}
\label{fig:where}
\end{figure}

Splitting conflict rules by what happens in the scene shows where the default matters most (Table~\ref{tab:app_types}).
When something is done to the scene, such as a block broken, water poured or a fire lit, the models gain
\aMAGapMin{} to \aMAGapMax{} points from naming the game on the main set and \aHAGapMin{} to \aHAGapMax{} on the hard
set. When only time or weather acts, they gain \aMPGapMin{} to \aMPGapMax{} and \aHPGapMin{} to \aHPGapMax{} points
(Figure~\ref{fig:where}), and on the hard set the difference is significant for all six models, also after Holm correction. We call this excess the \emph{passive gap}; every rule of Table~\ref{tab:blind} shows it.

In these scenes the image shows a cause whose effect the game does not compute: weeks of desert sun on a moss block or
a full cauldron, warm rain on a snow block, a campfire burning a week without wood. From the image the models let the
moss brown, the water evaporate, the snow slump and the fire go out, although in text they answer
\aHPTextMin--\aHPTextMax\% of these rules correctly; nothing in such a scene marks the game's indifference to time.

The passive gap is not a matter of wording. \tWordRealRules{} of the hard passive rules name weather or seasons that
the game lacks, such as frost or a hot sun, and \tWordNativeRules{} name only the passing of time, sun or rain, which
the game has. On the latter the models gain \tWordNativeMin{} to \tWordNativeMax{} points from the name, more than on
action rules (\tWordActMin{} to \tWordActMax) for every model; real-world weather adds to the gap
(\tWordRealMin{} to \tWordRealMax{} points) but does not create it.

The gap follows what the game leaves unchanged. On the \tChgDynN{} conflict rules of the main set whose game outcome is
itself a change, such as tilled soil turning back into dirt or leaves vanishing once their tree is cut, naming the game
adds \tChgDynMin{} to \tChgDynMax{} points, significantly for \tChgDynUp{} models; on the \tChgStatN{} where the game
leaves the scene as it was, it adds \tChgStatMin{} to \tChgStatMax{} for every model. In the hard set,
\tChgHardDynN{} of the \tChgHardN{} conflict rules are of the first kind. The told answers do not follow a habit of
choosing that nothing happens: on the \tChgStillN{} changing rules whose real-world option is that nothing happens, no
model answers significantly worse when told.

\subsection{Two More Games}
\label{sec:games}

\begin{table}[t]
\centering
\footnotesize
\setlength{\tabcolsep}{0pt}
\begin{tabular*}{\columnwidth}{@{\extracolsep{\fill}}lcccccc@{}}
\toprule
& \multicolumn{4}{c}{Conflict} & \multicolumn{2}{c}{Consistent} \\
\cmidrule(lr){2-5} \cmidrule(l){6-7}
Model & Text & Told & Image & Told $-$ Image & Image & Told \\
\midrule
\multicolumn{7}{@{}l}{\textbf{Terraria}} \\
DeepSeek & 88.9 & 98.1 & 69.4 & +28.7 {\scriptsize[14.8, 43.5]} & 98.1 & 98.1 \\
Qwen-Max & 100.0 & 88.9 & 43.5 & +45.4 {\scriptsize[29.6, 60.2]} & 94.2 & 100.0 \\
Doubao-Pro & 86.1 & 88.9 & 51.9 & +37.0 {\scriptsize[22.2, 52.8]} & 98.1 & 100.0 \\
\addlinespace[2pt]
\multicolumn{7}{@{}l}{\textbf{Stardew Valley}} \\
DeepSeek & 75.5 & 67.3 & 39.8 & +27.6 {\scriptsize[7.3, 47.9]} & 100.0 & 100.0 \\
Qwen-Max & 94.9 & 88.8 & 58.2 & +30.6 {\scriptsize[15.2, 46.9]} & 100.0 & 97.7 \\
Doubao-Pro & 85.7 & 77.6 & 39.8 & +37.8 {\scriptsize[18.0, 57.3]} & 100.0 & 93.2 \\
\bottomrule
\end{tabular*}
\caption{Two more games: accuracy (\%) on conflict and consistent rules, with the gain from naming the game (95\% CI).}
\label{tab:games}
\end{table}

To test whether the pattern is specific to Minecraft, we built questions for two games with rules of their own:
\aTerN{} questions over \aTerRules{} rules of Terraria, a 2D sandbox whose labels are checked on a game server, and
\aSdvN{} questions over \aSdvRules{} rules of Stardew Valley, a farming game whose labels are checked by running each
scene in the game (Appendix~\ref{app:games}). Each rule has up to four variants (Figure~\ref{fig:examples}, bottom
row), and DeepSeek, Qwen3.8-Max and Doubao-Seed-2.1-Pro answer in the three conditions (Table~\ref{tab:games}). Naming
the game lifts them by \aTerGapMin{} to \aTerGapMax{} points on Terraria's conflict rules and by \aSdvGapMin{} to
\aSdvGapMax{} on Stardew Valley's, every interval above zero, and \aTerRealPooled\% and \aSdvRealPooled\% of the errors
on the bare image pick the real-world outcome. Recognition again
does not settle the world: Qwen3.8-Max names Terraria on \aTerRecogQM\% of its images and answers \aTerConfImageQM\% of
its conflict questions from them, against \aTerConfToldQM\% when told. Accuracy on consistent rules stays at or above
\aSdvConsMin\% in both games.

\section{What Selects the World}
\label{sec:select}

\subsection{Cues to the World}
\label{sec:cue}

\begin{table}[t]
\centering
\small
\setlength{\tabcolsep}{0pt}
\begin{tabular*}{\columnwidth}{@{\extracolsep{\fill}}l cc cc cc@{}}
\toprule
& \multicolumn{2}{c}{DeepSeek} & \multicolumn{2}{c}{GPT-6.1-Sol} & \multicolumn{2}{c}{Qwen-Flash} \\
\cmidrule(lr){2-3} \cmidrule(lr){4-5} \cmidrule(l){6-7}
Prompt & Pas. & Cons. & Pas. & Cons. & Pas. & Cons. \\
\midrule
Direct answer & 23.9 & \textbf{96.5} & 37.4 & 99.0 & 35.4 & \textbf{82.7} \\
\addlinespace[3pt]
Chain of thought & 33.6 & 85.9 & 37.5 & \textbf{100.0} & 53.0 & 78.5 \\
Gen.\ knowledge & 55.9 & 84.3 & 47.3 & 97.1 & 61.9 & 70.5 \\
\addlinespace[3pt]
\emph{The world shown} & 49.9 & 93.9 & 91.7 & 93.3 & 59.9 & 80.1 \\
\emph{A video game} & \textbf{78.6} & 83.3 & \textbf{99.7} & 93.3 & \textbf{81.8} & 61.9 \\
\addlinespace[3pt]
\textcolor{gray}{Game named} & \textcolor{gray}{56.2} & \textcolor{gray}{92.9} & \textcolor{gray}{93.6} & \textcolor{gray}{99.0} & \textcolor{gray}{59.2} & \textcolor{gray}{79.5} \\
\textcolor{gray}{+ vanilla, version} & \textcolor{gray}{77.8} & \textcolor{gray}{79.8} & \textcolor{gray}{99.7} & \textcolor{gray}{97.1} & \textcolor{gray}{74.5} & \textcolor{gray}{70.5} \\
\textcolor{gray}{Told} & \textcolor{gray}{85.0} & \textcolor{gray}{76.0} & \textcolor{gray}{100.0} & \textcolor{gray}{96.2} & \textcolor{gray}{74.9} & \textcolor{gray}{70.2} \\
\textcolor{gray}{Told + CoT} & \textcolor{gray}{86.4} & \textcolor{gray}{72.4} & \textcolor{gray}{100.0} & \textcolor{gray}{97.1} & \textcolor{gray}{89.8} & \textcolor{gray}{71.2} \\
\bottomrule
\end{tabular*}
\caption{Accuracy (\%) on the hard set by prompt: no cue, reasoning on request, and cues to the world, ordered by how plainly
they point away from reality. Pas.: conflict rules where only time or weather acts; Cons.: consistent rules. Grey rows name the game for
the model: ``This image is a screenshot from Minecraft.'', then ``\dots{} from vanilla Minecraft (Java Edition
1.20.1).'', and Told adds ``Answer according to the rules of vanilla Minecraft.'' (CoT: chain of thought). Bold:
best in a column among the prompts that do not name the game.}
\label{tab:method}
\end{table}

If the missing piece is world selection, a cue to the world should close the gap. We test this on the hard set with
DeepSeek, GPT-6.1-Sol and Qwen3.8-Flash, which span the range of rule knowledge among the models
(Table~\ref{tab:method}). Two prompts ask for reasoning and give no cue: chain of thought \citep{wei2022cot} and
generated knowledge \citep{liu2022gkp}. Two prompts give cues, ordered by how plainly they point away from reality,
and neither names a game. \emph{The world shown} only raises the question: ``Answer according to the rules that govern
the world shown in the image.'' \emph{A video game} also rules out the default: ``Answer according to the rules of the
video game shown in the image, not according to real-world physics.'' Four references name the game for the model: its
name alone, the name with ``vanilla'' and the version, Told, which adds the request to answer by the game's rules and
is answered with one letter as in Section~\ref{sec:phen}, and Told + chain of thought.

The neutral sentence alone moves every model. On passive conflict rules it adds \aNeutUpMin{} to \aNeutUpMax{} points
and recovers \aNeutShareMin--\aNeutShareMax\% of the gain from being told, the most for GPT-6.1-Sol, which knows every
one of these rules. Ruling out real-world physics without naming any game brings all three models close to the told
prompt, so most of what the told prompt contributes is a world other than reality. That cue also excludes the
realistic option by instruction, and it costs consistent rules (Qwen3.8-Flash falls from \aMethHSDirectQF{} to
\aMethHSFixedGameQF\%); the neutral sentence excludes nothing. Requests to reason do least: chain of thought and
generated knowledge stay \tPromptedMin{} to \tPromptedMax{} points below Told, and the main set keeps the same order
(Table~\ref{tab:app_method}).

Two further conditions bracket the bare image. Described in text with no world named, the fourth cell of a design that
crosses text or image with the world named or not, the hard passive rules are
answered correctly in \aUnnamedMin--\aUnnamedMax\% of cases by the six models; without a cue, the scene is taken as
real. The bare image lifts this by \aUnnamedUpMin{} to \aUnnamedUpMax{} points for every model, so the pixels do act as
a cue, the weakest of all. At the other end, stating the world is enough without any instruction. The game's
name alone recovers \aNameOnlyShareMin--\aNameOnlyShareMax\% of the gain from being told, and with ``vanilla'' and the
version \aNameVerShareMin--\aNameVerShareMax\%; the request to answer by the game's rules adds at most \aInstrMax{}
points. What the told prompt supplies is the world, not an instruction.

A deployment context is a cue for most models but not all. After a system message for a tool that analyses gameplay
recordings, which implies game footage without naming a game, the bare image gains \tDepHUpMin{} to \tDepHUpMax{}
points on hard conflict rules for \tDepHCUp{} of the six models, while DeepSeek loses \tDepHLossDS; on the main set the
changes run from \tDepMMin{} to \tDepMMax{} points, and every model stays below its told accuracy. Implying the game
does not replace naming it.

What the cues supply is not recognition. Asked only ``Which video game is this screenshot from?'', with no list to
choose from, the six models name Minecraft on \aOpenRecMin--\aOpenRecMax\% of the 1{,}240 images, and restricted to the
images a model itself names as Minecraft, its bare-image accuracy on conflict rules changes by at most
\tRecOnlyShiftMax{} points on either set. Recognized, the world is still seldom selected: open answers from the bare
image mention Minecraft in only \aOpenNamedMin--\aOpenNamedMax\% of cases, and \aOpenNamedGameMin--\aOpenNamedGameMax\%
of those describe the game outcome, against \aOpenUnnamedRealMin--\aOpenUnnamedRealMax\% real-world outcomes among the
rest.

Nor do the models read the question as one about reality. Shown the bare image, action, question and options and
asked which world the question is about instead of answering it, they choose the video game over the real world for
\tMetaGMMin--\tMetaGMMax\% of the main set's items and \tMetaGHMin--\tMetaGHMax\% of the hard set's. On the conflict
items that a model itself reads as about the game, naming the game still adds \tMetaGapMMin{} to \tMetaGapMMax{}
points on the main set and \tMetaGapHMin{} to \tMetaGapHMax{} on the hard set, \tMetaGapSig{} interval above zero.
Taking the question to concern the game does not make a model answer it by the game.

Each cue brings its world's default with it. The neutral sentence keeps consistent rules within noise for all three
models (at most \aNeutCostMax{} points). Firmer cues cost consistent rules where knowledge is incomplete: told,
DeepSeek falls from \aMethHSDirectDS{} to \aMethHSToldDS\%, while GPT-6.1-Sol, which answers every hard consistent rule
in text, keeps \aMethHSToldSol\%. Balanced accuracy still rises with every cue, from
\aBalHardDirectQF--\aBalHardDirectSol\% with a direct answer to \aBalHardToldQF--\aBalHardToldSol\% when told.

\subsection{The Role of Reasoning}
\label{sec:reason}

\begin{table}[t]
\centering
\footnotesize
\setlength{\tabcolsep}{1pt}
\begin{tabular*}{\columnwidth}{@{\extracolsep{\fill}}lccccc@{}}
\toprule
& \multicolumn{2}{c}{Reasoning} & & & \\
\cmidrule(lr){2-3}
Model, setting & (\%) & tokens & Image & Told & Told $-$ Image \\
\midrule
DeepSeek, off & 0 & 1 & 28.9 & 80.9 & +52.0 {\scriptsize[40.0, 63.4]} \\
DeepSeek, on & 100 & 397 & 51.6 & 96.0 & +44.4 {\scriptsize[32.4, 56.1]} \\
GPT-6.1-Sol, low & 35 & 4 & 47.0 & 98.2 & +51.2 {\scriptsize[38.1, 64.5]} \\
GPT-6.1-Sol, high & 76 & 26 & 55.3 & 98.4 & +43.1 {\scriptsize[29.8, 56.6]} \\
\bottomrule
\end{tabular*}
\caption{Reasoning and world selection on hard conflict rules. Reasoning: share (\%) of bare-image answers that
reason before the answer letter, and median output tokens of a bare-image answer (the letter alone is \aTokImageDS{} token for
DeepSeek and \aTokImageSol{} for GPT-6.1-Sol). Whatever the setting, naming the game still adds accuracy (95\% CI).}
\label{tab:reasoning}
\end{table}

More reasoning does not close the gap, and the reasoning traces show why. Requested in the prompt, a chain of thought
helps only when it names the game. On hard passive rules it moves DeepSeek from \aMethHCDirectDS{} to \aMethHCCoTDS\%
and leaves GPT-6.1-Sol close to its direct answer (Table~\ref{tab:method}). Of the \tCoTNoMentionN{} chains of the three
models that never mention Minecraft, \tCoTNoMentionRight{} reaches the game outcome; the rest reason about sunlight,
dye and frost and arrive at the real-world one. GPT-6.1-Sol mentions the game in \aCoTMentionSol\% of its chains and is
then always right. Once the world is fixed, step-by-step reasoning does help the model with the weakest rule knowledge:
Told + chain of thought lifts Qwen3.8-Flash by \aToldCoTvsToldHCQF{} points \aToldCoTvsToldHCCIQF{} over Told.

Reasoning built into the model does not replace the cue either (Table~\ref{tab:reasoning}). Raised from its lowest
setting to a high one, GPT-6.1-Sol gains \aHighDiff{} points \aHighDiffCI{} on hard conflict rules. With its thinking
phase on, DeepSeek rises from \aThinkOffImage{} to \aThinkImage\%, and being told still adds \aThinkGap{} points. The
thinking costs a median of \aTokImageThink{} output tokens per answer, while told GPT-6.1-Sol answers \aHToldSol\% of
these questions with a median of \aTokToldSol. In all \aNConfigs{} combinations of model and reasoning
setting that we ran on the hard set, naming the game adds accuracy with an interval above zero.

\begin{table*}[t]
\centering
\small
\begin{tabular*}{\textwidth}{@{\extracolsep{\fill}}>{\raggedright\arraybackslash}p{0.25\textwidth}r>{\raggedright\arraybackslash}p{0.63\textwidth}@{}}
\toprule
Kind of trace & $n$ & Excerpt (DeepSeek, thinking on, bare image) \\
\midrule
States the game rule and answers by it & \aTrRightByGame & ``In Minecraft, anvils do not weather/rust/change appearance
over time due to rain.'' (an anvil in years of rain; correct) \\
States the game rule, then answers by the real world & \aTrConcede & ``But in Minecraft leaves don't change due no rain;
however question logical real world.'' (a tree in a desert for months; wrong) \\
Names the game only to describe the image & \aTrDescribe & ``Image shows Minecraft honey block on grass. [\dots] Honey
is hygroscopic, absorbs water, can dissolve/run off.'' (honey in heavy rain; wrong) \\
Holds a wrong game rule & \aTrWrongRule & ``In Minecraft, a dry sponge absorbs water when it comes into contact with
it.'' (a dry sponge in rain; wrong) \\
\bottomrule
\end{tabular*}
\caption{The four kinds of reasoning on the hard set's passive conflict questions, with their counts among the
\aTrPasN{} traces (the first among the \aTrRight{} correct answers, one more of which was sorted as a concession; the
others among the \aTrWrong{} wrong ones) and one excerpt each, quoted as written.}
\label{tab:traces}
\end{table*}

DeepSeek's reasoning traces show where the answer turns. With thinking on, it mentions Minecraft in \aTrMentionImage{}
of its \aTrN{} traces on conflict rules without being asked. On passive rules it is still right in only
\aTrPasMentionAccImage\% of the traces that mention the game, against \aTrPasMentionAccTold\% when told. A judge model
that reads the thinking but not the final reply sorted each of its \aTrPasN{} passive traces by how it uses the game
(Table~\ref{tab:traces}; Appendix~\ref{app:judges}). Of the \aTrRight{} traces behind a correct answer, \aTrRightByGame{} work out the game
outcome from the game's rule. Of its \aTrWrong{} wrong answers, \aTrConcede{} state that Minecraft would leave the scene
unchanged and then answer by the real world, a pattern we call \emph{concession}; \aTrDescribe{} reason by the real
world alone and use the name only to describe the image, and \aTrWrongRule{} misremember the rule. One trace reads:
``In Minecraft, dirt doesn't change state due to rain, but in a general reasoning context, dirt gets muddy.'' The model
has recognized the game and recalled its rule, and it then decides that the question concerns the real world. Whatever
else drives the answer \citep{turpin2023unfaithful}, the rule was retrieved and set aside.

\section{Inside Two Open Models}
\label{sec:open}

\begin{figure}[t]
\centering
\includegraphics[width=\columnwidth]{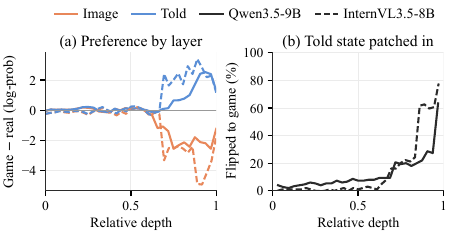}
\caption{Conflict items answered correctly only when told. (a) Game minus real-world log-probability at every
layer. (b) Items flipped to the game outcome by patching in the told answer-position state at one layer.}
\label{fig:whitebox}
\end{figure}

As a descriptive note, we run the two conditions on Qwen3.5-9B and InternVL3.5-8B with thinking off
(Appendix~\ref{app:whitebox}). Both name Minecraft on every image, gain \aWHCGapQ{} and \aWHCGapI{} points from the
name on hard conflict rules, and make real-world errors (\aWRealMin--\aWRealMax\%; Table~\ref{tab:wb_behav}), but they
know far fewer rules than the six models (Qwen3.5-9B answers \aWHCTextQ\% of hard conflict questions in text, and
InternVL3.5-8B stays below chance), so we only trace the \aWFlipNQ{} and \aWFlipNI{} conflict items that each answers
correctly only when told. Read through the output head \citep{nostalgebraist2020logitlens,belrose2023tuned}, the two
runs do not differ in their preference between the game and the real-world option until layer \aWSplitQ{} of
\aWLayersQ{} and \aWSplitI{} of \aWLayersI{} (Figure~\ref{fig:whitebox}a), and patching in the told answer-position
state \citep{meng2022locating} flips most items only with \aWPatchLeftQ{} to \aWPatchLeftI{} layers left
(Figure~\ref{fig:whitebox}b). The mean told-minus-bare-image state of a middle layer \citep{rimsky2024steering} acts
as a bias toward the game rather than a repair: it lifts Qwen3.5-9B's hard conflict accuracy to \aWStHCSteerQ\% and
lowers consistent rules, so balanced accuracy does not rise (\aWStHBalSteerQ\%; told: \aWStHBalToldQ\%;
Table~\ref{tab:wb_steer}); for InternVL3.5-8B it leaves hard conflict accuracy where a random direction does
(\aWStHCSteerI\% vs \aWStHCRandomI\%) and raises only consistent rules (\aWStHSSteerI\% vs \aWStHSRandomI\%).

\section{Discussion}
\label{sec:discussion}

\paragraph{Naming the world changes the task.} Benchmarks and judges that state the game in the prompt measure the
told condition, \aHPGapMin{} to \aHPGapMax{} points above the bare image on hard passive rules: whether a model knows
a world's rules, not whether it selects that world. Pipelines that score raw game footage need the second: saying that
the input is gameplay closes part of the gap for most models, and naming the game closes it.

\paragraph{Departures need consistent controls.} Accuracy on conflict rules alone rewards any default toward the game:
told, DeepSeek gains \aHGapDS{} points on hard conflict rules and loses \aHConsDropDS{} on consistent ones. Benchmarks of
departures from reality should report balanced accuracy over both kinds of rule.

\paragraph{A model can supply the name itself.} Asked first which world an image shows and then to answer in it, a
model reaches its told accuracy where it names the game (Appendix~\ref{app:arxiv_self}), so a pipeline can make the
selection an explicit step of its own.

\paragraph{Future work.} Natural extensions are a human bare-image condition with participants not told where the
images come from, real gameplay frames with the game's interface, and open models that know the rules well enough to
trace the selection inside them.

\section{Conclusion}

The six MLLMs we test know a great deal about the physics of a game and use little of it when the game is only shown.
They recognize the game, answer its rules in text and take the question to be about it, yet on a bare image they take
reality as the default. Selecting the world is a step of its own between recognizing a scene and answering about it:
a sentence that puts another world on the table moves the default, and naming the game swaps it, most of all for
processes the game leaves static, in all three games. Evaluations of models in other worlds should measure this step
rather than supply it in the prompt.

\clearpage
\appendix
\setcounter{topnumber}{3}\setcounter{dbltopnumber}{3}\setcounter{totalnumber}{5}
\renewcommand{\topfraction}{0.9}\renewcommand{\dbltopfraction}{0.9}\renewcommand{\textfraction}{0.08}
\renewcommand{\floatpagefraction}{0.75}\renewcommand{\dblfloatpagefraction}{0.75}
\makeatletter
\setlength{\@fptop}{0pt}\setlength{\@fpsep}{8pt plus 4pt}\setlength{\@dblfptop}{0pt}\setlength{\@dblfpsep}{8pt plus 4pt}
\makeatother
\onecolumn
\section{Acknowledgments}
\label{app:ack}

\raggedbottom
\definecolor{ackhead}{HTML}{7A4E2D}
\definecolor{ackA}{HTML}{F8F2E6}
\definecolor{ackB}{HTML}{FDFBF6}
\definecolor{acktext}{HTML}{2E2A25}
\newlength\ackname\setlength\ackname{80pt}
\newlength\ackuni\setlength\ackuni{138pt}
\newlength\acklead\setlength\acklead{24pt}
\newlength\ackmajor
\newlength\acktmp
\newcounter{ackrow}
\newcounter{ackicon}
\newcommand\ackfit[2]{\settowidth\acktmp{#2}%
  \ifdim\acktmp>#1\relax\resizebox{#1}{!}{#2}\else\makebox[#1][l]{#2}\fi}
\newcommand\ackband[2]{\par\noindent{\setlength\fboxsep{0pt}\colorbox{#1}{\makebox[\linewidth][l]{%
  \rule[-0.4em]{0pt}{1.5em}#2}}}\par\nointerlineskip}
\newcommand\ackperson[3]{\stepcounter{ackrow}\stepcounter{ackicon}%
  \ifnum\value{ackicon}>6 \setcounter{ackicon}{1}\fi
  \edef\ackcol{\ifodd\value{ackrow}ackA\else ackB\fi}%
  \ackband{\ackcol}{\hspace{7pt}\raisebox{-0.14em}{\includegraphics[height=9pt]{art/block\arabic{ackicon}}}\hspace{8pt}%
    \ackfit{\ackname}{\textbf{#1}}\ackfit{\ackuni}{#2}\ackfit{\ackmajor}{Major in #3}}}

We thank everyone who took part in our human study. Your predictions about what happens next in our blocky little
worlds gave us a human baseline for comparing model answers.

\vspace{4pt}

{\footnotesize\color{acktext}\setlength\parskip{0pt}%
\setlength\ackmajor{\dimexpr\linewidth-\acklead-\ackname-\ackuni-6pt\relax}%
\noindent\includegraphics[width=\linewidth]{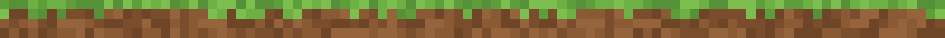}\par\nointerlineskip
\ackperson{Hang Liu}{Jilin University}{Software Engineering}
\ackperson{Yuze Li}{China Agricultural University}{Agricultural Mechanization and Automation}
\ackperson{Hanmeng Liang}{China Agricultural University}{Electronic Information}
\ackperson{Ziyi Wu}{China Agricultural University}{Mechanical Engineering}
\ackperson{Tianle Cheng}{China Agricultural University}{Computer Science}
\ackperson{Ziyue Jiang}{East China Normal University}{Microelectronics Science and Engineering}
\ackperson{Bohan Sun}{China Agricultural University}{Electronic Information Engineering}
\ackperson{Liangwei Ren}{China Agricultural University}{Electrical Engineering and Automation}
\ackperson{Yue He}{China Agricultural University}{Electronic Information Engineering}
\ackperson{Haotian Shi}{China Agricultural University}{Electrical Engineering and Automation}
\ackperson{Xiaoxiao Han}{Yanching Institute of Technology}{Taxation}
\ackperson{Yihang Wang}{China Agricultural University}{Communication Engineering}
\ackperson{Jiarun Li}{Tiangong University}{Aircraft Manufacturing Engineering}
\ackperson{Yujia Huo}{Taiyuan University of Technology}{Software Engineering}
\ackperson{Qinze Li}{Jilin University}{Pharmacy}
\noindent\includegraphics[width=\linewidth]{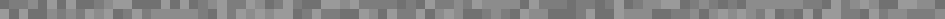}\par}

\definecolor{barimg}{HTML}{E07B24}
\definecolor{bartold}{HTML}{4A7FC1}
\definecolor{barhum}{HTML}{5E9E3A}
\definecolor{barempty}{HTML}{E4DED2}
\newcount\exfilled
\newcount\exsegment
\newcommand\exbar[3]{\noindent\makebox[32pt][l]{#1}%
  \exfilled=#2\relax\advance\exfilled by 5\relax\divide\exfilled by 10\relax
  \exsegment=0\relax
  \loop\ifnum\exsegment<10
    \ifnum\exsegment<\exfilled\textcolor{#3}{\rule{5pt}{3.5pt}}%
    \else\textcolor{barempty}{\rule{5pt}{3.5pt}}\fi
    \hspace{0.8pt}\advance\exsegment by 1\relax
  \repeat
  \hfill#2\%\par}
\newcommand\exopt[3]{\par\hangindent=15pt\hangafter=1\makebox[8pt][l]{#1}\makebox[7pt][l]{#2}#3\par}
\newlength\cellw
\newlength\exmax
\newlength\extmp
\newcounter{excol}
\newsavebox\exbi
\newsavebox\exbii
\newsavebox\exbiii
\newcommand\excell[7]{\stepcounter{excol}%
  \expandafter\gdef\csname eximg\roman{excol}\endcsname{#1}%
  \expandafter\gdef\csname exlab\roman{excol}\endcsname{#2}%
  \expandafter\gdef\csname exaccuracy\roman{excol}\endcsname{%
    \exbar{Image}{#5}{barimg}\exbar{Told}{#6}{bartold}\exbar{Human}{#7}{barhum}}%
  \expandafter\sbox\csname exb\roman{excol}\endcsname{\parbox[t]{\dimexpr\cellw-6pt\relax}{\scriptsize
    \color{acktext}\raggedright\setlength\parskip{0pt}\setlength\baselineskip{8.5pt}%
    \emph{#3}\par\vspace{1pt}#4}}}
\newsavebox\eximgbox
\newsavebox\exlabbox
\newcommand\exshow[1]{{\setlength\fboxsep{3pt}\colorbox{ackA}{\parbox[t]{\dimexpr\cellw-6pt\relax}{\scriptsize
  \color{acktext}\sbox\eximgbox{\includegraphics[width=\linewidth]{\csname eximg#1\endcsname}}%
  \sbox\exlabbox{\setlength\fboxsep{1.5pt}\colorbox{ackhead}{\textcolor{white}{\bfseries\csname exlab#1\endcsname}}}%
  \leavevmode\rlap{\usebox\eximgbox}\hspace{3pt}\raisebox{\dimexpr\ht\eximgbox-\ht\exlabbox-3pt\relax}[0pt][0pt]{\usebox\exlabbox}%
  \par\vspace{2pt}%
  \vbox to\exmax{\hbox{\usebox{\csname exb#1\endcsname}}\vss}\vspace{2pt}%
  \setlength\baselineskip{8.5pt}\csname exaccuracy#1\endcsname}}}}
\newcommand\exfit[1]{\setlength\extmp{\dimexpr\ht#1+\dp#1\relax}\ifdim\extmp>\exmax\setlength\exmax{\extmp}\fi}
\newcommand\exrowend{\setlength\exmax{0pt}\exfit\exbi\exfit\exbii\exfit\exbiii
  \par\noindent\exshow{i}\hfill\exshow{ii}\hfill\exshow{iii}\par\vspace{4pt}\setcounter{excol}{0}}

\vspace{7pt}
\noindent{\large\bfseries Example Questions from the Human Study}\par\vspace{2pt}
{\footnotesize\noindent\emph{What happens immediately after this action?} \caseok{} marks the correct choice.
Bars show accuracy across variants of each rule for the six models on the image alone (Image) or told the game (Told),
and for participants (Human; Appendix~\ref{app:human}).\par}

\vspace{4pt}
{\setlength\parskip{0pt}\setlength\parindent{0pt}\setlength\cellw{\dimexpr(\linewidth-12pt)/3\relax}%
\excell{figures/human/F14_v0}{Main set, conflict rule}{A bucket of lava is poured directly onto the water from above.}{\exopt{}{A}{The water flashes into steam, and the lava crumbles into dark hardened rock.}\exopt{\caseok}{B}{The water block just under the lava instantly turns into smooth stone.}\exopt{}{C}{The lava floats on top of the water and stays molten.}}{53}{59}{100}%
\excell{figures/human/F4_v0}{Main set, conflict rule}{The stone block under the cake is broken and removed.}{\exopt{}{A}{The cake falls down and lands on the ground.}\exopt{}{B}{The cake stays exactly where it was, floating in mid-air.}\exopt{\caseok}{C}{The cake disappears, and nothing at all is left behind.}}{37}{47}{38}%
\excell{figures/human/K12_v0}{Main set, consistent rule}{The oak planks block that the lantern hangs from is broken and removed.}{\exopt{}{A}{The lantern stays exactly where it was, floating in mid-air.}\exopt{\caseok}{B}{The lantern drops down; it does not stay in the air.}\exopt{}{C}{The lantern goes out but stays where it was.}}{96}{92}{58}\exrowend
\excell{figures/human/Y20_v0}{Hard set, passive rule}{Several days of very heavy rain fall on the pool.}{\exopt{\caseok}{A}{The pool is exactly as full as it was, and no water spills over.}\exopt{}{B}{The rain has washed the pool's water away, leaving it empty.}\exopt{}{C}{The pool has overflowed, and water has spread onto the grass around it.}}{32}{95}{100}%
\excell{figures/human/Y43_v0}{Hard set, action rule}{A block of ice is let go just above the middle of the pool.}{\exopt{\caseok}{A}{The ice stays exactly where it was, hanging in the air above the water.}\exopt{}{B}{The ice drops into the pool and floats on the water.}\exopt{}{C}{The ice melts into the pool at once.}}{2}{23}{92}%
\excell{figures/human/Y56_v0}{Hard set, consistent rule}{A few hours pass with the ice block sitting right next to the burning campfire.}{\exopt{}{A}{The ice block puts the campfire out.}\exopt{}{B}{The ice block stays completely unchanged; the campfire gives off no heat.}\exopt{\caseok}{C}{The ice block melts into water.}}{84}{49}{33}\exrowend}
\twocolumn

\section{Replication Details}
\label{app:replication}

\subsection{Rules, Variants and Verification}
\label{app:bench}

\paragraph{Scenes.} Each rule is a parameterized scene: a target block on a flat grass or sand field, optional supporting
or neighbouring blocks, and an action such as breaking a block, pouring a fluid, lighting a fire or letting time pass.
Its eight variants cycle through the rule's target materials (such as four kinds of wool or wood) and supports, and
variant $v$ is seen from camera position $v \bmod 4$ with a small random turn; option order is shuffled with a fixed
seed per variant. Bare-image accuracy on conflict rules differs little across camera positions (six-model means of
\tCamMImageMin--\tCamMImageMax\% on the main set and \tCamHImageMin--\tCamHImageMax\% on the hard set). The text
condition describes the scene of each variant in one or two sentences filled in from the same specification.

\paragraph{Verification.} A script builds each variant on a vanilla Java Edition 1.20.1 server with \texttt{setblock}
commands and drives a bot for actions that need a player. Time is advanced by raising the random-tick speed from 3 to
1{,}000 for 10 to 45 seconds, in the biome the scene is built in, rain by the weather command, and fire and falling
blocks run in real time. The expected blocks are then tested with \texttt{execute if block}; a scene passes only if
every check matches the label.

\paragraph{Construction.} A first round of rules served to set up the conditions and is not reported; the prompts of Section~\ref{sec:cue} were written on the first batch of the main set, and its last two
batches were written with predictions recorded before any model answered them. The principles of the hard set were
learned on a development pool of ten rules, outside both test sets, with GPT-6.1-Sol in the loop: there, rewording the
real-world option from total destruction to a mild, realistic change lowered its bare-image accuracy from 55\% to 19\%.
The 60 hard rules were then written to these principles before any model answered them; none was chosen, dropped or
reworded by a model's answers, and only the engine check could correct a label.

\subsection{Terraria and Stardew Valley}
\label{app:games}

\paragraph{Terraria.} The \aTerRules{} rules (14 passive conflict, 13 action conflict, 13 consistent) concern blocks
that the game keeps in the air without support, blocks that do fall, and blocks left to time. Each of the \aTerN{}
questions is built on a dedicated server (version 1.4.4) by a server plugin that places the tiles of the scene,
advances time and weather or removes the named block, and reads back the state of every tile; a question is kept only
if the read state matches its label. Images ($768{\times}480$) are rendered from the same scene with tile sprites
taken from the game files.

\paragraph{Stardew Valley.} The \aSdvRules{} rules (13 passive conflict, 12 action conflict, 11 consistent) concern
crops, trees, fires, chests, rocks and weeds on the farm. A game mod (version 1.6) clears a patch of the farm, sets the
date and weather, builds the scene, takes a screenshot, performs the steps of the question through the game's own tools
or its sleep command, and writes the state of every tile before and after. The \aSdvN{} questions are those whose
recorded state matches the label, and their images are crops of the game's own screenshots. Both games use the
conditions and the three kinds of option of Minecraft.

\subsection{Models and Settings}
\label{app:models}

\begin{table}[t]
\centering
\small
\setlength{\tabcolsep}{3pt}
\begin{tabular*}{\columnwidth}{@{\extracolsep{\fill}}lll@{}}
\toprule
Model & Setting & Runs \\
\midrule
DeepSeek-V4.1-Flash & thinking off & 3 / 3 \\
Qwen3.8-Max & thinking off & 1 / 1 \\
Qwen3.8-Flash & thinking off & 3$^\ddagger$ / 3 \\
Doubao-Seed-2.1-Pro & thinking off & 1 / 1 \\
Doubao-Seed-2.1-Lite & thinking off & 1 / 1 \\
GPT-6.1-Sol & lowest effort & 3$^\dagger$ / 3$^\dagger$ \\
\bottomrule
\end{tabular*}
\caption{Models and settings. Runs of the image conditions on the main / hard set. $^\dagger$Conflict rules; one run on
consistent rules and in the text condition. $^\ddagger$On the main set's batches 2--4; one run on batch 1. Tables
shorten the names to DeepSeek, Qwen-Max, Qwen-Flash, Doubao-Pro, Doubao-Lite and GPT-6.1-Sol.}
\label{tab:app_models}
\end{table}

All models are called through OpenAI-compatible APIs between 4 and 17 November 2026, with the image sent as a PNG
before the text, at the provider's default temperature (Table~\ref{tab:app_models}). Answers are parsed by
a fixed rule (an explicit ``Answer: X'' first, then a bold or bare letter); unparseable replies, \tUnparsedN{} of the
\tUnparsedTotal{} answers of the three conditions, count as wrong. Repeated runs give the same letter on
\tAgreeDSMin--\tAgreeDSMax\% of the items for DeepSeek, \tAgreeQFMin--\tAgreeQFMax\% for Qwen3.8-Flash and
\tAgreeSolMin--\tAgreeSolMax\% for GPT-6.1-Sol. Every reply is cached with its request, so all analyses replay offline.

\paragraph{Reasoning settings.} Table~\ref{tab:reasoning} counts the output tokens that the API bills for an answer,
reasoning included; an answer carries reasoning when the provider reports reasoning tokens for it. For the comparison
of settings, GPT-6.1-Sol answered the hard set's image and told conditions once more at the setting \emph{high}, and
DeepSeek once more with its thinking phase switched on, with an output limit of 8{,}000 tokens; the \aThinkReasked{}
answers that used up this limit were asked again with a limit of 32{,}000.

\subsection{Prompts}
\label{app:prompts}

\begin{figure*}[t]
\centering
\footnotesize
\fbox{\begin{minipage}{0.97\textwidth}
\textbf{Image.} \texttt{Look at the image.\textbackslash n Action: \{action\}\textbackslash n What happens immediately
after this action? (If the action describes time passing, answer for the end of that period.)\textbackslash n A.
\{option\}\textbackslash n B. \{option\}\textbackslash n C. \{option\}\textbackslash n Reply with a single letter.}\\[3pt]
\textbf{Told.} The image prompt, preceded by \texttt{This image is a screenshot from vanilla Minecraft (Java Edition
1.20.1). Answer according to the rules of vanilla Minecraft.}\\[3pt]
\textbf{Text.} No image; \texttt{In vanilla Minecraft (Java Edition 1.20.1): \{scene description\}} followed by the
action, the question, the options and the tail of the image prompt.\\[3pt]
\textbf{Text, no world named.} The text prompt without \texttt{In vanilla Minecraft (Java Edition 1.20.1):}.\\[3pt]
\textbf{Game named.} The image prompt, preceded by \texttt{This image is a screenshot from Minecraft.}; with version:
\texttt{This image is a screenshot from vanilla Minecraft (Java Edition 1.20.1).}\\[3pt]
\textbf{Chain of thought.} The image prompt with the tail \texttt{Think step by step, then end your reply with 'Answer:
<letter>'.}\\[3pt]
\textbf{Generated knowledge.} Call 1: \texttt{Before answering, list 3-5 facts or rules that are relevant for predicting
the answer to the question below. Do not answer the question yet.} with the image and the question. Call 2: the image
prompt, preceded by \texttt{Relevant knowledge:} and the listed facts.\\[3pt]
\textbf{The world shown.} The image prompt, preceded by \texttt{Answer according to the rules that govern the world
shown in the image.}\\[3pt]
\textbf{A video game.} The image prompt, preceded by \texttt{Answer according to the rules of the video game shown in
the image, not according to real-world physics.}\\[3pt]
\textbf{Which world.} The image prompt with its last line replaced by \texttt{Do not answer the question. Instead, say
which world the question is about:\textbackslash n G. what happens in the video game that the image comes
from\textbackslash n R. what would happen to such a scene in the real world\textbackslash n Reply with a single letter,
G or R.}; G and R swap places on every second item.\\[3pt]
\textbf{Gameplay tool.} The image prompt after the system message \texttt{You are the assistant of a tool that analyses
gameplay recordings. Users send a frame from a gameplay video and ask what happens next in the scene.}\\[3pt]
\textbf{Open answer.} The image or told prompt without the options, ending \texttt{Describe what happens to \{object\}
in one or two sentences.}\\[3pt]
\textbf{Recognition.} \texttt{Which video game is this screenshot from? Reply with the name of the game only.}
\end{minipage}}
\caption{Prompts. \texttt{\textbackslash n} marks a line break; braces mark the parts filled per question. Terraria and
Stardew Valley use the same prompts with their own game and version.}
\label{fig:app_prompts}
\end{figure*}

Figure~\ref{fig:app_prompts} gives every prompt; all conditions share the question block.

\subsection{Statistics and Judges}
\label{app:stats}

\begin{table*}[t]
\centering
\footnotesize
\begin{tabular}{@{}lll@{}}
\toprule
Analysis & Models and runs & Questions \\
\midrule
Image, Told and Text (Section~\ref{sec:phen}) & all six, runs as in Table~\ref{tab:app_models} & 760 / 480 \\
\midrule
\emph{The world shown} & DeepSeek 3, GPT-6.1-Sol 3, Qwen-Flash 3 & 480 \\
\emph{A video game} & DeepSeek 3, GPT-6.1-Sol 1, Qwen-Flash 3 & 480 \\
Chain of thought & DeepSeek 3, GPT-6.1-Sol 1, Qwen-Flash 3 & 480 \\
Generated knowledge & DeepSeek 3, GPT-6.1-Sol 1, Qwen-Flash 3 & 480 \\
Game named; + version & DeepSeek, Qwen-Flash 3; GPT-6.1-Sol 3 (1 cons.) & 480 \\
Text, no world named & DeepSeek 3, other models 1 & 480 \\
\midrule
Open answers & DeepSeek, GPT-6.1-Sol, Qwen-Flash 1 each & 296 \\
Reasoning traces & DeepSeek with thinking on 1 & 480 \\
Terraria / Stardew Valley & DeepSeek, Qwen-Max, Doubao-Pro 1 each & 160 / 142 \\
\midrule
Which world; gameplay tool & all six, 1 each & 760 / 480 \\
\midrule
Open models (Section~\ref{sec:open}) & Qwen3.5-9B, InternVL3.5-8B 1 each & 760 / 480; 80 dev \\
Human reference & 15 participants, 100 rules each & one variant per rule \\
\bottomrule
\end{tabular}
\caption{What every analysis rests on. Questions: main / hard set, or the items of the analysis; the cues run on the
hard set. An item's score is the mean over its runs.}
\label{tab:app_counts}
\end{table*}

\paragraph{Bootstrap.} Each item's score is the mean over its runs. Accuracies and confidence intervals come from
2{,}000 bootstrap samples that resample rules with replacement and, within each drawn rule, its items; intervals are
percentile intervals. Differences between conditions or prompts are paired per item and resampled the same way. The
gain in balanced accuracy resamples conflict and consistent rules separately; the passive-minus-action difference
resamples the two rule types independently. Clustering by mechanism resamples the \nRevMechN{} groups of main-set
rules that share a game mechanism instead of rules.

\paragraph{Multiple comparisons.} The main comparisons are repeated as two-sided bootstrap tests and corrected with the
Holm procedure within each family of tests (Table~\ref{tab:app_holm}). Elsewhere, ``significant'' means that the 95\%
interval excludes zero. Table~\ref{tab:app_counts} lists the models, questions and runs behind every analysis; counts in
the text are summed over runs.

\paragraph{Judges.}
\label{app:judges}
A trace or chain of thought mentions the game if it contains the word ``Minecraft''. For the \aTrPasN{} passive items
of the reasoning traces, a text-only judge call (DeepSeek, temperature 0) reads each bare-image thinking phase, without
the final reply, together with the game outcome and the real-world outcome of its item and sorts it into one of the four kinds of Table~\ref{tab:traces}.
For the open answers, on the \aOpenN{} items of the hard passive conflict rules, a text-only judge call
(DeepSeek, temperature 0) reads the action, the three outcomes of the item and the answer, and says which outcome the
answer matches, or that it matches none or leaves the outcome open. The judge sees neither the condition nor the answer
key, and a second judge, Qwen3.8-Flash, gives the same label to \aOpenAgree\% of the answers.

\subsection{Open Models}
\label{app:whitebox}

\paragraph{Setup.} Qwen3.5-9B and InternVL3.5-8B run in bfloat16 on one GPU with the prompts and images of the six
models, the image before the text, through each model's chat template with thinking switched off. One forward pass per
question gives the answer: the option letter with the highest probability at the first answer position, which holds a
median of more than 99\% of the probability on the option letters. That pass also stores the hidden state of the answer
position at every layer. The recognition question is answered by greedy decoding.

\paragraph{Logit lens and patching.} Each stored state goes through the model's final norm and output head, and we take
the log-probability of the game option minus that of the real-world option, averaged over the conflict items of both
test sets that a model answers correctly when told and wrongly from the bare image (4 and 26 items go the other way).
For patching, the bare-image run of each such item is repeated once per layer with the answer-position output of that
layer replaced by the stored told state.

\paragraph{Steering.} The \aWDevRules{} development rules belong to neither test set. The
direction at a layer is the mean, over their \aWDevItems{} items, of the told state minus the bare-image state at the
answer position. On these rules we add it, times 1, 2 or 3, at every position or at the answer position only, at every
second layer from layer \aWSweepLoQ{} to \aWSweepHiQ{} (Qwen3.5-9B) and \aWSweepLoI{} to \aWSweepHiI{}
(InternVL3.5-8B), and keep the setting with the highest balanced accuracy: times 3 at every position of layer
\aWSteerLayerQ{} of Qwen3.5-9B (\tSweepBestQ\%, against \tSweepNoneQ\% without the direction) and of layer
\aWSteerLayerI{} of InternVL3.5-8B (\tSweepBestI\%, against \tSweepNoneI\%). At the answer position only, no setting
moves balanced accuracy by more than \tSweepAnswerMoveQ{} points.
The random direction is drawn once (seed 0) and scaled to the norm of the direction.

\subsection{Human Study}
\label{app:human}

Each participant was invited individually and took part voluntarily, without payment; all knew the purpose of the
study and agreed to the publication of their anonymized answers. The only personal data collected, before the
questions, was their play time and frequency of play in Minecraft. A bilingual phone page shows the image,
action, question and three options exactly as in the image condition, each line followed by a Chinese translation, and gives each participant one variant each of 100 of the 155 rules, in random order, drawn
within rule types (43 main conflict, 18 main consistent, 24 hard passive, 7 hard action, 8 hard consistent). Per
participant, conflict accuracy spans \tHumMMin--\tHumMMax\% (main) and \tHumHMin--\tHumHMax\% (hard); the
\tHumNOften{} who play often average \tHumMOften{} and \tHumHOften\%, the \tHumNLittle{} who play a little
\tHumMLittle{} and \tHumHLittle\%, the one who never played \tHumMNever{} and \tHumHNever\%.

\begin{table*}[p]
\centering
\scriptsize
\setlength{\tabcolsep}{3pt}\renewcommand{\arraystretch}{0.86}
\begin{tabular*}{\textwidth}{@{\extracolsep{\fill}}llll>{\raggedright\arraybackslash}p{0.68\textwidth}rrr@{}}
\toprule
Rule & Set & Family & Type & Rule in the game & T & Td & I \\
\midrule
F1 & M & Grav. & Pas. & A tall column of sand stays standing and does not slump into a heap. & 88 & 87 & 21 \\
F5 & M & Plant & Pas. & A cactus stops growing at three blocks tall. & 85 & 97 & 74 \\
F6 & M & Time & Pas. & A fully grown tree never gets any bigger. & 100 & 100 & 74 \\
F7 & M & Time & Pas. & Cobblestone standing in water never grows moss by itself. & 99 & 92 & 40 \\
F10 & M & Heat & Pas. & A soul torch right next to an ice block does not melt it. & 98 & 92 & 78 \\
F11 & M & Heat & Pas. & A block that burns up leaves no ash behind. & 100 & 96 & 98 \\
F13 & M & Heat & Pas. & Rain does not put out a torch. & 96 & 100 & 90 \\
H1 & M & Heat & Pas. & A campfire does not set fire to a wooden block next to it. & 100 & 89 & 64 \\
H2 & M & Heat & Pas. & A glass block touching lava is not affected. & 100 & 99 & 99 \\
H3 & M & Heat & Pas. & A full snow block touching lava does not melt. & 67 & 43 & 23 \\
H4 & M & Heat & Pas. & Ice sitting on a magma block does not melt. & 67 & 57 & 31 \\
H5 & M & Heat & Pas. & A wooden block on top of a magma block does not catch fire. & 92 & 86 & 53 \\
H6 & M & Heat & Pas. & Thin snow lying on a magma block does not melt. & 43 & 37 & 15 \\
H7 & M & Heat & Pas. & An ordinary torch placed next to TNT does not set it off. & 100 & 94 & 84 \\
H8 & M & Heat & Pas. & A cobweb next to a fire does not burn. & 98 & 61 & 40 \\
H9 & M & Time & Pas. & A flower left alone never wilts. & 100 & 100 & 75 \\
H10 & M & Time & Pas. & A placed cake never goes bad. & 100 & 100 & 100 \\
M1 & M & Heat & Pas. & An ice block right next to a glowstone block melts. & 71 & 67 & 65 \\
M2 & M & Plant & Pas. & Dry farmland with nothing planted on it turns back into dirt. & 100 & 95 & 97 \\
M3 & M & Fluid & Pas. & Lava from one source on flat ground in the Overworld spreads about three blocks and stops. & 98 & 84 & 67 \\
M4 & M & Heat & Pas. & A fire on a grass block does not spread across the grass. & 49 & 78 & 65 \\
M5 & M & Heat & Pas. & A crafting table next to lava does not catch fire. & 49 & 55 & 28 \\
M8 & M & Heat & Pas. & A wet sponge left in a desert in the Overworld stays wet. & 93 & 56 & 5 \\
N5 & M & Heat & Pas. & Lava left alone never cools down or hardens by itself. & 100 & 100 & 92 \\
N6 & M & Heat & Pas. & A fire on top of netherrack burns forever. & 100 & 98 & 99 \\
N12 & M & Fluid & Pas. & Water from one source block on flat ground spreads about seven blocks and then stops. & 100 & 78 & 69 \\
N15 & M & Heat & Pas. & Water never evaporates, even in a hot desert. & 100 & 90 & 6 \\
N16 & M & Heat & Pas. & A full snow block does not melt next to a lit torch. & 94 & 88 & 62 \\
N17 & M & Heat & Pas. & Packed ice and blue ice never melt, even right next to a torch. & 100 & 94 & 93 \\
N18 & M & Heat & Pas. & Sunlight alone never melts a thin layer of snow. & 42 & 35 & 17 \\
N24 & M & React. & Pas. & Iron blocks never rust. & 100 & 100 & 76 \\
N25 & M & React. & Pas. & Wooden planks never rot. & 100 & 100 & 100 \\
F2 & M & Grav. & Act. & A chain stays in place when the block it hangs from is removed. & 65 & 63 & 55 \\
F3 & M & Att. & Act. & A trapdoor stays in place when the block it is attached to is removed. & 78 & 82 & 67 \\
F4 & M & Att. & Act. & A cake vanishes without dropping anything when the block under it is removed. & 51 & 47 & 37 \\
F8 & M & Fluid & Act. & Flowing water does not put out or wash away a lantern. & 100 & 100 & 99 \\
F9 & M & React. & Act. & TNT that explodes in water does not damage the blocks around it. & 99 & 97 & 87 \\
F12 & M & React. & Act. & A piston cannot push an obsidian block. & 71 & 63 & 56 \\
F14 & M & React. & Act. & Lava poured onto water from above turns the water block under it into stone. & 65 & 59 & 53 \\
F15 & M & Att. & Act. & All the sugar cane above breaks into items when its lowest piece is broken. & 95 & 99 & 99 \\
M6 & M & React. & Act. & A stone block broken by hand without a pickaxe drops nothing. & 72 & 67 & 45 \\
M7 & M & Fluid & Act. & Flowing water does not wash away a sand block. & 100 & 99 & 99 \\
N2 & M & Grav. & Act. & Leaves keep floating in place right after all the logs of their tree are removed. & 96 & 96 & 96 \\
N3 & M & Fluid & Act. & When a water source block is removed, the flowing water that came from it disappears. & 96 & 90 & 96 \\
N4 & M & Fluid & Act. & Placing a dry sponge in water instantly removes the water in a wide area around it. & 96 & 96 & 94 \\
N7 & M & Grav. & Act. & A horizontal beam of oak planks stays in the air after its only pillar is removed. & 96 & 98 & 92 \\
N8 & M & Str. & Act. & An anvil that falls onto a glass block lands on it and does not break it. & 67 & 80 & 51 \\
N9 & M & Str. & Act. & Leaf blocks can hold up a heavy block such as stone without being affected. & 100 & 96 & 90 \\
N10 & M & Fluid & Act. & A wooden block placed under water stays where it was placed. & 98 & 100 & 98 \\
N11 & M & Fluid & Act. & A stone block placed at the surface of deep water stays where it was placed. & 84 & 78 & 66 \\
N13 & M & Fluid & Act. & A bucket of water scooped from the middle of a pool is replaced at once, and the pool stays exactly as full as before. & 89 & 92 & 84 \\
N14 & M & Fluid & Act. & Flowing water breaks a flower it reaches and turns it into an item. & 77 & 79 & 74 \\
N19 & M & Att. & Act. & A lily pad breaks into an item when the water under it is removed. & 98 & 100 & 98 \\
N20 & M & React. & Act. & A glass block broken by hand leaves nothing behind. & 100 & 94 & 100 \\
N21 & M & React. & Act. & A log can be broken with a bare hand. & 98 & 94 & 96 \\
N22 & M & React. & Act. & A redstone torch placed next to TNT sets it off. & 55 & 52 & 52 \\
N23 & M & Grav. & Act. & A falling sand block that lands on a torch breaks into an item and the torch stays. & 83 & 94 & 81 \\
N26 & M & React. & Act. & A lava source block touched by water instantly becomes obsidian. & 98 & 96 & 94 \\
V1 & M & Att. & Act. & A door breaks into an item when the block under it is removed. & 72 & 83 & 67 \\
V2 & M & Att. & Act. & A carpet breaks into an item when the block under it is removed. & 66 & 64 & 66 \\
V3 & M & Fluid & Act. & Water cannot flow through iron bars or fences. & 65 & 59 & 40 \\
V4 & M & React. & Act. & A block of redstone touching a piston makes it extend. & 98 & 72 & 71 \\
V5 & M & React. & Act. & A block of redstone next to a redstone lamp lights it up. & 100 & 84 & 68 \\
V6 & M & React. & Act. & Ice broken by hand with only air under it leaves nothing behind. & 66 & 56 & 62 \\
V7 & M & Grav. & Act. & A falling sand block that lands on a slab breaks into an item. & 36 & 49 & 31 \\
V8 & M & Plant & Act. & Leaves vanish where they are some time after their tree's logs are removed. & 100 & 99 & 94 \\
V9 & M & React. & Act. & A dirt path turns into dirt when a solid block is placed on it. & 83 & 54 & 36 \\
K1 & M & Heat & Cons. & An ice block right next to a lit torch melts. & 42 & 46 & 75 \\
K2 & M & Fluid & Cons. & Water poured on a raised block flows over its edges down to the ground. & 100 & 100 & 100 \\
K3 & M & Grav. & Cons. & A sand block placed above water falls through the water to the bottom. & 98 & 100 & 100 \\
K4 & M & Grav. & Cons. & Concrete powder falls when the block under it is removed. & 95 & 98 & 100 \\
K5 & M & Heat & Cons. & A fire on top of a stone block burns out after a short time. & 42 & 56 & 67 \\
K6 & M & React. & Cons. & A copper block slowly turns green over time. & 99 & 100 & 100 \\
K7 & M & Plant & Cons. & A grass block covered by a solid block turns into dirt. & 81 & 79 & 92 \\
K8 & M & Fluid & Cons. & Water cannot flow through a glass block. & 100 & 100 & 100 \\
K9 & M & Fluid & Cons. & Water reaching a torch washes it away. & 100 & 96 & 85 \\
K10 & M & Fluid & Cons. & A wet sponge placed in water does not absorb any water. & 56 & 86 & 83 \\
K11 & M & Heat & Cons. & A thin layer of snow right next to a lit torch melts. & 65 & 46 & 96 \\
\bottomrule
\end{tabular*}
\caption{Rule catalogue: every Minecraft rule, as the game behaves in its first variant. Set: main (M) or hard (H);
type: only time or weather acts (Pas.), something is done to the scene (Act.) or consistent (Cons.). Mean accuracy (\%)
of the six models in text (T), when told (Td) and on the bare image (I).}
\label{tab:catalogue}
\end{table*}

\begin{table*}[p]
\centering
\scriptsize
\setlength{\tabcolsep}{3pt}\renewcommand{\arraystretch}{0.86}
\begin{tabular*}{\textwidth}{@{\extracolsep{\fill}}llll>{\raggedright\arraybackslash}p{0.68\textwidth}rrr@{}}
\toprule
Rule & Set & Family & Type & Rule in the game & T & Td & I \\
\midrule
K12 & M & Att. & Cons. & A hanging lantern drops when the block above it is removed. & 96 & 92 & 96 \\
K13 & M & Att. & Cons. & A block placed two blocks away from a cactus does not affect it. & 100 & 94 & 94 \\
N1 & M & Att. & Cons. & Sugar cane does not break at the moment the water next to its base is removed. & 78 & 76 & 77 \\
Q1 & M & Att. & Cons. & A ladder drops when the block it is attached to is removed. & 83 & 82 & 92 \\
Q2 & M & Fluid & Cons. & Flowing water puts out a fire it reaches. & 88 & 93 & 92 \\
Q3 & M & Heat & Cons. & An ice block right next to lava melts. & 55 & 67 & 82 \\
Q4 & M & Plant & Cons. & Grass spreads from grass blocks onto nearby bare dirt. & 100 & 99 & 100 \\
Q5 & M & Grav. & Cons. & Two stacked sand blocks both fall when the block under them is removed. & 98 & 90 & 94 \\
Q6 & M & Fluid & Cons. & Lava poured on a raised block flows over its edges down to the ground. & 94 & 95 & 99 \\
Q7 & M & Att. & Cons. & A lever drops when the block it is attached to is removed. & 96 & 99 & 100 \\
Q8 & M & Plant & Cons. & A sapling drops when the block it is planted on is removed. & 79 & 98 & 94 \\
Q9 & M & Heat & Cons. & Thin snow right next to lava melts. & 74 & 83 & 94 \\
Q10 & M & React. & Cons. & TNT that explodes on dry land leaves a crater. & 98 & 95 & 100 \\
Q11 & M & Plant & Cons. & Wheat planted in watered soil grows to full height over time. & 83 & 97 & 100 \\
Q12 & M & Heat & Cons. & A bricks block touching lava is not damaged. & 100 & 99 & 91 \\
Q13 & M & React. & Cons. & A lever switched on next to a piston makes it extend. & 79 & 72 & 88 \\
Q14 & M & Fluid & Cons. & Water flows through an open gap in a wall. & 89 & 90 & 97 \\
Y1 & H & Time & Pas. & A wool block left out in the rain stays exactly the same. & 100 & 100 & 92 \\
Y2 & H & Heat & Pas. & Coloured wool left in the sun keeps its colour. & 100 & 99 & 57 \\
Y3 & H & Time & Pas. & A bookshelf left out in the rain stays exactly the same. & 100 & 100 & 41 \\
Y4 & H & Time & Pas. & A wooden fence keeps its colour however many years pass. & 100 & 100 & 24 \\
Y5 & H & Plant & Pas. & A moss block in the desert stays green. & 100 & 97 & 0 \\
Y6 & H & Plant & Pas. & Lily pads stay green through the winter. & 100 & 97 & 4 \\
Y7 & H & Fluid & Pas. & A pond on an ordinary grassy plain does not freeze over. & 41 & 60 & 0 \\
Y8 & H & Heat & Pas. & Ice on a pond does not melt in sunshine. & 30 & 35 & 0 \\
Y9 & H & Plant & Pas. & Wheat keeps growing on dry soil with no water nearby. & 98 & 79 & 3 \\
Y10 & H & Time & Pas. & Ripe wheat that is never harvested stays standing and golden. & 100 & 99 & 85 \\
Y11 & H & Time & Pas. & A carved pumpkin never rots. & 100 & 100 & 86 \\
Y12 & H & Heat & Pas. & A campfire burns forever without anyone adding wood. & 96 & 88 & 53 \\
Y13 & H & Heat & Pas. & Rain does not put out a campfire. & 81 & 49 & 2 \\
Y14 & H & Heat & Pas. & Rain does not put out a fire burning on netherrack. & 90 & 68 & 24 \\
Y15 & H & Heat & Pas. & A torch burns forever. & 98 & 100 & 100 \\
Y16 & H & Heat & Pas. & Rain does not cool a magma block. & 100 & 100 & 96 \\
Y17 & H & Heat & Pas. & Warm rain does not melt a snow block. & 94 & 91 & 12 \\
Y18 & H & Fluid & Pas. & Water in a cauldron never evaporates. & 100 & 76 & 1 \\
Y19 & H & Fluid & Pas. & Water in a hole in sand stays there forever. & 100 & 99 & 32 \\
Y20 & H & Fluid & Pas. & Rain does not add any water to a pool. & 96 & 95 & 32 \\
Y21 & H & React. & Pas. & Rain does not harden concrete powder. & 94 & 83 & 22 \\
Y22 & H & Fluid & Pas. & A dry sponge left in the rain stays dry. & 94 & 88 & 1 \\
Y23 & H & Heat & Pas. & A wet sponge next to a campfire stays wet. & 85 & 56 & 26 \\
Y24 & H & Time & Pas. & A dirt path does not turn muddy in the rain. & 100 & 97 & 19 \\
Y25 & H & Plant & Pas. & Frost does not harm sugar cane. & 100 & 99 & 55 \\
Y26 & H & Plant & Pas. & Vines stay green and leafy through the winter. & 94 & 96 & 26 \\
Y27 & H & Plant & Pas. & Sea pickles out of water survive unchanged. & 90 & 79 & 41 \\
Y28 & H & Time & Pas. & Wind and rain do not damage a cobweb. & 100 & 97 & 44 \\
Y29 & H & Time & Pas. & A bone block left out in the weather never changes. & 100 & 100 & 62 \\
Y30 & H & Time & Pas. & A dried kelp block left in the rain never rots. & 100 & 100 & 89 \\
Y31 & H & Heat & Pas. & A honeycomb block in the desert sun keeps its shape. & 100 & 99 & 61 \\
Y32 & H & Time & Pas. & Rain does not wash away a honey block. & 100 & 100 & 49 \\
Y33 & H & Plant & Pas. & A tree in the desert stays green without any rain. & 98 & 99 & 12 \\
Y34 & H & Plant & Pas. & A flower in a pot never needs watering. & 100 & 99 & 15 \\
Y35 & H & Time & Pas. & A brick wall never weathers or cracks. & 100 & 100 & 8 \\
Y36 & H & Time & Pas. & Rain does not soften packed mud. & 96 & 90 & 26 \\
Y37 & H & React. & Pas. & An anvil left in the rain never rusts. & 100 & 99 & 66 \\
Y38 & H & React. & Act. & Punching a block of leaves for a few seconds removes the whole block. & 100 & 99 & 90 \\
Y39 & H & Grav. & Act. & A barrel stays floating when the block under it is removed. & 67 & 66 & 54 \\
Y40 & H & Att. & Act. & A flower pot stays floating when the block under it is removed. & 72 & 72 & 65 \\
Y41 & H & Fluid & Act. & Water flowing next to a campfire does not put it out. & 48 & 29 & 23 \\
Y42 & H & Heat & Act. & Water poured onto a magma block does not cool it. & 96 & 94 & 89 \\
Y43 & H & Grav. & Act. & A block of ice let go in the air stays where it is. & 30 & 23 & 2 \\
Y44 & H & Att. & Act. & A vine disappears when the block it clings to is removed. & 53 & 71 & 74 \\
Y45 & H & Fluid & Act. & Water poured at the end of a row of wheat flows over the row and breaks the wheat into dropped items. & 76 & 93 & 81 \\
Y46 & H & React. & Act. & Breaking an ice block that sits on grass leaves water behind. & 34 & 53 & 33 \\
Y47 & H & Heat & Act. & A torch placed on a hay bale does not set it on fire. & 94 & 97 & 49 \\
Y48 & H & Plant & Cons. & Sugar cane next to water grows taller over time. & 91 & 95 & 100 \\
Y49 & H & Plant & Cons. & A cocoa pod on a jungle log grows and ripens over time. & 67 & 74 & 95 \\
Y50 & H & Plant & Cons. & A sweet berry bush grows berries over time. & 85 & 64 & 75 \\
Y51 & H & Plant & Cons. & A bamboo shoot grows into a tall stalk. & 90 & 83 & 100 \\
Y52 & H & Plant & Cons. & Carrots planted on watered soil grow to full size. & 90 & 91 & 99 \\
Y53 & H & Grav. & Cons. & A hanging stone spike falls when the block above it is removed. & 61 & 38 & 50 \\
Y54 & H & React. & Cons. & Breaking ice over water leaves water in its place. & 100 & 94 & 99 \\
Y55 & H & React. & Cons. & Fire burning next to TNT sets it off. & 72 & 66 & 83 \\
Y56 & H & Heat & Cons. & Ice next to a burning campfire melts. & 51 & 49 & 84 \\
Y57 & H & Heat & Cons. & Fire spreads to a block of leaves next to it. & 81 & 94 & 98 \\
Y58 & H & Plant & Cons. & A full-grown pumpkin vine grows a pumpkin next to it. & 72 & 28 & 69 \\
Y59 & H & Heat & Cons. & Fire spreads to a hay bale next to it. & 92 & 70 & 86 \\
Y60 & H & Fluid & Cons. & Flowing water washes away thin snow. & 84 & 90 & 86 \\
\bottomrule
\end{tabular*}
\caption{Every Minecraft rule (continued).}
\end{table*}

\onecolumn
\raggedbottom
\section{Supplementary Tables and Figures}
\label{app:results}

\noindent\begin{minipage}[t]{0.485\textwidth}\vspace{0pt}
\tabbox{\textwidth}{\centering
\footnotesize
\setlength{\tabcolsep}{3pt}
\begin{tabular*}{\columnwidth}{@{\extracolsep{\fill}}lcccccc@{}}
\toprule
& \multicolumn{3}{c}{Main set} & \multicolumn{3}{c}{Hard set} \\
\cmidrule(lr){2-4} \cmidrule(l){5-7}
Family & Rules & Image & Told & Rules & Image & Told \\
\midrule
Light and heat & 21/6 & 58 & 77 & 12/3 & 48 & 81 \\
Time & 4/0 & 72 & 98 & 12/0 & 52 & 99 \\
Plants & 3/4 & 88 & 97 & 8/6 & 19 & 93 \\
Fluids & 11/7 & 81 & 87 & 7/1 & 24 & 77 \\
Gravity & 6/3 & 63 & 81 & 2/1 & 28 & 44 \\
Attachment & 6/5 & 72 & 79 & 2/0 & 69 & 72 \\
Reactions & 14/3 & 71 & 78 & 4/2 & 53 & 83 \\
Strength & 2/0 & 70 & 88 & 0/0 & -- & -- \\
\midrule
Total & 67/28 & 69 & 82 & 47/13 & 41 & 85 \\
\bottomrule
\end{tabular*}
\caption{Rules per family (conflict / consistent; each rule has eight variants) and the mean accuracy (\%) of the six
models on the family's conflict rules, from the bare image and when told.}
\label{tab:app_families}}\par\vspace{10pt}
\tabbox{\textwidth}{\centering
\small
\setlength{\tabcolsep}{3pt}
\begin{tabular*}{\columnwidth}{@{\extracolsep{\fill}}>{\raggedright\arraybackslash}p{0.5\columnwidth}ccc@{}}
\toprule
Comparison (six models) & Tests & $p<.05$ & Holm \\
\midrule
Told $-$ Image, conflict rules, both sets & 12 & 12 & 12 \\
Gain in balanced accuracy, main set & 6 & 2 & 2 \\
Gain in balanced accuracy, hard set & 6 & 6 & 4 \\
Passive minus action gain, hard set & 6 & 6 & 6 \\
\bottomrule
\end{tabular*}
\caption{The main comparisons as bootstrap tests (two-sided, 2{,}000 samples): how many of the per-model tests reach
$p<.05$, without and with Holm correction within each row.}
\label{tab:app_holm}}
\end{minipage}\hfill
\begin{minipage}[t]{0.485\textwidth}\vspace{0pt}
\tabbox{\textwidth}{\centering
\small
\setlength{\tabcolsep}{3.5pt}
\begin{tabular*}{\columnwidth}{@{\extracolsep{\fill}}lcc@{}}
\toprule
& Image errors: & Told errors: \\
Model & real-world & game exception \\
\midrule
\multicolumn{3}{@{}l}{\emph{Main set}} \\
DeepSeek & 91 & 100 \\
Qwen-Max & 81 & 100 \\
Qwen-Flash & 92 & 81 \\
Doubao-Pro & 98 & 100 \\
Doubao-Lite & 91 & 97 \\
GPT-6.1-Sol & 100 & \textendash \\
Pooled & \aMRealPooled & \aMOverPooled \\
\midrule
\multicolumn{3}{@{}l}{\emph{Hard set}} \\
DeepSeek & 96 & 100 \\
Qwen-Max & 96 & 100 \\
Qwen-Flash & 95 & 92 \\
Doubao-Pro & 96 & 97 \\
Doubao-Lite & 96 & 97 \\
GPT-6.1-Sol & 99 & 100 \\
Pooled & \aHRealPooled & \aHOverPooled \\
\bottomrule
\end{tabular*}
\caption{Where errors go (\%): the share of bare-image errors on conflict rules that pick the real-world outcome, and
the share of told errors on consistent rules that pick the game exception (\textendash: no told errors).}
\label{tab:direction}}\par\vspace{10pt}
\tabbox{\textwidth}{\centering\scriptsize
\setlength{\tabcolsep}{3pt}\renewcommand{\arraystretch}{0.92}
\begin{tabular*}{\textwidth}{@{\extracolsep{\fill}}>{\raggedright\arraybackslash}p{0.54\textwidth}>{\raggedright\arraybackslash}p{0.38\textwidth}@{}}
\toprule
Game mechanism & Rules \\
\midrule
Blocks need no support & N2, N7, F1, F2, F3 \\
Loads never break blocks & N8, N9 \\
Falling blocks break on small blocks & N23, V7 \\
Attached blocks pop off or vanish & N19, V1, V2, F4, F15 \\
Fluid flow follows fixed ranges and sources & N3, N4, N12, N13, M3 \\
No buoyancy & N10, N11 \\
Water does not wash away or seep & N14, M7, V3, F8 \\
No evaporation or drying & N15, M8 \\
Melting depends on block light only & N16, N17, N18, M1, H3, H4, H6, F10 \\
Lava does not melt or ignite neighbours & H2, M5, H5 \\
Flames do not spread or ignite & M4, H1, H7, H8 \\
Fire persists or vanishes without residue & N6, F11, F13 \\
Lava and water react by fixed rules & N5, N26, F14 \\
Nothing ages & N24, N25, H9, H10, F6, F7 \\
Breaking yields fixed drops & N20, N21, M6, V6 \\
Redstone and immovable blocks & N22, V4, V5, F12 \\
Soil and plant rules & M2, V8, V9, F5 \\
Explosions in water & F9 \\
\bottomrule
\end{tabular*}
\caption{The 18 game mechanisms behind the main set's 67 conflict rules, the
groups resampled by the clustered bootstrap (Section~\ref{sec:phen}).}
\label{tab:mech}}
\end{minipage}

\vspace{7pt}
\noindent\tabbox{\textwidth}{\centering
\small
\setlength{\tabcolsep}{0pt}
\begin{tabular*}{\textwidth}{@{\extracolsep{\fill}}lccccccc@{}}
\toprule
& \multicolumn{2}{c}{Text} & \multicolumn{2}{c}{Image} & \multicolumn{2}{c}{Told} & \\
\cmidrule(lr){2-3} \cmidrule(lr){4-5} \cmidrule(lr){6-7}
Model & Lean & Bal. & Lean & Bal. & Lean & Bal. & Bal.\ gain \\
\midrule
\multicolumn{8}{@{}l}{\emph{Main set}} \\
DeepSeek & +3.1 {\scriptsize[$-$10.2, 17.5]} & 82.0 & $-$37.6 {\scriptsize[$-$49.1, $-$25.2]} & 74.6 & $-$11.8 & 81.6 & +6.9 {\scriptsize[1.1, 12.4]} \\
Qwen-Max & $-$2.3 {\scriptsize[$-$12.3, 7.8]} & 90.3 & $-$17.0 {\scriptsize[$-$28.4, $-$6.1]} & 83.5 & +1.8 & 84.4 & +0.9 {\scriptsize[$-$4.6, 6.0]} \\
Qwen-Flash & $-$1.9 {\scriptsize[$-$14.2, 11.4]} & 75.4 & $-$25.8 {\scriptsize[$-$37.8, $-$12.8]} & 71.8 & $-$11.9 & 73.1 & +1.3 {\scriptsize[$-$3.3, 6.0]} \\
Doubao-Pro & +3.7 {\scriptsize[$-$9.2, 17.6]} & 82.7 & $-$19.1 {\scriptsize[$-$31.2, $-$7.4]} & 82.0 & $-$0.1 & 85.7 & +3.7 {\scriptsize[$-$1.2, 8.6]} \\
Doubao-Lite & +4.2 {\scriptsize[$-$10.0, 19.5]} & 80.2 & $-$32.1 {\scriptsize[$-$43.3, $-$21.4]} & 77.2 & $-$7.7 & 80.5 & +3.3 {\scriptsize[$-$2.0, 8.1]} \\
GPT-6.1-Sol & +0.0 {\scriptsize[0.0, 0.0]} & 100.0 & $-$11.2 {\scriptsize[$-$18.2, $-$5.3]} & 94.0 & $-$0.1 & 100.0 & +6.0 {\scriptsize[3.1, 9.5]} \\
\multicolumn{8}{@{}l}{\emph{Hard set}} \\
DeepSeek & +4.4 {\scriptsize[$-$14.0, 24.2]} & 83.0 & $-$67.6 {\scriptsize[$-$79.0, $-$55.2]} & 62.7 & +5.0 & 78.5 & +15.8 {\scriptsize[5.2, 25.0]} \\
Qwen-Max & +7.1 {\scriptsize[$-$10.8, 27.9]} & 85.3 & $-$16.3 {\scriptsize[$-$36.5, 5.9]} & 65.9 & +32.8 & 75.1 & +9.2 {\scriptsize[0.3, 17.5]} \\
Qwen-Flash & +4.4 {\scriptsize[$-$12.4, 23.1]} & 76.2 & $-$46.2 {\scriptsize[$-$59.3, $-$30.3]} & 59.6 & +1.2 & 70.8 & +11.2 {\scriptsize[3.3, 18.1]} \\
Doubao-Pro & +25.6 {\scriptsize[11.3, 41.2]} & 76.3 & $-$40.9 {\scriptsize[$-$57.2, $-$24.0]} & 65.1 & +23.7 & 74.3 & +9.2 {\scriptsize[0.3, 17.9]} \\
Doubao-Lite & +10.9 {\scriptsize[$-$6.6, 32.7]} & 83.4 & $-$50.6 {\scriptsize[$-$70.1, $-$28.6]} & 56.4 & +15.5 & 76.0 & +19.6 {\scriptsize[9.9, 27.8]} \\
GPT-6.1-Sol & $-$1.3 {\scriptsize[$-$4.8, 0.0]} & 99.3 & $-$52.1 {\scriptsize[$-$65.0, $-$39.1]} & 73.0 & +2.1 & 97.2 & +24.2 {\scriptsize[17.1, 31.1]} \\
\bottomrule
\end{tabular*}
\caption{Lean (conflict minus consistent accuracy, points) and balanced accuracy (their mean, \%) in each condition,
with 95\% CIs; Bal.\ gain: Told $-$ Image in balanced accuracy. In text the lean is close to zero, so accuracy in text
is not driven by a preference for the game's outcome; on the bare image it is negative.}
\label{tab:app_default}}

\vspace{7pt}
\noindent\tabbox{\textwidth}{\centering
\small
\begin{tabular}{@{}l ccc ccc ccc@{}}
\toprule
& \multicolumn{3}{c}{Conflict, passive} & \multicolumn{3}{c}{Conflict, action} & \multicolumn{3}{c}{Consistent} \\
\cmidrule(lr){2-4} \cmidrule(lr){5-7} \cmidrule(l){8-10}
Model & Text & Told & Image & Text & Told & Image & Text & Told & Image \\
\midrule
\multicolumn{10}{@{}l}{\emph{Main set}} \\
DeepSeek & 84.5 & 72.0 & 41.8 & 82.7 & 79.0 & 68.7 & 80.5 & 87.5 & 93.5 \\
Qwen-Max & 87.1 & 93.0 & 76.2 & 91.1 & 78.2 & 73.9 & 91.5 & 83.5 & 92.0 \\
Qwen-Flash & 79.7 & 68.1 & 54.7 & 69.6 & 66.2 & 62.7 & 76.3 & 79.0 & 84.7 \\
Doubao-Pro & 89.8 & 88.3 & 69.9 & 79.6 & 83.2 & 74.6 & 80.8 & 85.7 & 91.5 \\
Doubao-Lite & 89.1 & 75.0 & 55.9 & 76.1 & 78.2 & 66.1 & 78.1 & 84.4 & 93.3 \\
GPT-6.1-Sol & 100.0 & 100.0 & 77.0 & 100.0 & 99.9 & 98.8 & 100.0 & 100.0 & 99.6 \\
\multicolumn{10}{@{}l}{\emph{Hard set}} \\
DeepSeek & 93.6 & 85.0 & 23.9 & 54.2 & 65.8 & 47.5 & 80.8 & 76.0 & 96.5 \\
Qwen-Max & 94.3 & 93.6 & 53.4 & 68.8 & 83.8 & 73.8 & 81.7 & 58.7 & 74.0 \\
Qwen-Flash & 85.8 & 74.9 & 35.4 & 51.2 & 58.3 & 40.8 & 74.0 & 70.2 & 82.7 \\
Doubao-Pro & 94.9 & 93.9 & 44.9 & 67.5 & 57.5 & 43.8 & 63.5 & 62.5 & 85.6 \\
Doubao-Lite & 94.9 & 89.9 & 26.7 & 66.2 & 61.3 & 47.5 & 77.9 & 68.3 & 81.7 \\
GPT-6.1-Sol & 100.0 & 100.0 & 37.4 & 93.8 & 91.7 & 82.5 & 100.0 & 96.2 & 99.0 \\
\bottomrule
\end{tabular}
\caption{Accuracy (\%) by rule type. Passive: only time or weather acts on the scene; action: something is done to it.
On the hard set the type is part of each rule; on the main set it is read from the action text.}
\label{tab:app_types}}

\vspace{7pt}
\noindent\tabbox{\textwidth}{\centering\small
\begin{tabular}{l cccc cccc}
\toprule
 & \multicolumn{4}{c}{Qwen3.5-9B} & \multicolumn{4}{c}{InternVL3.5-8B} \\
\cmidrule(lr){2-5}\cmidrule(lr){6-9}
Rules & Text & Told & Image & Told $-$ Image & Text & Told & Image & Told $-$ Image \\
\midrule
Main, conflict & \aWMCTextQ & \aWMCToldQ & \aWMCImageQ & \aWMCGapQ & \aWMCTextI & \aWMCToldI & \aWMCImageI & \aWMCGapI \\
Main, passive conflict & \aWMPTextQ & \aWMPToldQ & \aWMPImageQ & \aWMPGapQ & \aWMPTextI & \aWMPToldI & \aWMPImageI & \aWMPGapI \\
Main, consistent & \aWMSTextQ & \aWMSToldQ & \aWMSImageQ & \aWMSGapQ & \aWMSTextI & \aWMSToldI & \aWMSImageI & \aWMSGapI \\
Hard, conflict & \aWHCTextQ & \aWHCToldQ & \aWHCImageQ & \aWHCGapQ & \aWHCTextI & \aWHCToldI & \aWHCImageI & \aWHCGapI \\
Hard, passive conflict & \aWHPTextQ & \aWHPToldQ & \aWHPImageQ & \aWHPGapQ & \aWHPTextI & \aWHPToldI & \aWHPImageI & \aWHPGapI \\
Hard, consistent & \aWHSTextQ & \aWHSToldQ & \aWHSImageQ & \aWHSGapQ & \aWHSTextI & \aWHSToldI & \aWHSImageI & \aWHSGapI \\
\bottomrule
\end{tabular}
\caption{Open models in the three conditions (accuracy, \%; chance 33\%). Told $-$ Image on hard conflict rules:
Qwen3.5-9B \aWHCGapQ{} \aWHCGapCIQ, InternVL3.5-8B \aWHCGapI{} \aWHCGapCII; on main conflict rules \aWMCGapQ{}
\aWMCGapCIQ{} and \aWMCGapI{} \aWMCGapCII. Bare-image errors on conflict rules that pick the real-world outcome:
\aWMRealQ{} and \aWMRealI\% on the main set, \aWHRealQ{} and \aWHRealI\% on the hard set.}
\label{tab:wb_behav}}

\vspace{7pt}
\noindent\figbox{\textwidth}{\centering
\includegraphics[width=0.8\textwidth]{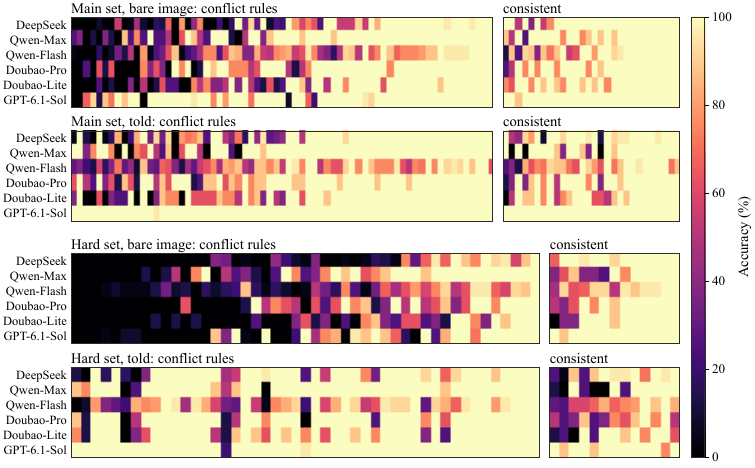}
\caption{Accuracy per rule (columns) and model (rows), on the bare image and when told. Within each set, conflict rules
fill the left block and consistent rules the right one, both sorted by the models' mean accuracy on the bare
image. Dark cells on the left of the bare-image panels are shared across models and turn bright when the game is
named.}
\label{fig:perrule}}

\vspace{7pt}
\noindent\tabbox{\textwidth}{\centering
\small
\setlength{\tabcolsep}{0pt}
\begin{tabular*}{\textwidth}{@{\extracolsep{\fill}}l cc cc cc @{\hspace{10pt}} cc cc cc@{}}
\toprule
& \multicolumn{6}{c}{Main set} & \multicolumn{6}{c}{Hard set} \\
\cmidrule(lr){2-7} \cmidrule(l){8-13}
& \multicolumn{2}{c}{DeepSeek} & \multicolumn{2}{c}{GPT-6.1-Sol} & \multicolumn{2}{c}{Qwen-Flash} & \multicolumn{2}{c}{DeepSeek} & \multicolumn{2}{c}{GPT-6.1-Sol} & \multicolumn{2}{c}{Qwen-Flash} \\
\cmidrule(lr){2-3} \cmidrule(lr){4-5} \cmidrule(lr){6-7} \cmidrule(lr){8-9} \cmidrule(lr){10-11} \cmidrule(lr){12-13}
Prompt & Conf. & Cons. & Conf. & Cons. & Conf. & Cons. & Conf. & Cons. & Conf. & Cons. & Conf. & Cons. \\
\midrule
Direct answer & 48.0 & 94.9 & 86.8 & \textbf{100.0} & 52.3 & 90.8 & 23.9 & \textbf{96.5} & 37.4 & 99.0 & 35.4 & \textbf{82.7} \\
\addlinespace[3pt]
Chain of thought & 48.2 & 95.8 & 88.1 & \textbf{100.0} & 64.1 & \textbf{97.9} & 33.6 & 85.9 & 37.5 & \textbf{100.0} & 53.0 & 78.5 \\
Generated knowledge & 64.4 & 96.4 & 89.9 & \textbf{100.0} & 66.2 & 97.0 & 55.9 & 84.3 & 47.3 & 97.1 & 61.9 & 70.5 \\
\addlinespace[3pt]
\emph{The world shown} & 53.8 & 92.9 & -- & -- & 58.2 & 90.2 & 49.9 & 93.9 & 91.7 & 93.3 & 59.9 & 80.1 \\
\emph{A video game} & 62.5 & 90.5 & \textbf{100.0} & \textbf{100.0} & 66.6 & 87.5 & 78.6 & 83.3 & 99.7 & 93.3 & 81.8 & 61.9 \\
\addlinespace[3pt]
\textcolor{gray}{Told} & \textcolor{gray}{70.9} & \textcolor{gray}{90.5} & \textcolor{gray}{99.9} & \textcolor{gray}{100.0} & \textcolor{gray}{64.2} & \textcolor{gray}{85.7} & \textcolor{gray}{85.0} & \textcolor{gray}{76.0} & \textcolor{gray}{100.0} & \textcolor{gray}{96.2} & \textcolor{gray}{74.9} & \textcolor{gray}{70.2} \\
\textcolor{gray}{Told + chain of thought} & \textcolor{gray}{71.9} & \textcolor{gray}{91.7} & \textcolor{gray}{99.7} & \textcolor{gray}{98.2} & \textcolor{gray}{76.4} & \textcolor{gray}{94.3} & \textcolor{gray}{86.4} & \textcolor{gray}{72.4} & \textcolor{gray}{100.0} & \textcolor{gray}{97.1} & \textcolor{gray}{89.8} & \textcolor{gray}{71.2} \\
\bottomrule
\end{tabular*}
\caption{Accuracy (\%) on conflict (Conf.) and consistent (Cons.) rules by prompt on both sets. Grey rows name the game for the model. Main set: batches 2\textendash 4, which played no part in writing the prompts
(\textendash: not run). Hard set: conflict rules where only time or weather acts, and all consistent rules. Bold: best in a column among
the prompts that do not name the game.}
\label{tab:app_method}}

\vspace{7pt}
\noindent\tabbox{\textwidth}{\centering\small
\setlength{\tabcolsep}{3.5pt}
\begin{tabular}{l cc c cc c cc c cc c}
\toprule
 & \multicolumn{6}{c}{Qwen3.5-9B} & \multicolumn{6}{c}{InternVL3.5-8B} \\
\cmidrule(lr){2-7}\cmidrule(lr){8-13}
 & \multicolumn{3}{c}{Main} & \multicolumn{3}{c}{Hard} & \multicolumn{3}{c}{Main} & \multicolumn{3}{c}{Hard} \\
\cmidrule(lr){2-4}\cmidrule(lr){5-7}\cmidrule(lr){8-10}\cmidrule(lr){11-13}
Condition & Conf. & Cons. & Bal. & Conf. & Cons. & Bal. & Conf. & Cons. & Bal. & Conf. & Cons. & Bal. \\
\midrule
Image & 39.7 & 80.4 & 60.0 & 31.9 & 74.0 & 53.0 & 29.1 & 76.8 & 52.9 & 13.6 & 78.8 & 46.2 \\
Image + direction & 53.7 & 55.4 & 54.5 & 54.0 & 46.2 & 50.1 & 38.4 & 80.8 & 59.6 & 24.7 & 95.2 & 60.0 \\
Image + random direction & 26.5 & 79.0 & 52.8 & 8.0 & 88.5 & 48.2 & 36.6 & 62.1 & 49.3 & 25.0 & 75.0 & 50.0 \\
\midrule
Told & 50.6 & 73.2 & 61.9 & 55.3 & 51.0 & 53.1 & 34.7 & 81.7 & 58.2 & 25.0 & 73.1 & 49.0 \\
Told $-$ direction & 39.2 & 85.3 & 62.2 & 37.0 & 66.3 & 51.7 & 42.5 & 54.0 & 48.3 & 27.1 & 39.4 & 33.3 \\
\bottomrule
\end{tabular}
\caption{Steering on the test sets (accuracy, \%). Conf.\ and Cons.: conflict and consistent rules; Bal.: their mean. The direction is the mean told $-$ image state over the dev rules, added at every position of one layer (Qwen3.5-9B: layer \aWSteerLayerQ, InternVL3.5-8B: layer \aWSteerLayerI) with strength 3; the random direction has the same norm.}
\label{tab:wb_steer}}

\twocolumn

\section{Further Details and Results}
\label{app:arxiv}

\subsection{Which World and Gameplay Tool, per Model}
\label{app:arxiv_interp}

\begin{table*}[t]
\centering
\scriptsize
\setlength{\tabcolsep}{0pt}
\begin{tabular*}{\textwidth}{@{\extracolsep{\fill}}l cc cc cc cc cc@{}}
\toprule
 & \multicolumn{4}{c}{Which world} & \multicolumn{6}{c}{Gameplay tool} \\
\cmidrule(lr){2-5}\cmidrule(l){6-11}
 & \multicolumn{2}{c}{Game (\%)} & \multicolumn{2}{c}{Told $-$ Image, read as game} & \multicolumn{2}{c}{Conflict (\%)} &
\multicolumn{2}{c}{Conflict, vs.\ Image} & \multicolumn{2}{c}{Consistent, vs.\ Image} \\
\cmidrule(lr){2-3}\cmidrule(lr){4-5}\cmidrule(lr){6-7}\cmidrule(lr){8-9}\cmidrule(l){10-11}
Model & Main & Hard & Main & Hard & Main & Hard & Main & Hard & Main & Hard \\
\midrule
DeepSeek & 96.3 & 79.2 & +18.6 {\tiny[11.9, 25.6]} & +45.5 {\tiny[33.2, 57.8]} & 50.7 & 21.5 & $-$5.1 {\tiny[$-$10.1, $-$0.6]} & $-$7.4 {\tiny[$-$12.8, $-$2.0]} & $-$1.0 {\tiny[$-$4.8, 2.7]} & +1.6 {\tiny[$-$4.2, 9.0]} \\
Qwen-Max & 97.0 & 72.3 & +7.4 {\tiny[2.7, 12.9]} & +18.1 {\tiny[8.1, 29.2]} & 80.2 & 72.9 & +5.2 {\tiny[0.9, 9.7]} & +15.2 {\tiny[7.4, 23.4]} & $-$0.9 {\tiny[$-$5.8, 4.5]} & $-$9.6 {\tiny[$-$24.0, 3.8]} \\
Qwen-Flash & 99.5 & 92.7 & +8.0 {\tiny[2.5, 13.9]} & +33.9 {\tiny[24.6, 43.3]} & 58.8 & 47.5 & $-$0.1 {\tiny[$-$5.7, 5.0]} & +10.9 {\tiny[4.4, 18.2]} & +0.6 {\tiny[$-$6.8, 8.2]} & $-$1.0 {\tiny[$-$12.2, 8.8]} \\
Doubao-Pro & 93.7 & 61.5 & +10.6 {\tiny[4.5, 17.4]} & +28.1 {\tiny[16.4, 41.5]} & 75.2 & 58.8 & +2.8 {\tiny[$-$2.1, 7.6]} & +14.1 {\tiny[6.9, 22.1]} & $-$2.2 {\tiny[$-$7.6, 2.7]} & $-$14.4 {\tiny[$-$30.8, 1.0]} \\
Doubao-Lite & 95.1 & 72.1 & +13.6 {\tiny[7.6, 20.0]} & +42.0 {\tiny[30.0, 55.3]} & 63.8 & 48.4 & +2.6 {\tiny[$-$3.0, 8.8]} & +17.3 {\tiny[8.8, 26.3]} & $-$9.8 {\tiny[$-$18.8, $-$1.3]} & $-$3.8 {\tiny[$-$16.3, 7.7]} \\
GPT-6.1-Sol & 93.7 & 62.1 & +5.5 {\tiny[2.4, 9.3]} & +22.6 {\tiny[13.1, 35.3]} & 95.1 & 77.9 & +6.8 {\tiny[2.2, 12.1]} & +30.9 {\tiny[21.0, 41.9]} & +0.4 {\tiny[0.0, 1.8]} & $-$1.9 {\tiny[$-$8.7, 2.9]} \\
\bottomrule
\end{tabular*}
\caption{The two probes of Section~\ref{sec:cue}, one run per model. Which world: share of all items read as about
the video game, and Told $-$ Image on the conflict items a model itself reads as about the game. Gameplay tool: accuracy
on the bare image after the tool's system message, and its difference from the bare image without it, paired per item.
Brackets: 95\% intervals resampling rules and items within rules.}
\label{tab:arxiv_interp}
\end{table*}

Table~\ref{tab:arxiv_interp} gives both probes of Section~\ref{sec:cue} for every model, one run each on all 1{,}240
items. In the which-world probe the two letters swap places on every second item; \tMetaParsed\% of its replies name
one of them, and the rest are left out. Told $-$ Image is taken over the conflict items that the model itself reads as
about the game, so it compares the told and bare-image answers on exactly the questions the model places in the game.
The gameplay tool's system message changes consistent rules little for most models; the exceptions are Doubao-Lite on
the main set (\tDepMSGainDL{} points) and Doubao-Pro on the hard set (\tDepHSGainDP).

\subsection{Rules the Game Changes}
\label{app:arxiv_change}

\begin{table*}[t]
\centering
\footnotesize
\setlength{\tabcolsep}{0pt}
\begin{tabular*}{\textwidth}{@{\extracolsep{\fill}}l ccc ccc c@{}}
\toprule
 & \multicolumn{3}{c}{Game changes the scene (28 rules)} &
\multicolumn{3}{c}{Game leaves it as it was (39 rules)} &
Real option: no change (14) \\
\cmidrule(lr){2-4}\cmidrule(lr){5-7}\cmidrule(l){8-8}
Model & Image & Told & Told $-$ Image & Image & Told & Told $-$ Image & Told $-$ Image \\
\midrule
DeepSeek & 61.5 & 74.4 & +12.9 {\tiny[4.3, 21.7]} & 51.8 & 76.6 & +24.8 {\tiny[14.3, 35.5]} & +8.6 {\tiny[1.8, 16.4]} \\
Qwen-Max & 70.1 & 72.3 & +2.2 {\tiny[$-$3.6, 8.0]} & 78.5 & 94.6 & +16.0 {\tiny[7.7, 25.0]} & +0.9 {\tiny[$-$5.4, 7.1]} \\
Qwen-Flash & 65.0 & 66.4 & +1.3 {\tiny[$-$8.6, 11.5]} & 54.5 & 67.6 & +13.1 {\tiny[7.2, 19.3]} & $-$5.7 {\tiny[$-$20.5, 8.9]} \\
Doubao-Pro & 80.4 & 85.3 & +4.9 {\tiny[$-$3.1, 13.8]} & 66.7 & 85.9 & +19.2 {\tiny[10.6, 28.8]} & +13.4 {\tiny[1.8, 26.8]} \\
Doubao-Lite & 63.4 & 75.4 & +12.1 {\tiny[3.6, 21.4]} & 59.6 & 77.6 & +17.9 {\tiny[9.3, 27.6]} & +14.3 {\tiny[0.0, 31.2]} \\
GPT-6.1-Sol & 99.3 & 99.9 & +0.6 {\tiny[$-$0.6, 2.7]} & 80.6 & 100.0 & +19.4 {\tiny[9.4, 30.0]} & +0.0 {\tiny[0.0, 0.0]} \\
\bottomrule
\end{tabular*}
\caption{Main-set conflict rules by the game outcome (Section~\ref{sec:where}): accuracy (\%) on the bare image and
when told. The last column takes the changing rules whose real-world option is that nothing happens. Brackets: 95\%
intervals resampling rules and items within rules.}
\label{tab:arxiv_change}
\end{table*}

Table~\ref{tab:arxiv_change} splits the main set's conflict rules as in Section~\ref{sec:where}. A rule counts as
changing when the game outcome is itself a change, read from the game option of the rule: N3, N4, N13, N14, N19--N23,
N26, M1, M2, M4, M6, V1--V9, F4, F9, F11, F14 and F15 on the main set, and Y9, Y38 and Y44--Y46 on the hard set. On the
changing rules the bare image comes close to the told prompt for four models; on the rules that the game leaves as they
were, every model gains from the name. Of the changing rules, 14 offer ``nothing happens'' as their real-world option,
and there no model answers significantly worse when told. In text, errors on consistent rules also favour the game:
\aMOverTPooled\% (main) and \aHOverTPooled\% (hard) of them pick the game exception.

\subsection{Reasoning Settings}
\label{app:arxiv_reason}

GPT-6.1-Sol does not accept a setting without reasoning and runs at its lowest one, where it reasons before some of its
answers (Table~\ref{tab:reasoning}). At the setting \emph{high} it answers \aHighImage\% of the hard conflict
questions from the bare image (\aHighPImage\% of the passive ones), \aHighTold\% when told and \aHighConsImage\% of the
consistent ones from the bare image, so being told still adds \aHighGap{} points \aHighGapCI.

DeepSeek answered the hard set's image and told conditions once more with its thinking phase switched on (one run;
\aThinkReasked{} answers that used up the output limit of 8{,}000 tokens were asked again with a limit of 32{,}000). On
conflict rules, accuracy rises from \aThinkOffImage{} to \aThinkImage\% on the bare image (\aThinkDiff{} points
\aThinkDiffCI) and from \aThinkOffTold{} to \aThinkTold\% when told, so being told still adds \aThinkGap{} points
\aThinkGapCI; on passive rules the gain is \aThinkPGap{} \aThinkPGapCI{} (\aThinkPImage{} to \aThinkPTold\%).
\aThinkRealShare\% of the remaining bare-image errors pick the real-world outcome. On consistent rules accuracy is
\aThinkConsImage\% on the bare image and \aThinkConsTold\% when told (\aThinkOffConsImage{} and \aThinkOffConsTold\%
with thinking off). A bare-image answer takes a median of \aTokImageThink{} output tokens and a told answer
\aTokToldThink.

\subsection{Open Answers}
\label{app:arxiv_open}

On the \aOpenN{} items of the hard set's passive conflict rules, DeepSeek, GPT-6.1-Sol and Qwen3.8-Flash answered once
without the options, on the bare image and when told. From the bare image they describe the game outcome in
\aOpenGameImageDS, \aOpenGameImageSol{} and \aOpenGameImageQF\% of their answers, and their multiple-choice accuracy on
the same items is \aMethHCDirectDS, \aMethHCDirectSol{} and \aMethHCDirectQF\%: the options neither produce the
real-world answers nor hide game answers. Naming the game adds \aOpenGapMin{} to \aOpenGapMax{} points in this format
too. Told, a model writes out what it recalls, which favours GPT-6.1-Sol and Qwen3.8-Flash; DeepSeek recognizes the game
outcome among three options more often than it recalls it unaided. A third example of GPT-6.1-Sol: after weeks of
desert sun on a moss block, from the bare image ``The moss dries out under the hot desert sun, turning brown and
brittle as it withers.'', and when told ``The moss remains unchanged. In vanilla Minecraft, moss does not dry out or die
from desert heat or a lack of rain.''

\paragraph{Answer options.} Each question's game option states the outcome the engine produces. We wrote the
real-world option as the change that a real scene of the same objects would undergo, and the third option as an
outcome that holds in neither world; for example, after months of rain ``The white wool block has shrunk into a small,
hard lump'' (Y1), and after the water under a lily pad is scooped up ``The lily pad turns into a flower'' (N19). All
options are in the reproducibility package.

\subsection{Cues, Paired}
\label{app:arxiv_cues}

On hard passive conflict rules, \emph{the world shown} leads direct answering by \aNeutvsDirectHCDS{}
\aNeutvsDirectHCCIDS{} points for DeepSeek, \aNeutvsDirectHCSol{} \aNeutvsDirectHCCISol{} for GPT-6.1-Sol and
\aNeutvsDirectHCQF{} \aNeutvsDirectHCCIQF{} for Qwen3.8-Flash, and the told prompt leads the world shown by
\aToldvsNeutHCDS{} \aToldvsNeutHCCIDS, \aToldvsNeutHCSol{} \aToldvsNeutHCCISol{} and \aToldvsNeutHCQF{}
\aToldvsNeutHCCIQF. On consistent rules the world shown differs from direct answering by \aNeutvsDirectHSDS{}
\aNeutvsDirectHSCIDS, \aNeutvsDirectHSSol{} \aNeutvsDirectHSCISol{} and \aNeutvsDirectHSQF{} \aNeutvsDirectHSCIQF. The
told prompt leads chain of thought by \aToldvsCoTHCDS{} \aToldvsCoTHCCIDS, \aToldvsCoTHCSol{} \aToldvsCoTHCCISol{} and
\aToldvsCoTHCQF{} \aToldvsCoTHCCIQF, and generated knowledge by \aToldvsGKPHCDS{} \aToldvsGKPHCCIDS,
\aToldvsGKPHCSol{} \aToldvsGKPHCCISol{} and \aToldvsGKPHCQF{} \aToldvsGKPHCCIQF. The video-game prompt reaches
\aMethHCFixedGameDS, \aMethHCFixedGameSol{} and \aMethHCFixedGameQF\% on passive conflict rules, against
\aMethHCToldDS, \aMethHCToldSol{} and \aMethHCToldQF\% when told.

On the main set the order of the cues is the same (Table~\ref{tab:app_method}). The neutral sentence was run there for
DeepSeek and Qwen3.8-Flash: it adds \nRevNeutDDS{} \nRevNeutDCIDS{} and \nRevNeutDQw{} \nRevNeutDCIQw{} points on
conflict rules, \nRevNeutShareDS\% and \nRevNeutShareQw\% of the gain of being told, and changes consistent rules by
\nRevNeutConsDDS{} and \nRevNeutConsDQw{} points.

\subsection{Self-Named World and Own Notes}
\label{app:arxiv_self}

\begin{figure*}[t]
\centering
\includegraphics[width=\textwidth]{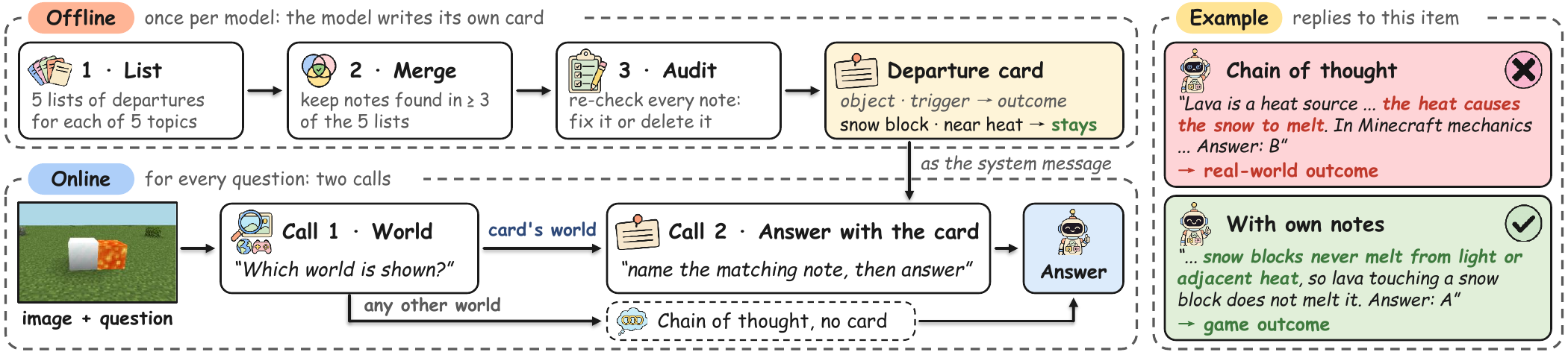}
\caption{Self-named world with own notes. Offline, each model writes and audits its own card of departures from
real-world physics. For each question a first call names the world shown; only for the card's world does a second call
answer with the card, naming the note that applies. Right: replies of DeepSeek to the item of Figure~\ref{fig:teaser}.}
\label{fig:method}
\end{figure*}

\begin{table*}[!t]
\centering
\small
\setlength{\tabcolsep}{0pt}
\begin{tabular*}{\textwidth}{@{\extracolsep{\fill}}l cc cc cc @{\hspace{10pt}} cc cc cc@{}}
\toprule
& \multicolumn{6}{c}{Main set} & \multicolumn{6}{c}{Hard set} \\
\cmidrule(lr){2-7} \cmidrule(l){8-13}
& \multicolumn{2}{c}{DeepSeek} & \multicolumn{2}{c}{GPT-6.1-Sol} & \multicolumn{2}{c}{Qwen-Flash} & \multicolumn{2}{c}{DeepSeek} & \multicolumn{2}{c}{GPT-6.1-Sol} & \multicolumn{2}{c}{Qwen-Flash} \\
\cmidrule(lr){2-3} \cmidrule(lr){4-5} \cmidrule(lr){6-7} \cmidrule(lr){8-9} \cmidrule(lr){10-11} \cmidrule(lr){12-13}
Prompt & Conf. & Cons. & Conf. & Cons. & Conf. & Cons. & Conf. & Cons. & Conf. & Cons. & Conf. & Cons. \\
\midrule
Direct answer & 48.0 & 94.9 & 86.8 & 100.0 & 52.3 & 90.8 & 23.9 & 96.5 & 37.4 & 99.0 & 35.4 & 82.7 \\
\emph{A video game} & 62.5 & 90.5 & 100.0 & 100.0 & 66.6 & 87.5 & 78.6 & 83.3 & 99.7 & 93.3 & 81.8 & 61.9 \\
\addlinespace[3pt]
Self-named world & 71.9 & 91.7 & 99.7 & 98.2 & 75.9 & 94.3 & 86.4 & 72.4 & 100.0 & 97.1 & 89.1 & 71.2 \\
~~+ own notes & 79.9 & 97.3 & 98.8 & 99.1 & 73.9 & 93.2 & 93.9 & 66.0 & 99.0 & 99.0 & 91.7 & 68.3 \\
\addlinespace[3pt]
\textcolor{gray}{Told} & \textcolor{gray}{70.9} & \textcolor{gray}{90.5} & \textcolor{gray}{99.9} & \textcolor{gray}{100.0} & \textcolor{gray}{64.2} & \textcolor{gray}{85.7} & \textcolor{gray}{85.0} & \textcolor{gray}{76.0} & \textcolor{gray}{100.0} & \textcolor{gray}{96.2} & \textcolor{gray}{74.9} & \textcolor{gray}{70.2} \\
\textcolor{gray}{Told + chain of thought} & \textcolor{gray}{71.9} & \textcolor{gray}{91.7} & \textcolor{gray}{99.7} & \textcolor{gray}{98.2} & \textcolor{gray}{76.4} & \textcolor{gray}{94.3} & \textcolor{gray}{86.4} & \textcolor{gray}{72.4} & \textcolor{gray}{100.0} & \textcolor{gray}{97.1} & \textcolor{gray}{89.8} & \textcolor{gray}{71.2} \\
\bottomrule
\end{tabular*}
\caption{Self-named world, with and without the model's own notes, beside direct answering, the video-game cue and the
told prompts (accuracy, \%; grey rows name the game). Main set: batches 2\textendash 4; hard set: passive conflict
rules and all consistent rules.}
\label{tab:arxiv_self}
\end{table*}

A further cue lets the model name the world itself (Figure~\ref{fig:method}). A first call shows the image, action,
question and options and asks, without answering, for the world shown:
\par\smallskip{\small\ttfamily\raggedright\noindent Do not answer the question. Return JSON with fields: world: one of
"real\_world", "minecraft\_vanilla", "minecraft\_modded", "other\_game", "unknown"; game\_name; objects.\par}
\smallskip\noindent If the model names vanilla Minecraft, a second call asks the told prompt with the tail of chain
of thought; otherwise it answers with chain of thought alone. On the items where the model names the game,
self-named world therefore equals Told + chain of thought (Table~\ref{tab:arxiv_self}).

The last cue adds knowledge to the selected world. Once per model and without any test item, the model lists how
vanilla Minecraft departs from real-world physics, five times for each of five topics (light and heat, gravity and
support, fluids, plants, materials and reactions). It merges the lists, keeping departures named in at least three of
five, and audits every entry with its reasoning switched on, correcting or deleting entries it judges wrong. The
audited cards hold \nCardDS{} (DeepSeek), \nCardSol{} (GPT-6.1-Sol) and \nCardQw{} (Qwen3.8-Flash) entries. The second
call places the card in the system message under the header
\par\smallskip{\small\ttfamily\raggedright\noindent Notes on how vanilla Minecraft differs from real-world physics. Use
a note only if it applies to the image.\par}
\smallskip\noindent and asks the model to name the applicable note before answering.
The audit, the header and the instruction to name the note were set with DeepSeek on the first main-set batch and then
fixed; the other two models use the same prompts with their own cards.

For the two models with incomplete rule knowledge the notes add conflict accuracy: over the video-game prompt they gain
\aGDCvsFixedGameHCDS{} \aGDCvsFixedGameHCCIDS{} points for DeepSeek and \aGDCvsFixedGameHCQF{} \aGDCvsFixedGameHCCIQF{}
for Qwen3.8-Flash on hard passive rules, and on the main set DeepSeek is best on both kinds of rule. GPT-6.1-Sol, which
needs no notes, is unchanged. A card is a list of departures, and on the hard set, whose consistent rules are processes
the game does simulate, its notes are stretched beyond their scope: consistent accuracy trails direct answering by
\aGDCvsDirectHSDS{} \aGDCvsDirectHSCIDS{} for DeepSeek and \aGDCvsDirectHSQF{} \aGDCvsDirectHSCIQF{} for Qwen3.8-Flash.
Replaying DeepSeek's answers recovers the note it named: ``plants left alone do not wilt'' applied to ripening cocoa
pods and a growing pumpkin, ``most blocks stay in the air when their support is removed'' applied to a hanging
stalactite, and ``heat does not spread between blocks'' applied to ice next to a campfire.

\subsection{Retest on the Main Set}
\label{app:arxiv_retest}

Knowing a rule can be tested out of sample. On items a model answers correctly in every text run and in two told runs,
a third told run errs on at most \nRevStableRetestMax\% and the bare image on
\nRevStableImageMin--\nRevStableImageMax\%.

\subsection{Open Models: Layer Choice and Steering}
\label{app:arxiv_wb}

\begin{table}[t]
\centering\small
\setlength{\tabcolsep}{4pt}
\begin{tabular}{r cc r cc}
\toprule
\multicolumn{3}{c}{Qwen3.5-9B} & \multicolumn{3}{c}{InternVL3.5-8B} \\
\cmidrule(lr){1-3}\cmidrule(lr){4-6}
Layer & Every & Answer & Layer & Every & Answer \\
\midrule
none & 62.5 & -- & none & 28.1 & -- \\
8 & 57.0 & 63.3 & 9 & 30.5 & 28.9 \\
10 & 66.4 & 60.2 & 11 & 36.7 & 28.9 \\
12 & 69.5 & 59.4 & 13 & 35.9 & 29.7 \\
14 & 60.9 & 60.2 & 15 & 42.2 & 29.7 \\
16 & 66.4 & 60.2 & 17 & 38.3 & 29.7 \\
18 & 65.6 & 61.7 & 19 & 45.3 & 30.5 \\
20 & 60.9 & 60.9 & 21 & 42.2 & 29.7 \\
22 & 57.8 & 60.9 & 23 & 36.7 & 30.5 \\
24 & 54.7 & 60.9 & 25 & 30.5 & 30.5 \\
26 & 60.2 & 60.9 & 27 & 25.0 & 29.7 \\
28 & 58.6 & 61.7 & 29 & 31.2 & 29.7 \\
30 & 58.6 & 60.2 & 31 & 30.5 & 30.5 \\
 &  &  & 33 & 28.9 & 29.7 \\
 &  &  & 35 & 29.7 & 29.7 \\
\bottomrule
\end{tabular}
\caption{Choice of layer on the dev rules: balanced accuracy (\%) of bare-image answers with the direction added at strength 3, at every position or at the answer position only. Strengths 1 and 2 give lower maxima.}
\label{tab:wb_sweep}
\end{table}
\begin{table}[t]
\centering\small
\setlength{\tabcolsep}{0pt}
\begin{tabular*}{\columnwidth}{@{\extracolsep{\fill}}l ccc ccc@{}}
\toprule
 & \multicolumn{3}{c}{Qwen3.5-9B} & \multicolumn{3}{c}{InternVL3.5-8B} \\
\cmidrule(lr){2-4}\cmidrule(l){5-7}
 & Conf. & Cons. & Bal. & Conf. & Cons. & Bal. \\
\midrule
Image & 31.9 & 74.0 & 53.0 & 13.6 & 78.8 & 46.2 \\
\quad + direction & 54.0 & 46.2 & 50.1 & 24.7 & 95.2 & 60.0 \\
\quad + random & 8.0 & 88.5 & 48.2 & 25.0 & 75.0 & 50.0 \\
\midrule
Told & 55.3 & 51.0 & 53.1 & 25.0 & 73.1 & 49.0 \\
\quad $-$ direction & 37.0 & 66.3 & 51.7 & 27.1 & 39.4 & 33.3 \\
\bottomrule
\end{tabular*}
\caption{Steering the open models on the hard set (accuracy, \%). The direction, the mean told minus bare-image state
of the development rules, is added at every position of one layer; the random direction has the same norm.}
\label{tab:wb_steer_hard}
\end{table}

Table~\ref{tab:wb_sweep} gives the choice of layer and strength on the development rules (Appendix~\ref{app:whitebox}),
and Table~\ref{tab:wb_steer_hard} the hard set with the chosen setting. Added at the answer position only, the direction
barely moves the answers at any layer, in line with the patching result. For InternVL3.5-8B, subtracting the direction
from the told prompt lowers consistent rules more than conflict ones.

\subsection{Human Study}
\label{app:arxiv_human}

The \nHumanN{} participants were undergraduates. Their reported play time is \nHumanHoursText. A
questionnaire on Minecraft experience preceded the questions, so participants knew which game the images came from but
were not asked to apply its rules; their accuracy therefore compares with the told condition of the models.

\subsection{Case Studies}
\label{app:cases}

\begin{figure*}[p]
\centering
\setlength{\tabcolsep}{0pt}\setlength{\parskip}{0pt}
\begin{tabular*}{\textwidth}{@{\extracolsep{\fill}}>{\footnotesize\raggedright\arraybackslash}p{0.315\textwidth}>{\footnotesize\raggedright\arraybackslash}p{0.315\textwidth}>{\footnotesize\raggedright\arraybackslash}p{0.315\textwidth}@{}}
\normalsize\textbf{Minecraft} & \normalsize\textbf{Terraria} & \normalsize\textbf{Stardew Valley} \\[1pt]
\includegraphics[width=\linewidth]{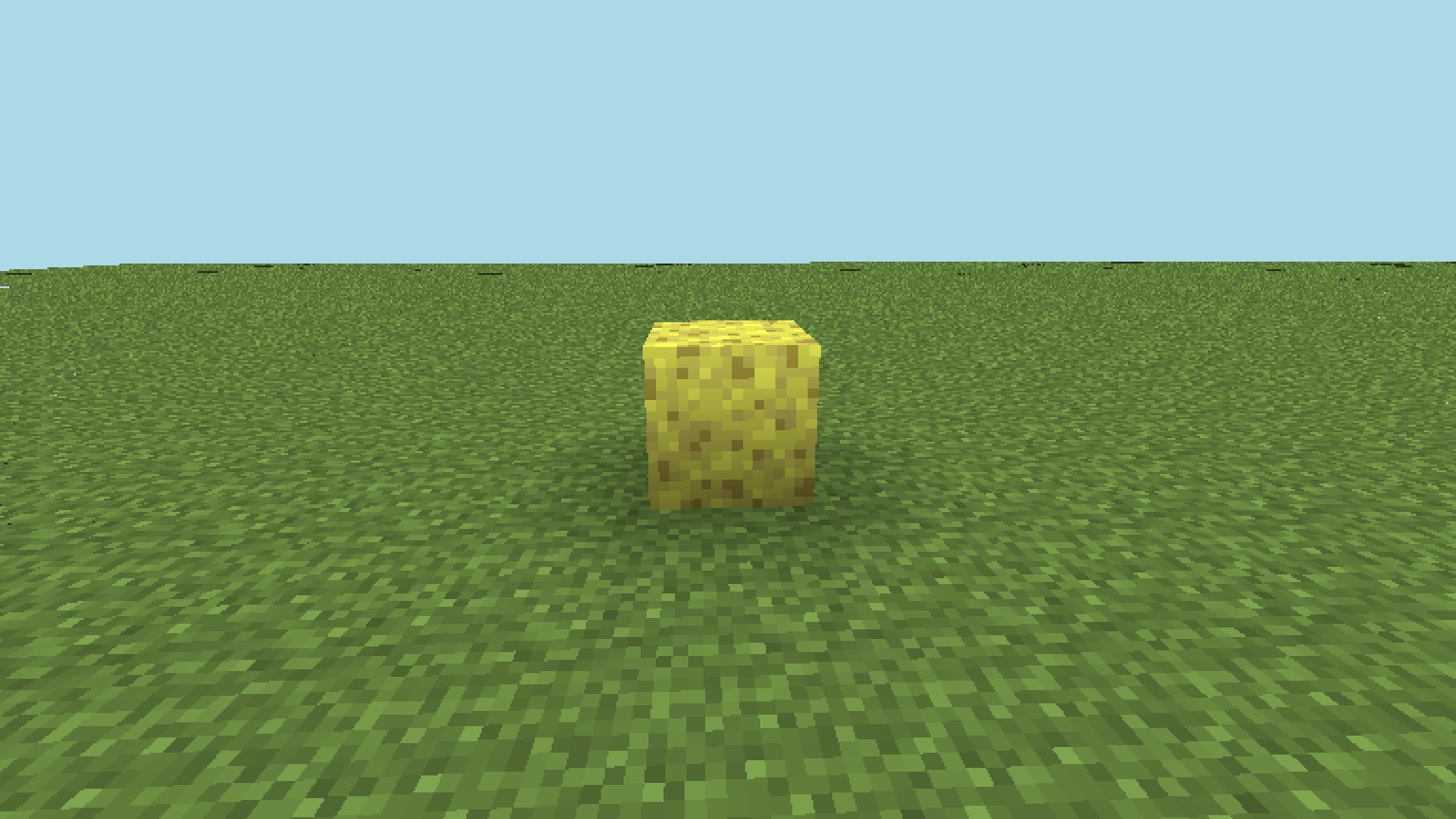} & \includegraphics[width=\linewidth]{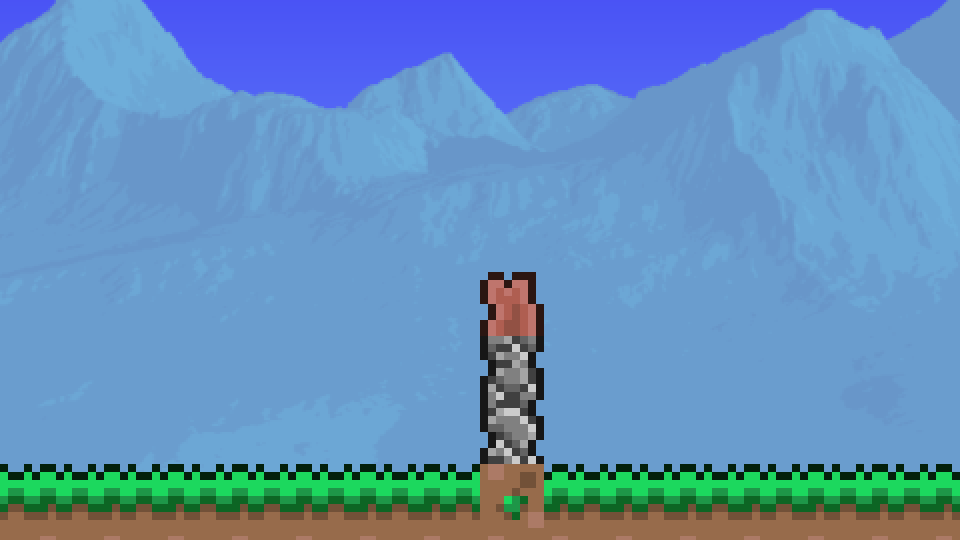} & \includegraphics[width=\linewidth]{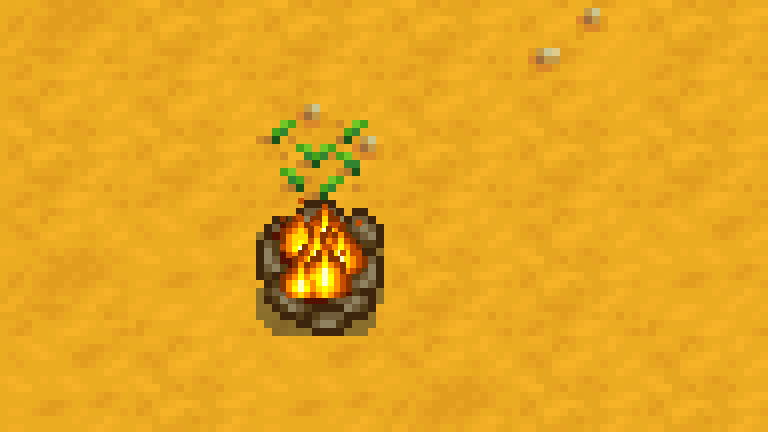} \\[1pt]
Heavy rain falls on the dry sponge all day. & After the grey stone block directly below the clay block is removed, what happens to the clay block? & The campfire is burning. Nobody adds any wood, and a week passes. What does it look like at the end of the week? \\[2pt]
\begin{tabular}[t]{@{}p{1.1em}@{}>{\raggedright\arraybackslash}p{\dimexpr\linewidth-1.1em\relax}@{}}
\caseok & \textbf{Game:} The sponge stays completely dry, exactly as it was. \\[1pt]
 & \textbf{Real world:} The sponge has soaked up the rain and is now heavy and wet. \\[1pt]
 & \textbf{Neither:} The sponge has shrunk to half its size. \\[1pt]
\end{tabular}\par\vspace{4pt} & \begin{tabular}[t]{@{}p{1.1em}@{}>{\raggedright\arraybackslash}p{\dimexpr\linewidth-1.1em\relax}@{}}
\caseok & \textbf{Game:} It stays exactly where it is. \\[1pt]
 & \textbf{Real world:} It immediately falls down one level. \\[1pt]
 & \textbf{Neither:} It slowly slides sideways. \\[1pt]
\end{tabular}\par\vspace{4pt} & \begin{tabular}[t]{@{}p{1.1em}@{}>{\raggedright\arraybackslash}p{\dimexpr\linewidth-1.1em\relax}@{}}
\caseok & \textbf{Game:} It is still burning, just as now. \\[1pt]
 & \textbf{Real world:} It has burned out, leaving a pile of ash. \\[1pt]
 & \textbf{Neither:} It has spread into a large blaze. \\[1pt]
\end{tabular}\par\vspace{4pt} \\
\centering \begin{tabular}{@{}l@{\hspace{3pt}}c@{\hspace{3pt}}c@{\hspace{3pt}}c@{}}
& \scriptsize DeepSeek & \scriptsize Qwen-Max & \scriptsize Doubao-Pro \\ \midrule
Text & \caseok & \caseok & \caseok \\
Told & \caseok & \caseok & \caseok \\
Image & \casereal & \casereal & \casereal \\
\end{tabular} & \centering \begin{tabular}{@{}l@{\hspace{3pt}}c@{\hspace{3pt}}c@{\hspace{3pt}}c@{}}
& \scriptsize DeepSeek & \scriptsize Qwen-Max & \scriptsize Doubao-Pro \\ \midrule
Text & \caseok & \caseok & \caseok \\
Told & \caseok & \caseok & \caseok \\
Image & \casereal & \casereal & \casereal \\
\end{tabular} & \centering \begin{tabular}{@{}l@{\hspace{3pt}}c@{\hspace{3pt}}c@{\hspace{3pt}}c@{}}
& \scriptsize DeepSeek & \scriptsize Qwen-Max & \scriptsize Doubao-Pro \\ \midrule
Text & \caseok & \caseok & \caseok \\
Told & \caseok & \caseok & \caseok \\
Image & \casereal & \casereal & \casereal \\
\end{tabular} \\
\end{tabular*}
\caption{One conflict rule per game: the answers of three models in text, when told and on the bare image (first run).
For Minecraft the action is shown; the shared question of Figure~\ref{fig:app_prompts} follows it. \caseok: correct;
\casereal: the real-world outcome.}
\label{fig:cases}
\end{figure*}

\begin{figure*}[p]
\centering
\begin{minipage}[c]{0.19\textwidth}
\includegraphics[width=\linewidth]{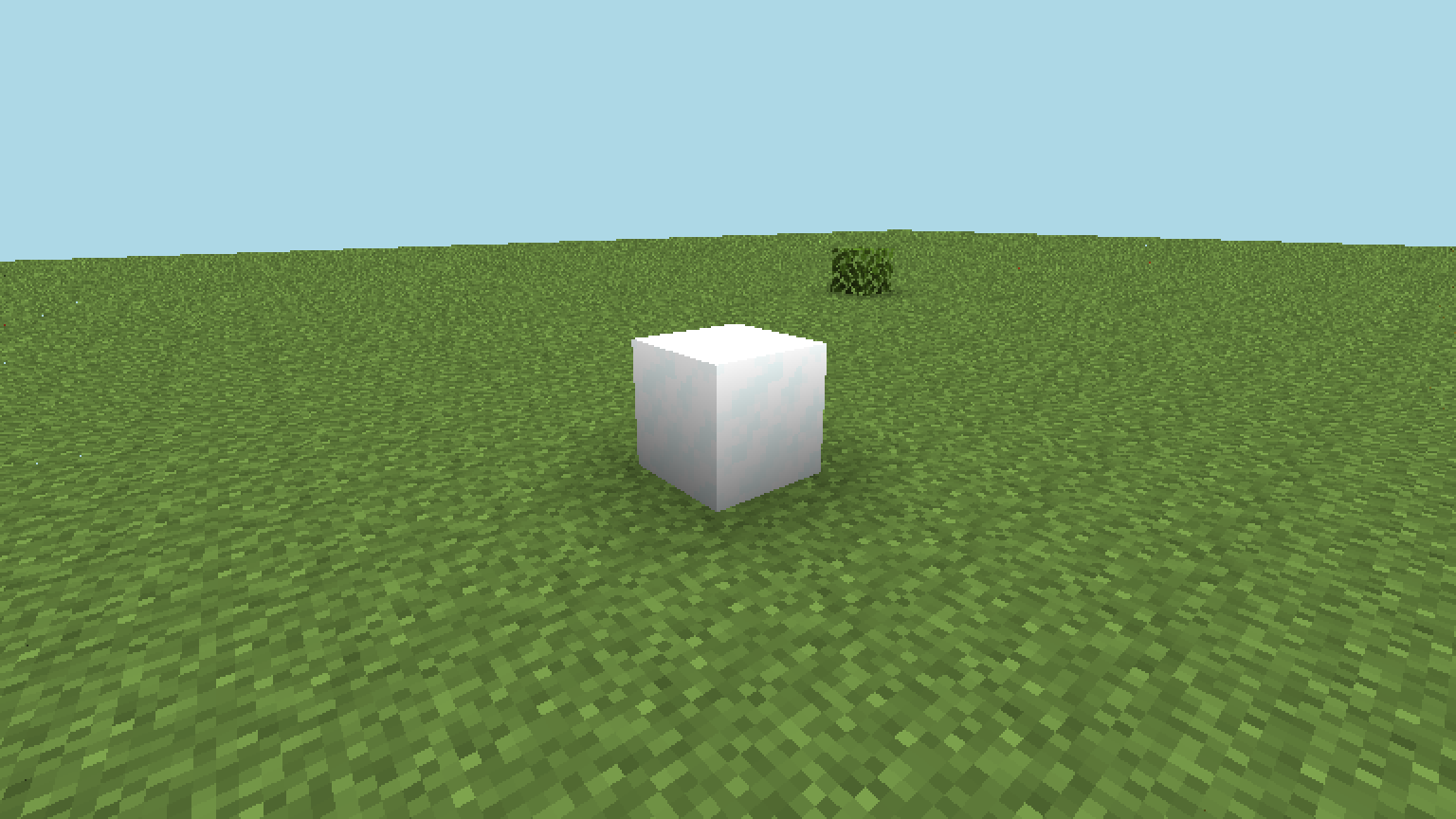}
\end{minipage}\hfill
\begin{minipage}[c]{0.47\textwidth}
\footnotesize\raggedright {\scriptsize\textsc{conflict, time or weather only (hard set)}}\par\vspace{1pt}
Several days of warm rain fall on the snow block. Nobody touches it.\par\vspace{1pt}
\begin{tabular}[t]{@{}p{1.1em}@{}>{\raggedright\arraybackslash}p{\dimexpr\linewidth-1.1em\relax}@{}}
\caseok & \textbf{Game:} The snow block stays completely unchanged. \\[1pt]
 & \textbf{Real world:} The snow block has turned to slush and shrunk to a small heap. \\[1pt]
 & \textbf{Neither:} The snow block has grown into a taller pile of snow. \\[1pt]
\end{tabular}
\end{minipage}\hfill
\begin{minipage}[c]{0.31\textwidth}
{\footnotesize\centering \begin{tabular}{@{}l@{\hspace{3pt}}c@{\hspace{3pt}}c@{\hspace{3pt}}c@{}}
& \scriptsize DeepSeek & \scriptsize Qwen-Max & \scriptsize Doubao-Pro \\ \midrule
Text & \caseok & \caseok & \caseok \\
Told & \caseok & \caseok & \caseok \\
Image & \casereal & \casereal & \casereal \\
\end{tabular}\par}
\end{minipage}

\vspace{4pt}\hrule\vspace{4pt}

\begin{minipage}[c]{0.19\textwidth}
\includegraphics[width=\linewidth]{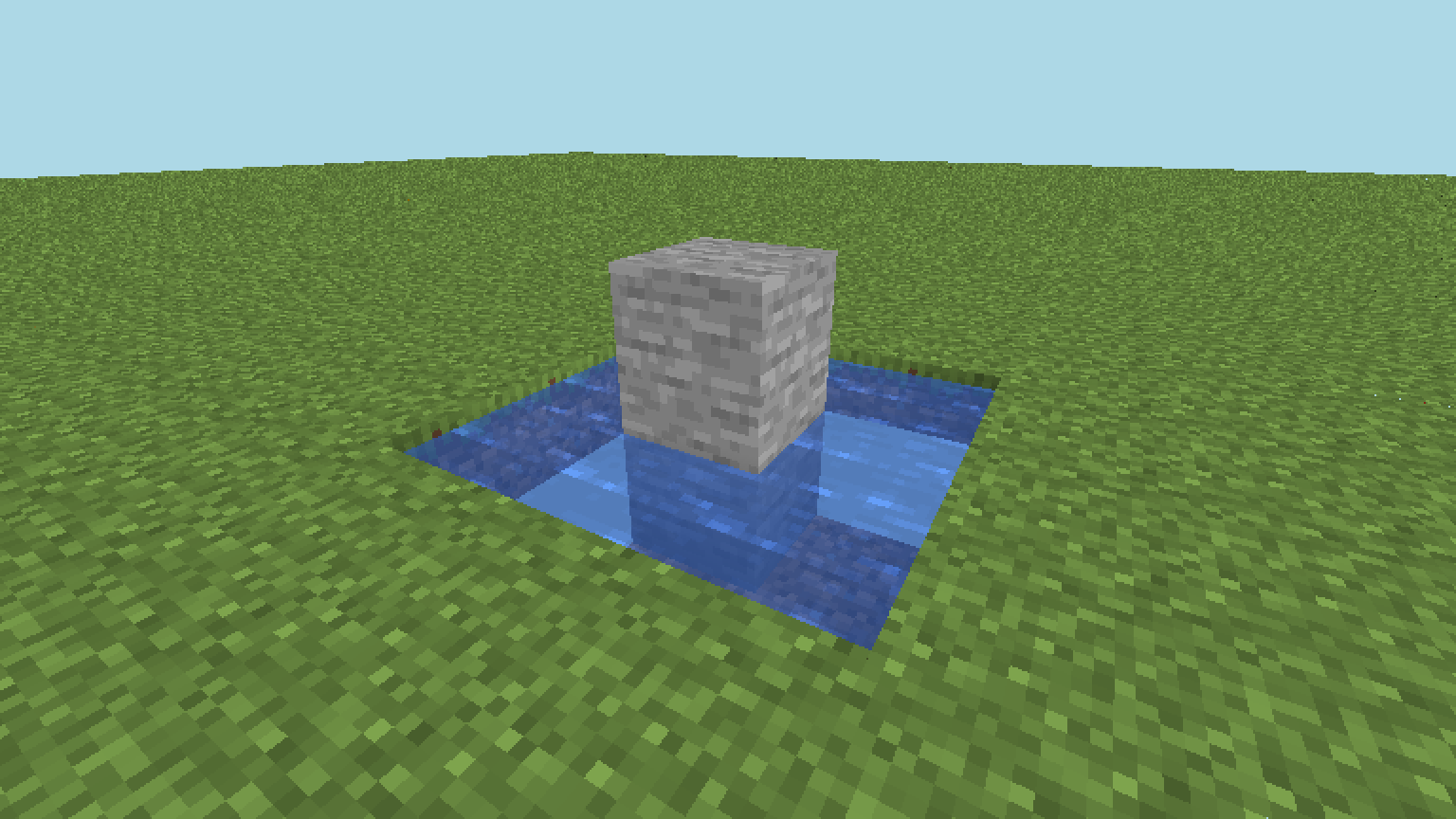}
\end{minipage}\hfill
\begin{minipage}[c]{0.47\textwidth}
\footnotesize\raggedright {\scriptsize\textsc{conflict, time or weather only (main set)}}\par\vspace{1pt}
Many years of rain pass with the stone standing in the water.\par\vspace{1pt}
\begin{tabular}[t]{@{}p{1.1em}@{}>{\raggedright\arraybackslash}p{\dimexpr\linewidth-1.1em\relax}@{}}
\caseok & \textbf{Game:} The stone looks exactly the same as it did at the start. \\[1pt]
 & \textbf{Real world:} The stone has become covered in green moss. \\[1pt]
 & \textbf{Neither:} The stone has turned into iron ore. \\[1pt]
\end{tabular}
\end{minipage}\hfill
\begin{minipage}[c]{0.31\textwidth}
{\footnotesize\centering \begin{tabular}{@{}l@{\hspace{3pt}}c@{\hspace{3pt}}c@{\hspace{3pt}}c@{}}
& \scriptsize DeepSeek & \scriptsize Qwen-Max & \scriptsize Doubao-Pro \\ \midrule
Text & \caseok & \caseok & \caseok \\
Told & \caseok & \caseok & \caseok \\
Image & \casereal & \casereal & \casereal \\
\end{tabular}\par}
\end{minipage}

\vspace{4pt}\hrule\vspace{4pt}

\begin{minipage}[c]{0.19\textwidth}
\includegraphics[width=\linewidth]{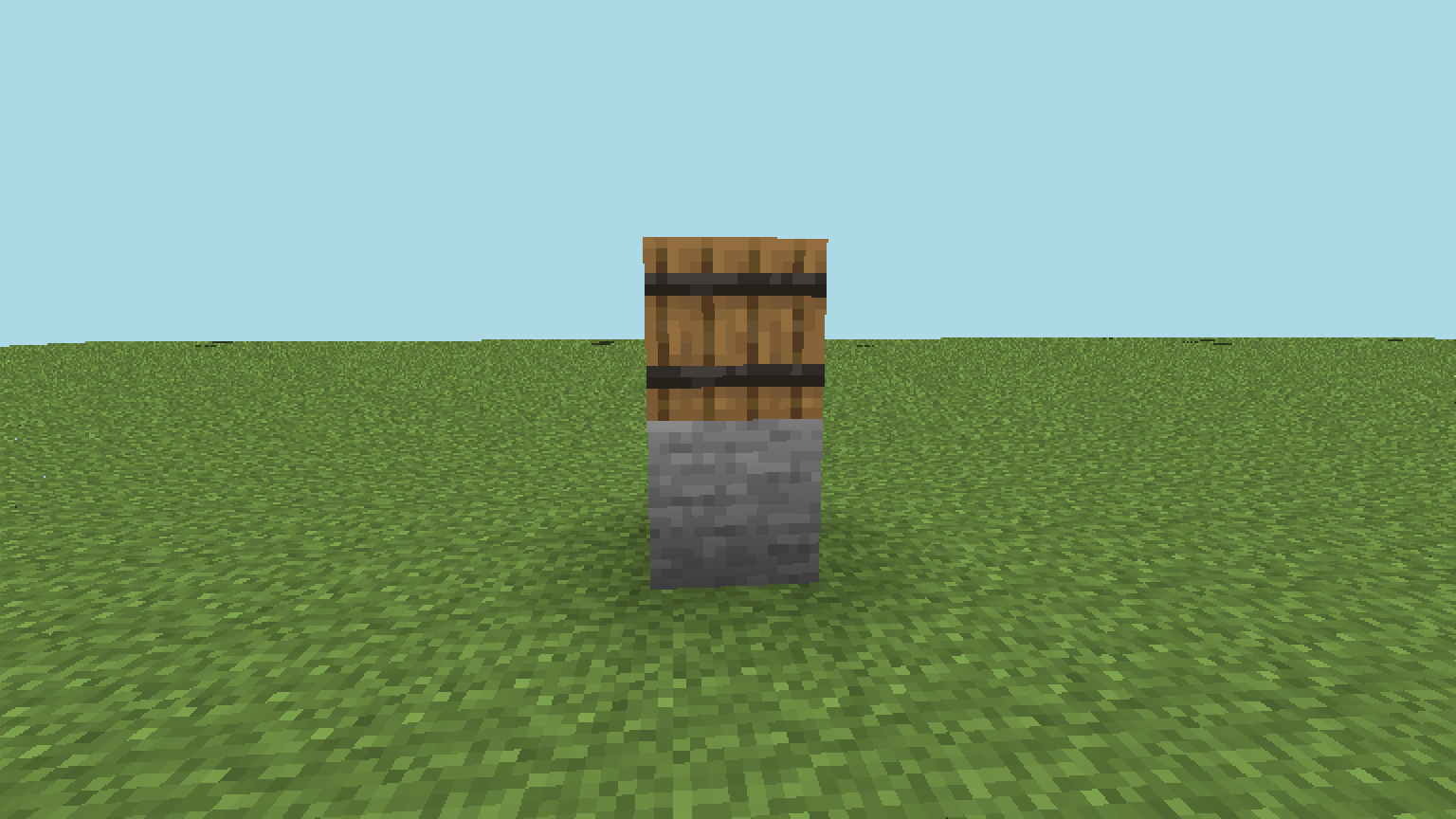}
\end{minipage}\hfill
\begin{minipage}[c]{0.47\textwidth}
\footnotesize\raggedright {\scriptsize\textsc{conflict, with an action (hard set)}}\par\vspace{1pt}
The stone block under the barrel is broken and removed.\par\vspace{1pt}
\begin{tabular}[t]{@{}p{1.1em}@{}>{\raggedright\arraybackslash}p{\dimexpr\linewidth-1.1em\relax}@{}}
\caseok & \textbf{Game:} The barrel stays exactly where it was, floating in mid-air. \\[1pt]
 & \textbf{Real world:} The barrel falls down onto the grass. \\[1pt]
 & \textbf{Neither:} The barrel breaks into a pile of wooden planks. \\[1pt]
\end{tabular}
\end{minipage}\hfill
\begin{minipage}[c]{0.31\textwidth}
{\footnotesize\centering \begin{tabular}{@{}l@{\hspace{3pt}}c@{\hspace{3pt}}c@{\hspace{3pt}}c@{}}
& \scriptsize DeepSeek & \scriptsize Qwen-Max & \scriptsize Doubao-Pro \\ \midrule
Text & \caseok & \caseok & \caseok \\
Told & \casereal & \caseok & \casereal \\
Image & \casereal & \caseok & \casereal \\
\end{tabular}\par}
\end{minipage}

\vspace{4pt}\hrule\vspace{4pt}

\begin{minipage}[c]{0.19\textwidth}
\includegraphics[width=\linewidth]{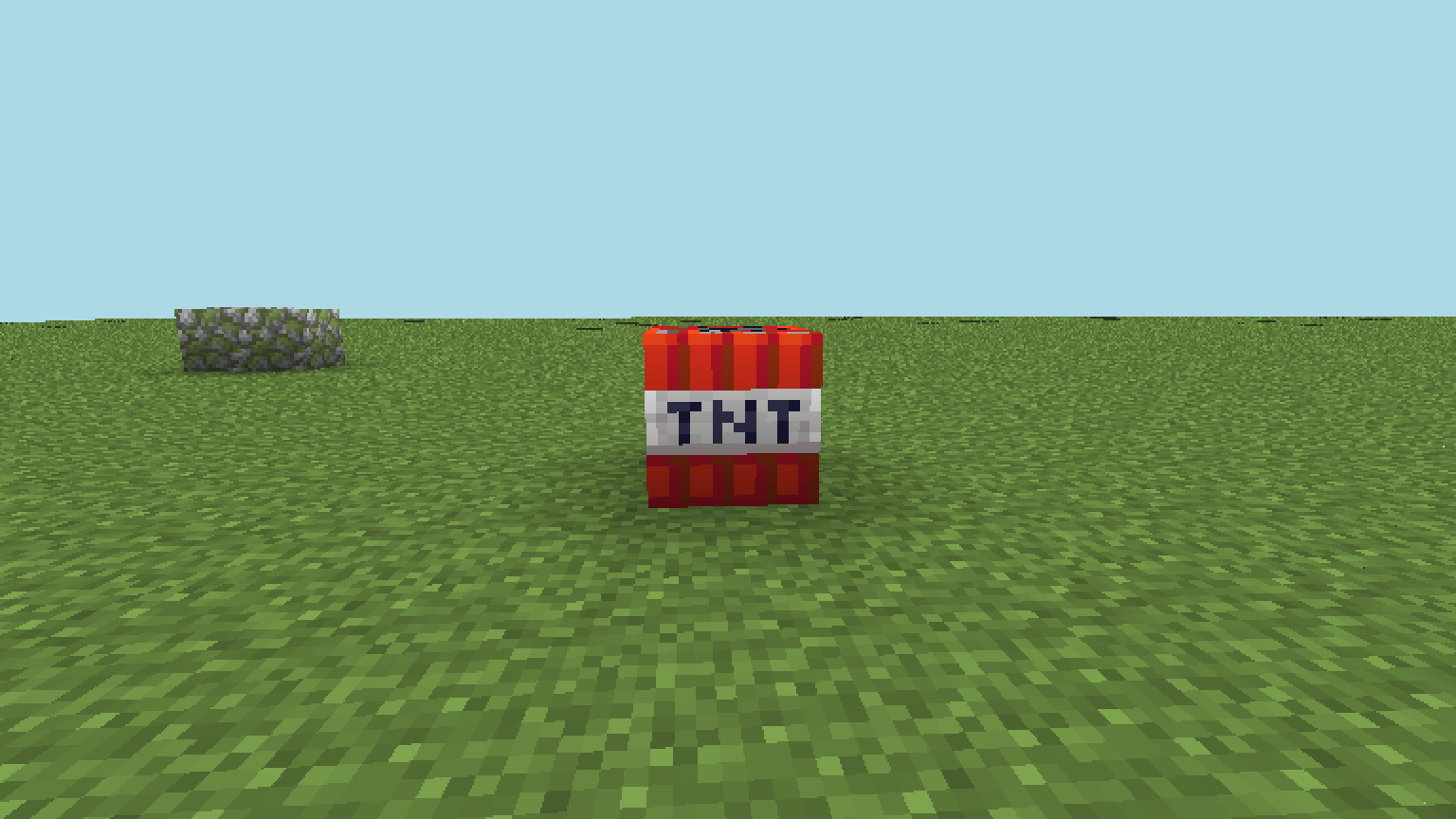}
\end{minipage}\hfill
\begin{minipage}[c]{0.47\textwidth}
\footnotesize\raggedright {\scriptsize\textsc{consistent (main set)}}\par\vspace{1pt}
The TNT block is lit with a flint and steel. A few seconds pass.\par\vspace{1pt}
\begin{tabular}[t]{@{}p{1.1em}@{}>{\raggedright\arraybackslash}p{\dimexpr\linewidth-1.1em\relax}@{}}
\caseok & \textbf{Both worlds:} The TNT explodes and leaves a crater in the ground. \\[1pt]
 & \textbf{Exception:} The TNT explodes, but the ground around it is not damaged at all. \\[1pt]
 & \textbf{Neither:} The TNT turns into a block of sand. \\[1pt]
\end{tabular}
\end{minipage}\hfill
\begin{minipage}[c]{0.31\textwidth}
{\footnotesize\centering \begin{tabular}{@{}l@{\hspace{3pt}}c@{\hspace{3pt}}c@{\hspace{3pt}}c@{}}
& \scriptsize DeepSeek & \scriptsize Qwen-Max & \scriptsize Doubao-Pro \\ \midrule
Text & \caseok & \caseok & \caseok \\
Told & \caseok & \caseok & \caseok \\
Image & \caseok & \caseok & \caseok \\
\end{tabular}\par}
\end{minipage}
\caption{Case studies from Minecraft: the image, the action, which the shared question of Figure~\ref{fig:app_prompts} follows, the options, and the first-run answers of three models;
above each question, the type of the rule. \caseok: correct; otherwise the kind of outcome chosen.}
\label{fig:cases-minecraft}
\end{figure*}

\begin{figure*}[p]
\centering
\begin{minipage}[c]{0.19\textwidth}
\includegraphics[width=\linewidth]{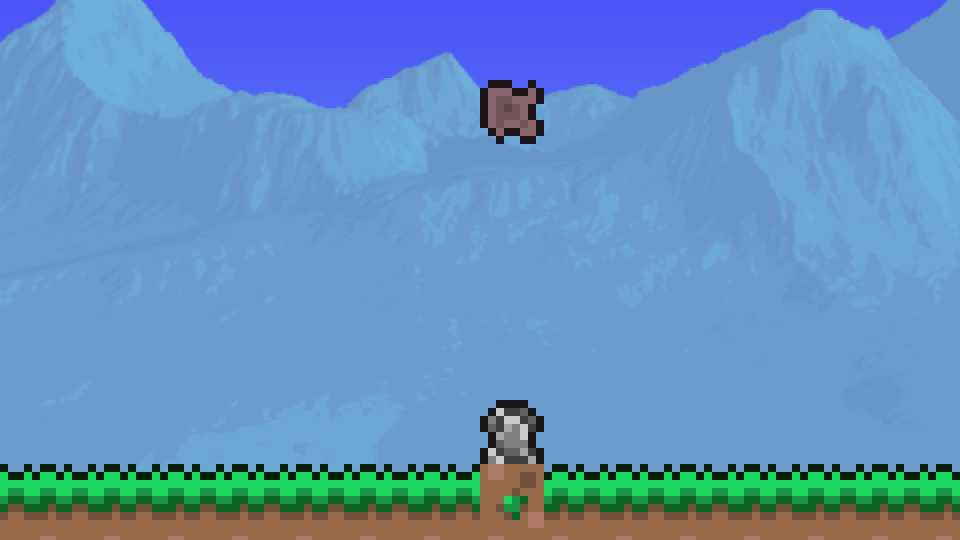}
\end{minipage}\hfill
\begin{minipage}[c]{0.47\textwidth}
\footnotesize\raggedright {\scriptsize\textsc{conflict, time or weather only}}\par\vspace{1pt}
After some time passes, what happens to the reddish clay block hanging in the air?\par\vspace{1pt}
\begin{tabular}[t]{@{}p{1.1em}@{}>{\raggedright\arraybackslash}p{\dimexpr\linewidth-1.1em\relax}@{}}
\caseok & \textbf{Game:} It stays exactly where it is. \\[1pt]
 & \textbf{Real world:} It falls and breaks apart on impact. \\[1pt]
 & \textbf{Neither:} It slowly melts away. \\[1pt]
\end{tabular}
\end{minipage}\hfill
\begin{minipage}[c]{0.31\textwidth}
{\footnotesize\centering \begin{tabular}{@{}l@{\hspace{3pt}}c@{\hspace{3pt}}c@{\hspace{3pt}}c@{}}
& \scriptsize DeepSeek & \scriptsize Qwen-Max & \scriptsize Doubao-Pro \\ \midrule
Text & \caseok & \caseok & \caseok \\
Told & \caseok & \caseok & \caseok \\
Image & \casereal & \casereal & \casereal \\
\end{tabular}\par}
\end{minipage}

\vspace{4pt}\hrule\vspace{4pt}

\begin{minipage}[c]{0.19\textwidth}
\includegraphics[width=\linewidth]{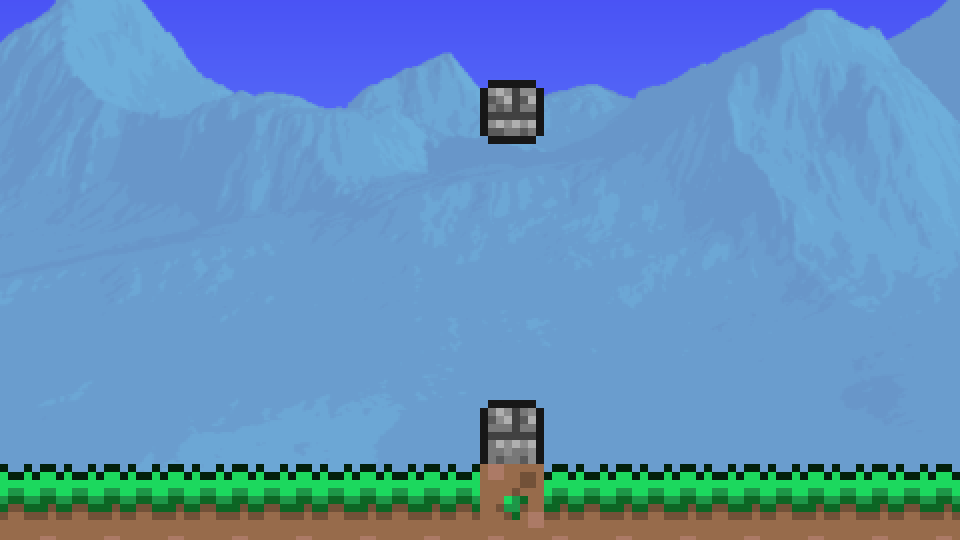}
\end{minipage}\hfill
\begin{minipage}[c]{0.47\textwidth}
\footnotesize\raggedright {\scriptsize\textsc{conflict, time or weather only}}\par\vspace{1pt}
After some time passes, what happens to the grey block floating in mid-air?\par\vspace{1pt}
\begin{tabular}[t]{@{}p{1.1em}@{}>{\raggedright\arraybackslash}p{\dimexpr\linewidth-1.1em\relax}@{}}
\caseok & \textbf{Game:} It stays exactly where it is. \\[1pt]
 & \textbf{Real world:} It falls straight down. \\[1pt]
 & \textbf{Neither:} It floats upward toward the surface. \\[1pt]
\end{tabular}
\end{minipage}\hfill
\begin{minipage}[c]{0.31\textwidth}
{\footnotesize\centering \begin{tabular}{@{}l@{\hspace{3pt}}c@{\hspace{3pt}}c@{\hspace{3pt}}c@{}}
& \scriptsize DeepSeek & \scriptsize Qwen-Max & \scriptsize Doubao-Pro \\ \midrule
Text & \caseok & \caseok & \casereal \\
Told & \caseok & \caseok & \caseok \\
Image & \casereal & \casereal & \casereal \\
\end{tabular}\par}
\end{minipage}

\vspace{4pt}\hrule\vspace{4pt}

\begin{minipage}[c]{0.19\textwidth}
\includegraphics[width=\linewidth]{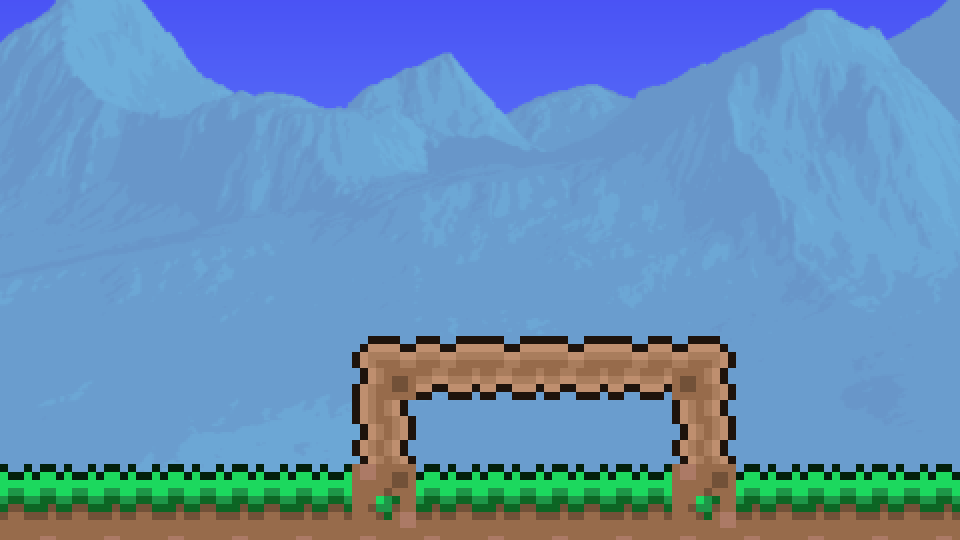}
\end{minipage}\hfill
\begin{minipage}[c]{0.47\textwidth}
\footnotesize\raggedright {\scriptsize\textsc{conflict, with an action}}\par\vspace{1pt}
After the rightmost supporting block is removed, leaving five blocks overhanging in mid-air, what happens?\par\vspace{1pt}
\begin{tabular}[t]{@{}p{1.1em}@{}>{\raggedright\arraybackslash}p{\dimexpr\linewidth-1.1em\relax}@{}}
\caseok & \textbf{Game:} All five blocks stay where they are. \\[1pt]
 & \textbf{Real world:} The unsupported end collapses. \\[1pt]
 & \textbf{Neither:} The whole row slides to the right. \\[1pt]
\end{tabular}
\end{minipage}\hfill
\begin{minipage}[c]{0.31\textwidth}
{\footnotesize\centering \begin{tabular}{@{}l@{\hspace{3pt}}c@{\hspace{3pt}}c@{\hspace{3pt}}c@{}}
& \scriptsize DeepSeek & \scriptsize Qwen-Max & \scriptsize Doubao-Pro \\ \midrule
Text & \caseok & \caseok & \caseok \\
Told & \caseok & \caseok & \caseok \\
Image & \casereal & \casereal & \casereal \\
\end{tabular}\par}
\end{minipage}

\vspace{4pt}\hrule\vspace{4pt}

\begin{minipage}[c]{0.19\textwidth}
\includegraphics[width=\linewidth]{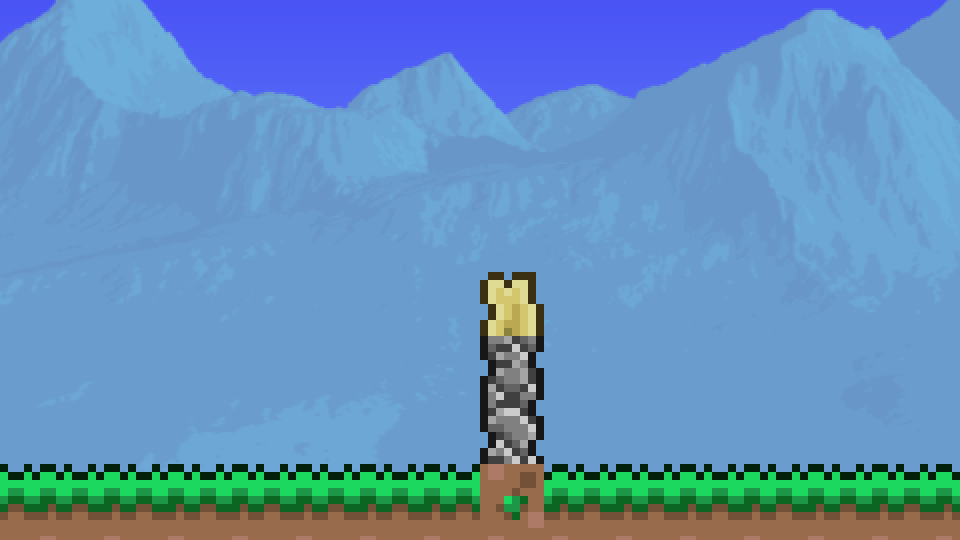}
\end{minipage}\hfill
\begin{minipage}[c]{0.47\textwidth}
\footnotesize\raggedright {\scriptsize\textsc{consistent}}\par\vspace{1pt}
After the stone block below the sand is removed, what happens to the sand?\par\vspace{1pt}
\begin{tabular}[t]{@{}p{1.1em}@{}>{\raggedright\arraybackslash}p{\dimexpr\linewidth-1.1em\relax}@{}}
\caseok & \textbf{Both worlds:} It falls straight down. \\[1pt]
 & \textbf{Exception:} It stays floating in mid-air. \\[1pt]
 & \textbf{Neither:} It spreads outward in all directions. \\[1pt]
\end{tabular}
\end{minipage}\hfill
\begin{minipage}[c]{0.31\textwidth}
{\footnotesize\centering \begin{tabular}{@{}l@{\hspace{3pt}}c@{\hspace{3pt}}c@{\hspace{3pt}}c@{}}
& \scriptsize DeepSeek & \scriptsize Qwen-Max & \scriptsize Doubao-Pro \\ \midrule
Text & \caseok & \caseok & \caseok \\
Told & \caseok & \caseok & \caseok \\
Image & \caseok & \caseok & \caseok \\
\end{tabular}\par}
\end{minipage}
\caption{Case studies from Terraria: the image, the question as asked, the options, and the first-run answers of three models;
above each question, the type of the rule. \caseok: correct; otherwise the kind of outcome chosen.}
\label{fig:cases-terraria}
\end{figure*}

\begin{figure*}[p]
\centering
\begin{minipage}[c]{0.19\textwidth}
\includegraphics[width=\linewidth]{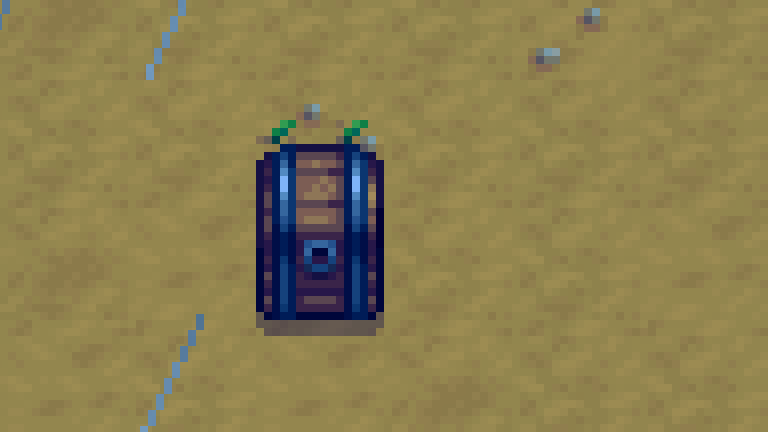}
\end{minipage}\hfill
\begin{minipage}[c]{0.47\textwidth}
\footnotesize\raggedright {\scriptsize\textsc{conflict, time or weather only}}\par\vspace{1pt}
The wooden chest stands outdoors and it rains every day for two weeks. What does the chest look like at the end?\par\vspace{1pt}
\begin{tabular}[t]{@{}p{1.1em}@{}>{\raggedright\arraybackslash}p{\dimexpr\linewidth-1.1em\relax}@{}}
\caseok & \textbf{Game:} It looks exactly the same as now. \\[1pt]
 & \textbf{Real world:} The wood has darkened and started to swell and warp. \\[1pt]
 & \textbf{Neither:} It has sunk into the ground. \\[1pt]
\end{tabular}
\end{minipage}\hfill
\begin{minipage}[c]{0.31\textwidth}
{\footnotesize\centering \begin{tabular}{@{}l@{\hspace{3pt}}c@{\hspace{3pt}}c@{\hspace{3pt}}c@{}}
& \scriptsize DeepSeek & \scriptsize Qwen-Max & \scriptsize Doubao-Pro \\ \midrule
Text & \caseok & \caseok & \caseok \\
Told & \caseok & \caseok & \caseok \\
Image & \casereal & \casereal & \casereal \\
\end{tabular}\par}
\end{minipage}

\vspace{4pt}\hrule\vspace{4pt}

\begin{minipage}[c]{0.19\textwidth}
\includegraphics[width=\linewidth]{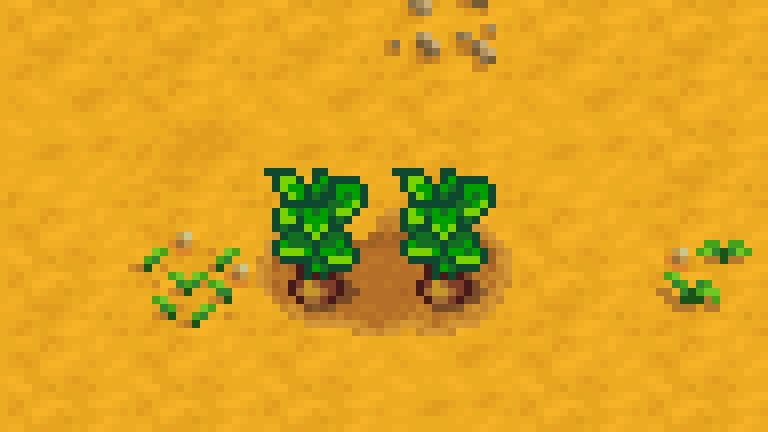}
\end{minipage}\hfill
\begin{minipage}[c]{0.47\textwidth}
\footnotesize\raggedright {\scriptsize\textsc{conflict, time or weather only}}\par\vspace{1pt}
The vegetables in the image are ripe. Nobody picks them, and a week of sunny weather passes. What do they look like at the end of the week?\par\vspace{1pt}
\begin{tabular}[t]{@{}p{1.1em}@{}>{\raggedright\arraybackslash}p{\dimexpr\linewidth-1.1em\relax}@{}}
\caseok & \textbf{Game:} They are still ripe and ready to pick, just as now. \\[1pt]
 & \textbf{Real world:} They become overripe and start to rot in the ground. \\[1pt]
 & \textbf{Neither:} They shrink back into small seedlings. \\[1pt]
\end{tabular}
\end{minipage}\hfill
\begin{minipage}[c]{0.31\textwidth}
{\footnotesize\centering \begin{tabular}{@{}l@{\hspace{3pt}}c@{\hspace{3pt}}c@{\hspace{3pt}}c@{}}
& \scriptsize DeepSeek & \scriptsize Qwen-Max & \scriptsize Doubao-Pro \\ \midrule
Text & \caseok & \caseok & \caseok \\
Told & \caseok & \caseok & \caseok \\
Image & \casereal & \casereal & \casereal \\
\end{tabular}\par}
\end{minipage}

\vspace{4pt}\hrule\vspace{4pt}

\begin{minipage}[c]{0.19\textwidth}
\includegraphics[width=\linewidth]{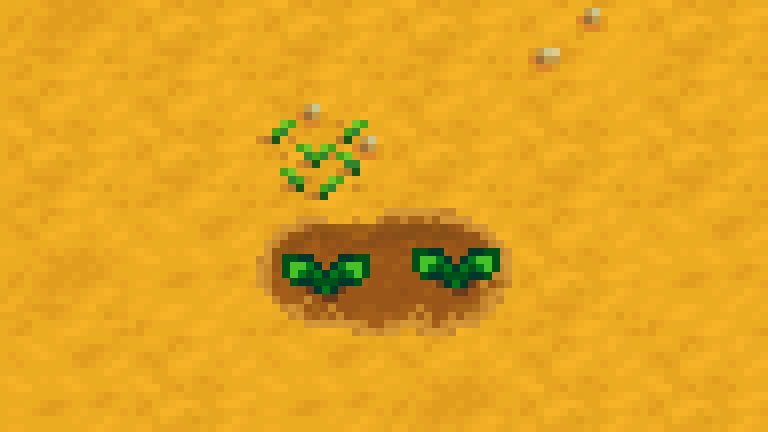}
\end{minipage}\hfill
\begin{minipage}[c]{0.47\textwidth}
\footnotesize\raggedright {\scriptsize\textsc{conflict, with an action}}\par\vspace{1pt}
Someone waters the seedlings in the tilled soil every morning for one week of sunny weather. What do they look like at the end of the week?\par\vspace{1pt}
\begin{tabular}[t]{@{}p{1.1em}@{}>{\raggedright\arraybackslash}p{\dimexpr\linewidth-1.1em\relax}@{}}
\caseok & \textbf{Game:} They are fully grown and ready to harvest. \\[1pt]
 & \textbf{Real world:} They are a little taller, but still young plants. \\[1pt]
 & \textbf{Neither:} They have dried out and died. \\[1pt]
\end{tabular}
\end{minipage}\hfill
\begin{minipage}[c]{0.31\textwidth}
{\footnotesize\centering \begin{tabular}{@{}l@{\hspace{3pt}}c@{\hspace{3pt}}c@{\hspace{3pt}}c@{}}
& \scriptsize DeepSeek & \scriptsize Qwen-Max & \scriptsize Doubao-Pro \\ \midrule
Text & \caseok & \caseok & \casereal \\
Told & \casereal & \casereal & \casereal \\
Image & \casereal & \casereal & \casereal \\
\end{tabular}\par}
\end{minipage}

\vspace{4pt}\hrule\vspace{4pt}

\begin{minipage}[c]{0.19\textwidth}
\includegraphics[width=\linewidth]{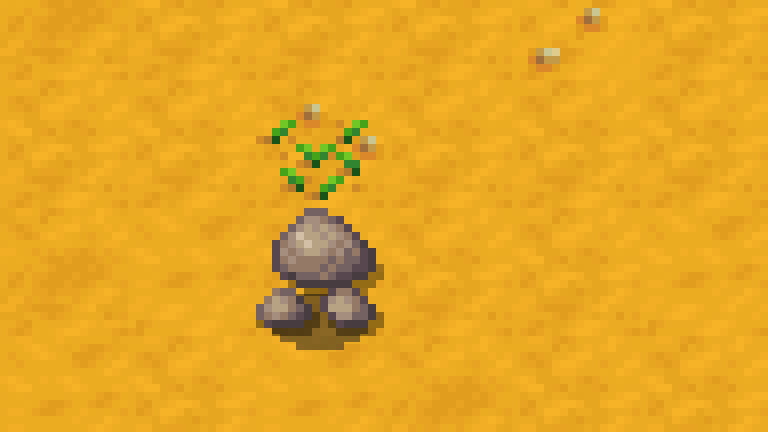}
\end{minipage}\hfill
\begin{minipage}[c]{0.47\textwidth}
\footnotesize\raggedright {\scriptsize\textsc{consistent}}\par\vspace{1pt}
Someone hits the rock with a pickaxe again and again. What happens to the rock?\par\vspace{1pt}
\begin{tabular}[t]{@{}p{1.1em}@{}>{\raggedright\arraybackslash}p{\dimexpr\linewidth-1.1em\relax}@{}}
\caseok & \textbf{Both worlds:} It breaks apart. \\[1pt]
 & \textbf{Exception:} It stays in place, without a scratch. \\[1pt]
 & \textbf{Neither:} It melts. \\[1pt]
\end{tabular}
\end{minipage}\hfill
\begin{minipage}[c]{0.31\textwidth}
{\footnotesize\centering \begin{tabular}{@{}l@{\hspace{3pt}}c@{\hspace{3pt}}c@{\hspace{3pt}}c@{}}
& \scriptsize DeepSeek & \scriptsize Qwen-Max & \scriptsize Doubao-Pro \\ \midrule
Text & \caseok & \caseok & \caseok \\
Told & \caseok & \caseok & \caseok \\
Image & \caseok & \caseok & \caseok \\
\end{tabular}\par}
\end{minipage}
\caption{Case studies from Stardew Valley: the image, the question as asked, the options, and the first-run answers of three models;
above each question, the type of the rule. \caseok: correct; otherwise the kind of outcome chosen.}
\label{fig:cases-stardew}
\end{figure*}

Figure~\ref{fig:cases} shows one conflict rule per game side by side, and
Figures~\ref{fig:cases-minecraft}--\ref{fig:cases-stardew} show four rules from each game: two conflict rules where only
time or weather acts, one with an action, and one consistent rule, each with one of its question images as the models
saw it and their first-run answers. On the conflict rules where only time acts, all three models answer correctly when
told and choose the real-world outcome from the bare image; in text, one answer of the 18 is wrong. The consistent
rules are answered correctly in every condition. Two action rules show that naming the game is not always enough: a
barrel left without support in Minecraft and crops watered every day in Stardew Valley, where some models choose the
real-world outcome even when told.


\subsection{Rules of Terraria and Stardew Valley}
\label{app:arxiv_games}

Tables~\ref{tab:app_rules_terraria} and~\ref{tab:app_rules_stardew} list every rule of the two further games with the
question and outcomes of its first variant and the mean accuracy of the three models that answered them.

\begin{table*}[p]
\centering
\scriptsize
\setlength{\tabcolsep}{2.5pt}\renewcommand{\arraystretch}{0.8}\setlength{\abovecaptionskip}{5pt}
\begin{tabular}{@{}ll>{\raggedright\arraybackslash}p{0.32\textwidth}>{\raggedright\arraybackslash}p{0.215\textwidth}>{\raggedright\arraybackslash}p{0.215\textwidth}rrr@{}}
\toprule
Rule & Type & Question & Game outcome & Other outcome & T & Td & I \\
\midrule
T\_P01 & Pas. & After some time passes, what happens to the brown block floating in mid-air? & It stays exactly where it is. & It slowly slides down and rests on the ground. & 92 & 100 & 83 \\
\rowcolor{black!6}T\_P02 & Pas. & After some time passes, what happens to the grey block floating in mid-air? & It stays exactly where it is. & It falls straight down. & 92 & 75 & 8 \\
T\_P03 & Pas. & After some time passes, what happens to the reddish clay block hanging in the air? & It stays exactly where it is. & It falls and breaks apart on impact. & 92 & 92 & 33 \\
\rowcolor{black!6}T\_P04 & Pas. & After some time passes, what happens to the transparent glass block floating in mid-air? & It stays exactly where it is. & It falls and shatters on the ground. & 100 & 100 & 75 \\
T\_P05 & Pas. & After some time passes, what happens to the dark mud block floating in mid-air? & It stays exactly where it is. & It slowly drips downward. & 100 & 100 & 75 \\
\rowcolor{black!6}T\_P06 & Pas. & After some time passes, what happens to the dark ash block floating in mid-air? & It stays exactly where it is. & It slowly drifts downward like falling ash. & 92 & 100 & 92 \\
T\_P07 & Pas. & After some time passes, what happens to the shiny black block floating in mid-air? & It stays exactly where it is. & It falls straight down. & 75 & 50 & 33 \\
\rowcolor{black!6}T\_P08 & Pas. & After some time passes, what happens to the red brick block floating in mid-air? & It stays exactly where it is. & It falls and breaks on impact. & 100 & 100 & 75 \\
T\_P09 & Pas. & After some time passes, what happens to the gray brick block floating in mid-air? & It stays exactly where it is. & It slowly slides down. & 100 & 100 & 100 \\
\rowcolor{black!6}T\_P10 & Pas. & After some time passes, what happens to the golden brick block floating in mid-air? & It stays exactly where it is. & It falls straight down. & 92 & 100 & 58 \\
T\_P11 & Pas. & After some time passes, what happens to the silver brick block floating in mid-air? & It stays exactly where it is. & It falls and dents the ground. & 100 & 100 & 92 \\
\rowcolor{black!6}T\_P12 & Pas. & After a long time passes with nothing touching it, what happens to the cobweb? & It stays exactly as it is. & It slowly disintegrates and falls away. & 83 & 100 & 50 \\
T\_P13 & Pas. & After a long time underground with no sunlight, what happens to the green grass-covered block? & It stays green. & The grass turns brown and dies. & 75 & 83 & 0 \\
\rowcolor{black!6}T\_P14 & Pas. & After a long time passes, what happens to the rust-brown iron ore block? & It stays exactly as it is. & It slowly corrodes and crumbles. & 92 & 100 & 42 \\
T\_A01 & Act. & After the bottom block of this small tower is removed, what happens to the two blocks above it? & They stay floating in the air. & They both collapse to the ground. & 92 & 100 & 50 \\
\rowcolor{black!6}T\_A02 & Act. & After the stone block directly below the wooden platform is removed, what happens to the platform? & It stays exactly where it is. & It falls straight down. & 83 & 100 & 33 \\
T\_A03 & Act. & After the grey stone block directly below the clay block is removed, what happens to the clay block? & It stays exactly where it is. & It immediately falls down one level. & 92 & 100 & 17 \\
\rowcolor{black!6}T\_A05 & Act. & What happens if you place a wooden platform one block above the ground and try to walk up through it from below? & You pass straight through the platform from below. & You are blocked by the platform. & 92 & 83 & 92 \\
T\_A06 & Act. & After one block is removed from the middle of this row, what happens to the blocks on either side? & They stay exactly where they are. & The blocks on each side slide toward the gap. & 100 & 100 & 83 \\
\rowcolor{black!6}T\_A07 & Act. & After the rightmost supporting block is removed, leaving five blocks overhanging in mid-air, what happens? & All five blocks stay where they are. & The unsupported end collapses. & 100 & 100 & 67 \\
T\_A08 & Act. & After the central support block is removed and the two side blocks are left hanging, what happens to those two side blocks? & Both side blocks stay floating. & Both side blocks fall immediately. & 92 & 100 & 67 \\
\rowcolor{black!6}T\_A09 & Act. & After the stone block below the ash block is removed, what happens to the ash? & It stays exactly where it is. & It crumbles and scatters downward. & 83 & 75 & 8 \\
T\_A10 & Act. & After the stone block below the obsidian is removed, what happens to the obsidian? & It stays exactly where it is. & It falls straight down. & 75 & 58 & 33 \\
\rowcolor{black!6}T\_A11 & Act. & After the stone block below the mud block is removed, what happens to the mud? & It stays exactly where it is. & It slowly drips down. & 92 & 92 & 67 \\
T\_A12 & Act. & After the stone block below the clay block is removed, what happens to the clay? & It stays exactly where it is. & It falls and breaks apart on impact. & 100 & 100 & 25 \\
\rowcolor{black!6}T\_A13 & Act. & After the bottom-left block of this tall wall is removed, what happens to the rest of the wall? & The rest of the wall stays in place. & The whole wall topples over. & 100 & 100 & 100 \\
T\_A14 & Act. & After the stone block below the golden brick is removed, what happens to the gold brick? & It stays exactly where it is. & It falls straight down. & 92 & 75 & 25 \\
\rowcolor{black!6}T\_C01 & Cons. & After the stone block below the sand is removed, what happens to the sand? & It falls straight down. & It stays floating in mid-air. & 100 & 100 & 100 \\
T\_C02 & Cons. & After a stone block is placed on solid ground, what happens to it over time? & It stays exactly where it was placed. & It sinks slowly into the ground. & 100 & 100 & 83 \\
\rowcolor{black!6}T\_C03 & Cons. & After a dirt block is placed on solid ground, what happens to it? & It stays exactly where it was placed. & It sinks into the ground below. & 100 & 100 & 100 \\
T\_A04 & Cons. & After the stone block directly below the sand block is removed, what happens to the sand? & It falls straight down. & It stays in place. & 100 & 92 & 100 \\
\rowcolor{black!6}T\_C04 & Cons. & After a mud block is placed on solid ground, what happens to it? & It stays exactly where it was placed. & It slowly spreads outward like a liquid. & 100 & 100 & 100 \\
T\_C05 & Cons. & After an ash block is placed on solid ground, what happens to it? & It stays exactly where it was placed. & It drifts downward like falling ash. & 92 & 100 & 92 \\
\rowcolor{black!6}T\_C06 & Cons. & After a column of three stone blocks is built on solid ground, what happens to the column? & All three blocks stay stacked. & The top block slides off because there is nothing holding it. & 100 & 100 & 100 \\
T\_C07 & Cons. & After a gold brick is placed on solid ground, what happens to it? & It stays exactly where it was placed. & It slowly sinks due to its weight. & 100 & 100 & 100 \\
\rowcolor{black!6}T\_C08 & Cons. & After an obsidian block is placed on solid ground, what happens to it? & It stays exactly where it was placed. & It slowly sinks into the ground below. & 100 & 100 & 83 \\
T\_C09 & Cons. & After a clay block is placed on solid ground, what happens to it? & It stays exactly where it was placed. & It slowly spreads sideways like wet clay. & 100 & 100 & 100 \\
\rowcolor{black!6}T\_C10 & Cons. & After a red brick block is placed on solid ground, what happens to it? & It stays exactly where it was placed. & It slides sideways because bricks are smooth. & 100 & 100 & 100 \\
T\_C11 & Cons. & After a glass block is placed on solid ground and nothing hits it, what happens to it? & It stays intact exactly where it was placed. & It spontaneously shatters from temperature changes. & 100 & 100 & 100 \\
\rowcolor{black!6}T\_C12 & Cons. & After the stone block below a stack of two sand blocks is removed, what happens to the sand blocks? & Both sand blocks fall straight down. & Both sand blocks float in mid-air. & 100 & 100 & 100 \\
\bottomrule
\end{tabular}
\caption{Every rule of Terraria, with its type, the question and outcomes of its first variant, and
the mean accuracy (\%) of DeepSeek, Qwen3.8-Max and Doubao-Seed-2.1-Pro in text (T), when told (Td) and on the bare image
(I). T\_A04 (sand falls once its support is removed) is filed as an action rule but holds in reality too, so it is scored as consistent.}
\label{tab:app_rules_terraria}
\end{table*}

\begin{table*}[p]
\centering
\scriptsize
\setlength{\tabcolsep}{2.5pt}\renewcommand{\arraystretch}{0.8}\setlength{\abovecaptionskip}{5pt}
\begin{tabular}{@{}ll>{\raggedright\arraybackslash}p{0.32\textwidth}>{\raggedright\arraybackslash}p{0.215\textwidth}>{\raggedright\arraybackslash}p{0.215\textwidth}rrr@{}}
\toprule
Rule & Type & Question & Game outcome & Other outcome & T & Td & I \\
\midrule
S\_P01 & Pas. & Nobody waters the young plants in the tilled soil, and three hot, sunny days pass. What do the plants look like at the end? & They look exactly the same as now. & They wilt, droop and start to dry out. & 67 & 67 & 0 \\
\rowcolor{black!6}S\_P02 & Pas. & The vegetables in the image are ripe. Nobody picks them, and a week of sunny weather passes. What do they look like at the end of the week? & They are still ripe and ready to pick, just as now. & They become overripe and start to rot in the ground. & 100 & 100 & 17 \\
S\_P03 & Pas. & The vegetables in the image are ripe. Nobody picks them, and it rains every day for a week. What do they look like at the end of the week? & They are still ripe and ready to pick, just as now. & They become waterlogged and start to rot. & 100 & 100 & 17 \\
\rowcolor{black!6}S\_P04 & Pas. & The torch stuck in the ground is lit. Nobody touches it for a week. What does it look like at the end of the week? & It is still burning, just as now. & It has burned down and gone out. & 100 & 100 & 33 \\
S\_P05 & Pas. & It is raining, and it keeps raining for three days. What does the lit torch look like at the end? & It is still burning, just as now. & The rain has put the flame out. & 100 & 100 & 33 \\
\rowcolor{black!6}S\_P06 & Pas. & The campfire is burning. Nobody adds any wood, and a week passes. What does it look like at the end of the week? & It is still burning, just as now. & It has burned out, leaving a pile of ash. & 83 & 100 & 0 \\
S\_P07 & Pas. & The fruit on the tree is ripe. Nobody picks it, and a week passes. What does the tree look like at the end of the week? & The fruit is still hanging on the tree. & The fruit has fallen to the ground and started to rot. & 92 & 100 & 67 \\
\rowcolor{black!6}S\_P08 & Pas. & Nobody waters the young plants in the tilled soil, and a whole week of hot, sunny weather passes. What do the plants look like at the end? & They look exactly the same as now. & They have withered and died. & 58 & 33 & 0 \\
S\_P09 & Pas. & It rains every day for a week, and nobody tends the seedlings in the tilled soil. What do they look like at the end of the week? & They are fully grown and ready to harvest. & They are a little taller, but still young plants. & 50 & 17 & 25 \\
\rowcolor{black!6}S\_P10 & Pas. & The weeds in the image are left alone for two weeks of warm, rainy spring weather. What do they look like at the end? & They look exactly the same as now. & They have grown taller and spread. & 67 & 44 & 0 \\
S\_P11 & Pas. & The wild plant growing in the grass is left alone for a week. What does it look like at the end of the week? & It looks exactly the same as now. & It has wilted and withered. & 92 & 92 & 42 \\
\rowcolor{black!6}S\_P13 & Pas. & The tree stump in the image is left alone for a whole spring. What does it look like at the end? & It looks exactly the same as now. & New shoots have sprouted from it. & 92 & 58 & 50 \\
S\_P14 & Pas. & The wooden chest stands outdoors and it rains every day for two weeks. What does the chest look like at the end? & It looks exactly the same as now. & The wood has darkened and started to swell and warp. & 100 & 100 & 0 \\
\rowcolor{black!6}S\_A01 & Act. & Someone chops the tree with an axe until it falls. What is left where the tree stood right afterwards? & A stump, with small pieces of wood scattered around it. & A stump, with the whole trunk lying on the ground beside it. & 58 & 75 & 75 \\
S\_A02 & Act. & Someone chops the tree with an axe until it falls. What happens to the plants growing right next to the tree? & They are untouched. & They are crushed by the falling tree. & 92 & 92 & 75 \\
\rowcolor{black!6}S\_A03 & Act. & Someone hits the rock once with a pickaxe. What happens to the rock right afterwards? & It disappears, and a few loose stones pop out where it was. & It cracks, but stays in place. & 75 & 33 & 67 \\
S\_A04 & Act. & Someone hits the fallen branch once with an axe. What happens to the branch right afterwards? & It disappears, and a few small pieces of wood pop out where it was. & It is split, and the two halves stay on the ground. & 100 & 92 & 100 \\
\rowcolor{black!6}S\_A05 & Act. & Someone waters the seedlings in the tilled soil every morning for one week of sunny weather. What do they look like at the end of the week? & They are fully grown and ready to harvest. & They are a little taller, but still young plants. & 67 & 0 & 0 \\
S\_A06 & Act. & Someone hits the wooden chest once with a pickaxe. What happens to the chest right afterwards? & It pops off the ground in one piece, completely intact. & Its wood splinters and cracks where it was hit. & 100 & 100 & 92 \\
\rowcolor{black!6}S\_A07 & Act. & Someone hits the lit torch once with a pickaxe. What happens to the torch right afterwards? & It pops out of the ground whole, and the flame is gone. & It snaps in two, and the burning end falls to the ground. & 100 & 100 & 100 \\
S\_A08 & Act. & Someone hits the fence post once with an axe. What happens to the fence post right afterwards? & It pops out of the ground in one piece, completely intact. & It gets a deep cut, but stays standing. & 100 & 92 & 67 \\
\rowcolor{black!6}S\_A09 & Act. & Someone hits the burning campfire once with a pickaxe. What happens right afterwards? & The whole campfire pops off the ground in one piece, and the fire is gone. & The burning logs are knocked apart and scattered. & 92 & 100 & 100 \\
S\_A10 & Act. & Someone cuts the weeds with a scythe. What is left where the weeds were right afterwards? & Nothing but a few loose fibers on the bare ground. & Short cut stems still rooted in the ground. & 92 & 92 & 25 \\
\rowcolor{black!6}S\_A11 & Act. & Someone chops the small sapling once with an axe. What happens to it right afterwards? & It disappears completely, leaving nothing on the ground. & It is cut off, and the cut stem lies on the ground. & 58 & 50 & 42 \\
S\_A12 & Act. & Someone chops the tree with an axe until it falls. What happens to the wooden chest standing right next to it? & It is untouched. & It is smashed by the falling tree. & 92 & 100 & 100 \\
\rowcolor{black!6}S\_C01 & Cons. & The tilled soil was watered this morning. A hot, sunny day passes and nobody waters it again. What does the soil look like the next morning? & It has dried out and turned lighter. & It looks exactly as wet as now. & 92 & 100 & 100 \\
S\_C02 & Cons. & The tilled soil is dry. It rains all night. What does the soil look like the next morning? & It is dark and wet. & It looks exactly as dry as now. & 100 & 100 & 100 \\
\rowcolor{black!6}S\_C04 & Cons. & Someone chops the tree with an axe many times. What happens to the tree? & It is cut down. & It stays standing, without a scratch. & 100 & 100 & 100 \\
S\_C05 & Cons. & Someone waters the young plants in the tilled soil every morning for three sunny days. What do they look like at the end? & They have grown bigger. & They look exactly the same as now. & 92 & 83 & 100 \\
\rowcolor{black!6}S\_C06 & Cons. & Someone pulls up one of the ripe vegetables by hand. What is left in that spot right afterwards? & Bare tilled soil. & The same ripe vegetable, still growing there. & 100 & 92 & 100 \\
S\_C07 & Cons. & Seeds were just planted in the tilled soil. Nobody waters them, and three hot, sunny days pass. What does the soil look like at the end? & Still bare, with no sprouts. & Covered in fully grown plants. & 100 & 100 & 100 \\
\rowcolor{black!6}S\_C08 & Cons. & Someone pours water from a watering can onto the dry tilled soil. What does the soil look like right afterwards? & It is dark and wet. & It looks exactly as dry as now. & 100 & 100 & 100 \\
S\_C09 & Cons. & Someone picks the ripe strawberries from one of the plants by hand. What is left in that spot right afterwards? & The strawberry plant, still growing there. & Bare tilled soil. & 67 & 100 & 100 \\
\rowcolor{black!6}S\_C10 & Cons. & Someone hits the rock with a pickaxe again and again. What happens to the rock? & It breaks apart. & It stays in place, without a scratch. & 100 & 100 & 100 \\
S\_C11 & Cons. & Someone cuts the weeds with a scythe. What happens to the weeds? & They are cut down. & They stay standing, untouched. & 100 & 100 & 100 \\
\rowcolor{black!6}S\_C12 & Cons. & Someone turns the grassy ground over with a hoe. What does that spot look like right afterwards? & Freshly tilled, bare soil. & The same untouched ground as before. & 100 & 92 & 100 \\
\bottomrule
\end{tabular}
\caption{Every rule of Stardew Valley, as in Table~\ref{tab:app_rules_terraria}.}
\label{tab:app_rules_stardew}
\end{table*}

\subsection{Code and Data}
\label{app:package}

\noindent{\small Code: \href{\codeurl}{\nolinkurl{github.com/GaoJianXin05/RuleCraft}}\\
Data: \href{\dataurl}{\nolinkurl{huggingface.co/datasets/GGjx/RuleCraft}}\par}
\smallskip

\noindent The data hold every rule and all 1{,}240 questions
with their options, answer keys and images, the engine check of every variant, the Terraria and Stardew Valley
questions with their images and engine checks, every model answer the paper reports, including the open answers, their
judge labels, the reasoning traces and the three models' notes, and the pseudonymized answers of the human study. The
code builds the scenes, checks them in the engine and renders the images, runs every experiment, prints every prompt
verbatim, and recomputes Tables~\ref{tab:phen}, \ref{tab:games} and \ref{tab:reasoning} and the human results from
the released answers without any API call. Code is released under the MIT license and data under CC BY 4.0.

\end{document}